\documentclass[10pt,twocolumn,letterpaper]{article}
\usepackage[T1]{fontenc}   %

\usepackage[pagenumbers]{cvpr} %

\usepackage{float}
\usepackage{multirow}
\usepackage{currfile}
\usepackage{caption}
\usepackage{booktabs}
\usepackage{colortbl}
\usepackage{makecell}   %
\usepackage{adjustbox}  %
\usepackage{array}
\usepackage{arydshln}  %
\newcolumntype{I}{!{\color{black!45}\vrule width 0.4pt}}  %
\usepackage{amsmath,amssymb}
\usepackage{tikz}
\usetikzlibrary{arrows.meta,positioning,calc,fit,backgrounds}
\usepackage{enumitem}
\setlist{nosep,leftmargin=*}
\usepackage{xspace}

\newcommand{\qalign}{Q-Align\xspace}
\newcommand{\liqe}{LIQE\xspace}
\newcommand{\dinov}{DINOv3\xspace}

\definecolor{posgreen}{RGB}{27,120,85}
\definecolor{negred}{RGB}{178,60,55}
\newcommand{\good}[1]{\textcolor{posgreen}{#1}}
\newcommand{\bad}[1]{\textcolor{negred}{#1}}
\newcommand{\nsig}[1]{\textcolor{black!45}{#1}}

\newcommand{\numAdaIntUpPPR}{seven}

\newcommand{\numCalN}{953}
\newcommand{\numCalReachDt}{77.3}
\newcommand{\numCalReachRT}{82.8}

\newcommand{\numCsrN}{499}
\newcommand{\numCsrSeen}{450}

\newcommand{\numDtLossMaxFrontier}{two}

\newcommand{\numDtPosCells}{26}

\newcommand{\numDtWinMinFrontier}{five}
\newcommand{\numExpGPTQAPPR}{0.082}

\newcommand{\numExpNeurOPPSNR}{20.43}
\newcommand{\numExpOrigDE}{10.72}
\newcommand{\numExpOrigLPIPS}{0.135}
\newcommand{\numExpOrigPSNR}{18.45}
\newcommand{\numExpOrigQAFiveK}{0.091}
\newcommand{\numExpOursDE}{10.79}
\newcommand{\numExpOursLPIPS}{0.164}
\newcommand{\numExpOursPSNR}{18.68}
\newcommand{\numExpOursQAFiveK}{0.078}
\newcommand{\numExpOursQAPPR}{0.129}

\newcommand{\numGPTPosCells}{24}

\newcommand{\numHumAccPaq}{53.1}
\newcommand{\numHumAccQA}{64.0}
\newcommand{\numHumAccRater}{67.3}

\newcommand{\numHumClaudeAll}{50.5}
\newcommand{\numHumClaudeAllCI}{46.8--54.2}

\newcommand{\numHumClaudeGap}{11.9}
\newcommand{\numHumClaudeGapCI}{3.3--20.6}

\newcommand{\numHumGPTAll}{54.3}

\newcommand{\numHumGPTp}{p{=}0.031}
\newcommand{\numHumInvalid}{73}

\newcommand{\numHumPairs}{1{,}837}
\newcommand{\numHumRaters}{157}

\newcommand{\numHumSupAll}{46.3}
\newcommand{\numHumSupAllCI}{41.5--51.2}

\newcommand{\numHumTie}{9.8}
\newcommand{\numHumTieShift}{0.8}
\newcommand{\numHumVotes}{9{,}253}
\newcommand{\numLossMaxClaude}{two}

\newcommand{\numLossMaxRetouch}{two}

\newcommand{\numMeanAdaIntQAFiveK}{+0.134}

\newcommand{\numMeanAdaIntQASPAQ}{$-$0.011}

\newcommand{\numMeanNeurOPQAFiveK}{+0.137}

\newcommand{\numMeanNeurOPQASPAQ}{$-$0.009}

\newcommand{\numMeanOursQASPAQ}{+0.125}

\newcommand{\numNeurOPUpPPR}{five}

\newcommand{\numOursPosCells}{26}

\newcommand{\numOursUpPPR}{nine}

\newcommand{\numQHighFiveKClaude}{one}
\newcommand{\numQHighFiveKOurs}{two}
\newcommand{\numQHighPPRClaude}{one}
\newcommand{\numQHighPPROurs}{one}
\newcommand{\numQHighSPAQClaude}{three}
\newcommand{\numQHighSPAQOurs}{four}
\newcommand{\numQLowFiveKClaude}{one}
\newcommand{\numQLowFiveKOurs}{four}
\newcommand{\numQLowPPRClaude}{none}
\newcommand{\numQLowPPROurs}{six}
\newcommand{\numQLowSPAQClaude}{none}
\newcommand{\numQLowSPAQOurs}{four}

\newcommand{\numSolverSec}{42}

\newcommand{\numVEditorDtFiveK}{1\,/\,6\,/\,2}

\newcommand{\numVEditorDtPPR}{0\,/\,4\,/\,5}
\newcommand{\numVEditorDtSPAQ}{2\,/\,4\,/\,3}
\newcommand{\numWTLAdaIntFiveK}{7\,/\,1\,/\,1}

\newcommand{\numWTLClaudeFiveK}{4\,/\,3\,/\,2}
\newcommand{\numWTLClaudePPR}{3\,/\,5\,/\,1}
\newcommand{\numWTLClaudeSPAQ}{6\,/\,3\,/\,0}

\newcommand{\numWTLGPTFiveK}{9\,/\,0\,/\,0}
\newcommand{\numWTLGPTPPR}{8\,/\,1\,/\,0}
\newcommand{\numWTLGPTSPAQ}{9\,/\,0\,/\,0}

\newcommand{\numWTLHeldClaudeFiveK}{2\,/\,2\,/\,2}
\newcommand{\numWTLHeldClaudePPR}{3\,/\,2\,/\,1}
\newcommand{\numWTLHeldClaudeSPAQ}{3\,/\,3\,/\,0}

\newcommand{\numWTLNeurOPFiveK}{8\,/\,0\,/\,1}

\newcommand{\numWinMaxClaude}{six}

\newcommand{\numWinMinClaude}{three}
\newcommand{\numWinMinGPT}{eight}
\newcommand{\numWinMinRetouch}{seven}
\newcommand{\numWinMinSupOwn}{eight}
\newcommand{\numClaudeCalls}{1,881}\newcommand{\numClaudeValid}{1,881}\newcommand{\numClaudeSec}{3.3}
\newcommand{\numGapRsqFiveK}{0.54}\newcommand{\numGapSharpRFiveK}{0.58}
\newcommand{\numGapRsqPPR}{0.52}\newcommand{\numGapSharpRPPR}{0.67}
\newcommand{\numGapRsqSPAQ}{0.23}\newcommand{\numGapSharpRSPAQ}{0.28}
\newcommand{\numGapClipR}{0.09}\newcommand{\numGapTopSPAQ}{highlight level}
\newcommand{\numPaqDesignFiveK}{+1.78}\newcommand{\numPaqNativeFiveK}{+6.99}
\newcommand{\numFigQOrig}{4.01}\newcommand{\numFigQOurs}{4.48}\newcommand{\numFigAOrig}{3.04}\newcommand{\numFigAOurs}{3.73}\newcommand{\numFigQDt}{4.23}\newcommand{\numFigADt}{3.71}
\newcommand{\figonemoves}{\mpill{\icosha}{sharpness}{+28}\\[1.5pt]\mpill{\icoden}{noise}{+17}\\[1.5pt]\mpill{\icocol}{colourfulness}{+16}\\[1.5pt]\mpill{\icohaz}{haze}{-7}\\[1.5pt]\mpill{\icocon}{local contrast}{+5}\\[1.5pt]\mpill{\icobri}{highlight clip.}{+4}}
\newcommand{\figonesliders}{sharpness $+1.00$, colourfulness $+0.68$, contrast $+0.22$, temperature $+0.17$, shadow depth $+0.16$, highlight level $-0.14$, noise $-0.10$}
\newcommand{\figoneslidersdt}{fine detail $+1.00$, colourfulness $+0.92$, sharpness $+0.53$, highlight clipping $+0.31$, brightness $+0.22$, shadow depth $+0.19$, noise $-0.18$, highlight level $+0.14$}
\newcommand{\numTrainLiftMed}{0.25}\newcommand{\numTrainLiftHi}{0.89}\newcommand{\numTestLiftMedFiveK}{0.63}\newcommand{\numTestLiftAboveFiveK}{5}\newcommand{\numTestLiftMedPPR}{0.40}\newcommand{\numTestLiftAbovePPR}{0}\newcommand{\numTestLiftMedSPAQ}{0.87}\newcommand{\numTestLiftAboveSPAQ}{47}
\newcommand{\numPaperSize}{10M}\newcommand{\numPaperRows}{10{,}000{,}000}\newcommand{\numPaperHiPct}{68}\newcommand{\numPaperSteps}{144{,}000}\newcommand{\numPaperBest}{141{,}000}\newcommand{\numPaperParams}{54M}\newcommand{\numPaperTrainMin}{58}

\newcommand{\numGPTValid}{1,879}\newcommand{\numGPTFailed}{2}\newcommand{\numGPTSec}{14.6}
\newcommand{\numAttrCandN}{33}
\newcommand{\numCapOthers}{0.55}
\newcommand{\numClipK}{4}
\newcommand{\numClipMZeroPct}{0.5}
\newcommand{\numClipT}{250.5}
\newcommand{\numDcpOthers}{0.74}
\newcommand{\numDefClipRise}{40}
\newcommand{\numDefCopies}{five}
\newcommand{\numDefLiqeHi}{3.85}
\newcommand{\numDefLiqeLo}{2.3}
\newcommand{\numDefMaxHi}{1.15}
\newcommand{\numDefMaxLo}{0.6}
\newcommand{\numDefMaxMid}{1.0}
\newcommand{\numDefOpsHi}{three}
\newcommand{\numDefOpsLo}{two}
\newcommand{\numDefSMin}{0.3}
\newcommand{\numDefThirdCap}{0.85}
\newcommand{\numDtHazeT}{0, 0.2, 0.4, 0.7 and 1}
\newcommand{\numDtHazeV}{0, 0.31, 0.54, 0.82 and 1}
\newcommand{\numEightMSteps}{72{,}000}

\newcommand{\numExposureBrightRho}{1.00}
\newcommand{\numLapNoiseLeak}{+1.00}
\newcommand{\numLapOthers}{0.58}

\newcommand{\numMlvNoiseLeak}{+0.22}
\newcommand{\numMlvOthers}{0.63}
\newcommand{\numNoiseJpegOurs}{0.97}
\newcommand{\numNoiseJpegPCA}{0.58}
\newcommand{\numNoiseResizeOurs}{0.91}
\newcommand{\numNoiseResizePCA}{0.78}
\newcommand{\numPilotPhotos}{5}
\newcommand{\numQualPcts}{90th, 50th and 10th}
\newcommand{\numRobustN}{1{,}000}
\newcommand{\numSelSetN}{500}
\newcommand{\numSelTau}{0.99}
\newcommand{\numSelTauGrid}{0.90, 0.97 and 0.99}
\newcommand{\numSolverCorr}{twice}
\newcommand{\numSolverDt}{$b{=}0.2$; weight 0.3 for highlight clipping, 1.4 for sharpness and 1 for the rest; slack $-0.5$ for highlight level, haze, highlight clipping and local contrast, and 0 for the rest.}
\newcommand{\numSolverFrame}{1024}
\newcommand{\numSolverHarm}{4}
\newcommand{\numSolverHuber}{1.5}
\newcommand{\numSolverMaxRenders}{50}
\newcommand{\numSolverRT}{$b{=}0.2$; weight 1 and slack 0 for every attribute.}
\newcommand{\numSolverTol}{2}
\newcommand{\numSpaqRotated}{202}
\newcommand{\numSpaqWideGamut}{30}
\newcommand{\numTenNoiseLeak}{+1.00}
\newcommand{\numTenOthers}{0.60}
\newcommand{\numTrainBatch}{32{,}768}
\newcommand{\numTrainBatchGPU}{4096}
\newcommand{\numTrainEvalEvery}{1000}
\newcommand{\numTrainGPUs}{8}
\newcommand{\numTrainLR}{3{\cdot}10^{-3}}
\newcommand{\numTrainLRFloor}{3{\cdot}10^{-4}}
\newcommand{\numTrainSelFromPct}{20}
\newcommand{\numTrainWD}{0.05}
\newcommand{\numTrainWarmPct}{10}
\definecolor{cvprblue}{rgb}{0.21,0.49,0.74}
\usepackage[pagebackref,breaklinks,colorlinks,allcolors=cvprblue]{hyperref}

\def\confName{CVPR}
\def\confYear{2027}

\title{ATune: Unsupervised Photo Retouching Through Image-Editor Controls}

\author{Divyanshu Goyal\textsuperscript{*}\\
{\tt\small divgoyal@adobe.com}
\and
Gabriel Ruiz\textsuperscript{*}\\
{\tt\small garuiz@adobe.com}
\and
Amir Meisami\textsuperscript{\ddag}\\
{\tt\small meisami@adobe.com}
}

\newcommand{\mainsec}[2]{\cref{#1}}
\newcommand{\maintab}[2]{\cref{#1}}

\begin{document}
\twocolumn[{%
\renewcommand\twocolumn[1][]{#1}%
\maketitle
\vspace{-1.4em}
\begin{center}
\input{fig/pipeline}
\end{center}
\vspace{0.4em}
}]
{\let\thefootnote\relax\footnotetext{\textsuperscript{*}Equal contribution.\quad\textsuperscript{\ddag}Corresponding author.}}
\begin{abstract}
Photo retouching models are trained on pairs of original and retouched photographs, usually edited by experts. Expert
edits are costly, so the two main public sets, FiveK and PPR10K, hold only 16{,}161 originals. Good photographs on their own are far easier to find:
commercial stock collections and free ones such as Pixabay and Unsplash hold millions. We learn retouching from them alone: we damage each photograph the way a camera or a careless edit
would, which gives a worse and a better version of the same photograph. From these pairs, our model, ATune, learns how 13 measured
attributes, such as brightness, contrast and sharpness, should change to raise a photograph's quality.
The edit is declarative: ATune sets target values for the attributes, and a simple search adjusts the sliders of an
out-of-the-box photo editor until the photograph reaches them. Every
edit is a \emph{list of slider values} a person can read and adjust. The same model can drive any editor that makes
those changes; we test RawTherapee and darktable. We train on \numPaperSize{} photographs from a commercial stock collection and test
on FiveK, PPR10K and SPAQ. Nine quality judges score every edit, and in a user study \numHumRaters{} human raters
made \numHumVotes{} choices between our edit and another method's edit of the same photograph.
When Claude Opus 5 and GPT-6 Astra edit the same photographs in the same editor, raters prefer our edits to GPT-6 Astra's
(\numHumGPTAll\%) and rate them even with Claude's (\numHumClaudeAll\%). We also compare with supervised retouchers on their home ground: NeurOP on FiveK and AdaInt on PPR10K, each trained on
that set's own expert edits. Raters chose our edits in \numHumSupAll\% of those comparisons. ATune, a \numPaperParams-parameter model trained without a
single expert edit, now matches frontier models at setting an editor's sliders.
\end{abstract}

\section{Introduction}
\label{sec:intro}

Photo retouching models learn from pairs of original and expert-retouched photographs. Expert edits are costly, so
MIT-Adobe FiveK~\cite{bychkovsky2011fivek} from 2011 and PPR10K~\cite{liang2021ppr10k} from 2021 hold only 16{,}161
originals between them. A model trained on so few
photographs fits that one collection, and it does worse on photographs from other sources (\cref{sec:results}). Good
photographs on their own are far easier to find. Stock
collections, commercial ones and free ones such as Pixabay and Unsplash, hold millions of
photographs that already look right, and none of them comes with a retouched pair.

A good photograph can be turned into a training pair by degrading it: underexpose it, flatten its contrast or blur it,
and the original is the target. Millions of such pairs teach a model which changes raise a photograph's quality.
Distort-and-Recover~\cite{park2018distort} trains this way too, but its model outputs a sequence of pixel operations, so
its edits are tied to those operations. ATune outputs a goal: how 13 measured attributes of the photograph, such as brightness, contrast, sharpness and haze, should change.
The first is an \emph{imperative} edit and the second a \emph{declarative} one. Any editor whose tools move these attributes can
reach the goal, so the same model works across editors without retraining.

The method has three steps (\cref{fig:pipeline}). \textbf{Measure:} each of the 13 attributes is measured directly from the
photograph, and the measures were selected by a reliability test (\cref{sec:attributes}). \textbf{Predict:} ATune learns how the degraded copy's attributes
must change to get back to the original, given how much the quality should rise. At test time that quality rise is a
\emph{dial}: we ask for the changes that bring a new photograph up to the collection's $\tau$-th quality percentile
(\cref{sec:method-target}). \textbf{Act:} a simple search adjusts the calibrated controls of an out-of-the-box open-source
editor, RawTherapee~\cite{rawtherapee} or darktable~\cite{darktable}, rendering and re-measuring until the attributes
land (\cref{sec:editing}). Every edit is a short list of named slider values a photographer can read and adjust.

We train ATune on \numPaperSize{} photographs from a commercial stock collection and test on FiveK, PPR10K and SPAQ. We compare with two
kinds of method. The first is frontier models: Claude Opus 5~\cite{anthropic2026opus5} and GPT-6
Astra~\cite{openai2026gpt6} get the same photograph and editor as ATune and set its sliders in one call, with reasoning on (Claude at maximum effort, GPT-6 Astra at high). The second is seven learned retouchers. Two of them, AdaInt and NeurOP, trained on FiveK's and PPR10K's training splits, so they are tested on
collections they learned from; CSRNet's weights trained on \numCsrSeen{} of
the \numCsrN{} FiveK test photographs themselves. Nine published quality judges score every edit, and in a user study
\numHumRaters{} human raters made \numHumVotes{} choices between our edit and another method's edit of the same photograph
(\cref{sec:setup,sec:human}).

Against GPT-6 Astra, at least \numWinMinGPT{} of the nine judges prefer our edits on every test set, and raters chose ours in
\numHumGPTAll\% of votes. Against Claude, \numWinMinClaude{} to \numWinMaxClaude{} judges prefer ours and at most
\numLossMaxClaude{} prefer Claude's, while raters' votes are a tie at \numHumClaudeAll\% (95\% CI
\numHumClaudeAllCI). The same holds on darktable, where at least \numDtWinMinFrontier{} judges prefer our edits to
either model's on every test set (\cref{sec:frontier,sec:human}). Against the seven learned retouchers, at least
\numWinMinRetouch{} judges prefer ours on every test set (\cref{tab:wtl}). On their home ground, NeurOP on FiveK and AdaInt
on PPR10K, \numWinMinSupOwn{} judges prefer ours, while raters chose ours in \numHumSupAll\% of comparisons (95\% CI
\numHumSupAllCI; \cref{sec:human}). Finally, we train models on 1M to \numPaperSize{} photographs and find that more training photographs help: no judge
ever prefers the model trained on the smaller set (\cref{sec:ablations}).

Our contributions:
\begin{enumerate}
  \item \textbf{A declarative, editor-independent edit.} ATune describes an edit as a change in 13 measured
  attributes, with no pixels and no sliders. A simple search then reaches that change with any editor's sliders, so
  the same model drives RawTherapee and darktable, and every edit is a readable list of slider values
  (\cref{sec:method,sec:frontier}).
  \item \textbf{Retouching learned without expert edits.} Degrading good photographs and learning to reverse the
  degradation turns \numPaperSize{} unedited photographs into training data. The model improves as the training set
  grows, with no judge preferring a model trained on fewer photographs (\cref{sec:ablations}).
  \item \textbf{Parity with frontier models from a \numPaperParams-parameter model.} With Claude Opus 5 and GPT-6
  Astra setting the same editor's sliders, raters prefer ATune's edits to GPT-6 Astra's and rate them even with
  Claude's, and the judges favour ATune's over both (\cref{sec:frontier,sec:human}).
\end{enumerate}

We compare only with methods that adjust the photograph's tone and colour and keep its content. Generative editors
that synthesise new pixels~\cite{wu2025qwenimage} are out of scope.

\section{Related work}
\label{sec:related}

\paragraph{Supervised retouching.} Most learned retouchers are trained on the FiveK dataset~\cite{bychkovsky2011fivek} and learn to reproduce its
expert retouches as pixels, curves or filters: HDRNet~\cite{gharbi2017hdrnet},
DPE~\cite{chen2018dpe}, CSRNet~\cite{he2020csrnet}, DeepLPF~\cite{moran2020deeplpf}, image-adaptive 3D
LUTs~\cite{zeng2020lut}, AdaInt~\cite{yang2022adaint} and neural colour operators (NeurOP)~\cite{wang2022neurop}. FiveK has
five experts' edits, but these methods are usually trained on one of them, expert C, and so reproduce that expert's
style. Their output is the edited image itself, so a
user cannot see or adjust the edit as slider values. The exception is Neural Photo-Finishing~\cite{tseng2022neural},
which predicts a raw editor's slider settings, but it needs a target style to imitate and a neural network that mimics the
editor in order to train.

\paragraph{Unpaired and self-supervised enhancement.} Several methods avoid expert pairs by learning from a signal
other than a retouched target. Exposure~\cite{hu2018exposure}, Distort-and-Recover~\cite{park2018distort} and Kosugi
and Yamasaki~\cite{kosugi2020unpaired} use reinforcement learning to pick a sequence of editing operations, such as
brightening or adding contrast, from a fixed set that each method defines and implements itself. Distort-and-Recover
degrades good photographs to make its training signal, as we do, and is the closest prior method; synthetic
degradation is also how restoration methods make their training pairs~\cite{brooks2019unprocessing,wang2021realesrgan}. Kosugi and Yamasaki
and UEGAN~\cite{ni2020uegan} instead learn from a discriminator, which still needs FiveK's expert retouches as the set
of desirable images. Others learn from a quality score: Talebi and Milanfar~\cite{talebi2018learned} train an
enhancement network to raise the score of a learned aesthetic judge, Deng \etal~\cite{deng2018aesthetic} learn their
operators adversarially from good and poor aesthetic labels, and Gao \etal~\cite{gao2025quality} let the user set a
target quality score for a portrait enhancer. EnlightenGAN~\cite{jiang2021enlightengan}, Zero-DCE~\cite{guo2020zerodce}
and SCI~\cite{ma2022sci} learn without pairs but target low light only. All of these methods
output an imperative edit, a policy or network over one fixed set of operations, so moving to another editor means
training again. ATune outputs a declarative edit, a change in measured attributes learned from millions of photographs
with no paired edit, and a simple search reaches that change with the tools of any out-of-the-box editor.

\paragraph{Language-model retouching agents.} Multimodal language models can now operate photo editors.
MonetGPT~\cite{dutt2025monetgpt} fine-tunes a multimodal model on procedurally edited expert photographs to propose
editing operations with their parameters. JarvisArt~\cite{lin2025jarvisart} is fine-tuned, with supervision and then reinforcement
learning, to drive Lightroom's tools from a user's instruction; its MMArt release provides the PPR10K retoucher renders
we evaluate against. Similar agents orchestrate restoration tools~\cite{chen2024restoreagent,zhu2025agenticir}. All of
these need a large model, curated instruction data and expert edits to fine-tune on. We instead compare with frontier
models used as they are: Claude Opus 5~\cite{anthropic2026opus5} and GPT-6 Astra~\cite{openai2026gpt6} set the sliders
of the same editor as ATune.

\section{Method}
\label{sec:method}

\subsection{The attribute space}
\label{sec:attributes}

When a photographer judges a photograph, they talk about qualities such as its brightness, contrast, sharpness,
colour balance, noise etc. ATune works with 13 such attributes. It uses them for three things: to describe a
photograph, to decide how the photograph should change, and to check that the editor delivered the change. Because
everything depends on these numbers, each attribute needs a measure that is reliable, that an editor can move, and
that does not duplicate another. A bad measure would corrupt the training targets, the edits and the check on the
edits at once.

We therefore chose each measure in a selection study on public photographs, KADID-10k~\cite{lin2019kadid} and
SPAQ~\cite{fang2020spaq}. A candidate measure had to track a known dose of its own
distortion, ignore other distortions, and add something the other measures did not. All of the chosen measures except
the two clipping descriptors are stable: resizing, JPEG re-encoding or a small crop leaves the ordering of
photographs by that measure almost unchanged. The protocol, the
coverage and the rejected candidates are in the supplementary material.

\Cref{tab:attributes} lists the 13 attributes in three groups, tone, colour and detail, with the measure behind each.
The clipping and noise measures are less obvious than the rest. \textbf{Clipping} is measured separately for the
shadows and the highlights as a signed value: positive when pixels are clipped to pure black or white (growing with
the clipped area), negative when there is still room before clipping, so it passes smoothly through zero.
\textbf{Noise} is estimated from the pixel variation in the smoothest patches of the image, such as a wall or a sky.
We chose this estimator because it is stable under JPEG re-encoding and resizing (the ordering of SPAQ photographs
before and after agrees with Spearman $\numNoiseJpegOurs$ and $\numNoiseResizeOurs$), where the classical PCA
estimator~\cite{liu2013noise} is not ($\numNoiseJpegPCA$, $\numNoiseResizePCA$). Its limitation is that fine texture and sharpening also add variation in
smooth patches, so it cannot separate them from noise~\cite{colom2013ponomarenko}; it reads as the amount of
fine-grained variation, whatever its source.

\begin{table}[t]
\centering
\caption{\textbf{The 13 attributes} and the measure behind each one. All are computed on the analysis frame
(\cref{sec:attributes}).}
\label{tab:attributes}
\footnotesize\setlength{\tabcolsep}{4pt}\renewcommand{\arraystretch}{1.08}
\begin{adjustbox}{max width=\linewidth}
\begin{tabular}{@{}lll@{}}
\toprule
Group & Attribute & Measure \\
\midrule
Tone & brightness & median log intensity \\
 & contrast & interquartile range of CIELAB $L^*$ \\
 & shadow depth & $L^*$ at the 5th percentile \\
 & highlight level & $L^*$ at the 95th percentile \\
 & local contrast & midtone-weighted unsharp-mask energy, large radius \\
 & haze & mean dark channel~\cite{he2009haze} \\
 & shadow clipping & signed clipped area or headroom, at black \\
 & highlight clipping & signed clipped area or headroom, at white \\
\midrule
Colour & colourfulness & Hasler and S\"usstrunk~\cite{hasler2003colorfulness} \\
 & colour temperature & McCamy CCT~\cite{mccamy1992cct} of the mean colour \\
 & tint & mean $a^*$ \\
\midrule
Detail & sharpness & spread of the max.\ local variation map~\cite{bahrami2014mlv} \\
 & noise & Immerk\ae r residual~\cite{immerkaer1996noise}, flattest $8{\times}8$ blocks, two scales \\
\bottomrule
\end{tabular}
\end{adjustbox}
\end{table}

\textbf{One frame, one ruler.} Every attribute is measured on the same \emph{analysis frame}: the decoded image
resampled once to a 1024-px short side with Lanczos resampling~\cite{duchon1979lanczos}, in the corpus, inside the search and at evaluation. Each
attribute is divided into 100 equal-width \emph{buckets} between its 0.5th and 99.5th percentiles over the training
corpus (equal steps of $\log(1{+}x)$ for sharpness, temperature and noise). These 100 bucket boundaries are
computed once from the training corpus and then used unchanged for every photograph, including the test sets. Buckets are the unit for everything that follows: the model
predicts changes in buckets, and the search measures how far an edit still is from its target in buckets.

\subsection{Supervision from a quality-ranked corpus}
\label{sec:corpus}

\paragraph{Corpus.} We draw photographs from a licensed commercial stock collection: photographs only, no items flagged
as AI-generated, short side at least 1024\,px. We sample from every band of their precomputed \qalign{} quality score,
from weak to polished, so that the model learns to edit photographs at every quality level. Our training corpus has \numPaperSize{}
photographs, \numPaperHiPct\% of them at \qalign{} $\geq 4$.

\paragraph{Transitions.} The corpus holds only finished photographs, so we generate the degraded versions. Each
photograph gets five \emph{copies}, and each copy receives two or three of twelve camera-like defects: exposure, black
point, highlight blow-out or roll-off, contrast, temperature, tint, colourfulness, haze veil, local-contrast
flattening, blur or oversharpening, three kinds of noise, and JPEG compression. Defect strengths are drawn from a range
that grows with the original's quality, so a weak photograph is not damaged beyond recognition. The original is the
target for each of its copies, so no expert edit is needed, and because the originals range from mediocre to excellent,
the training transitions start at every quality level rather than all ending at the top.

Quality is measured on the same frame as the attributes, with \liqe~\cite{zhang2023liqe}. For every copy we record its
13 attributes $\mathbf{o}_0$ and its \liqe{} score, expressed as a percentile $p_{\mathrm{start}}$ of the \liqe{} scores
of the training photographs. The dial uses the same scale: $\tau{=}0.99$ means ``as good, by \liqe, as the corpus's
99th-percentile photograph''. So \qalign{} shaped the sample and \liqe{} defines the dial.

\subsection{ATune}
\label{sec:method-target}

ATune answers one question: \emph{given what this photograph shows and where its attributes sit, how should they move
for its quality to rise by a given amount?} Its inputs are the content of the photograph, as a frozen \dinov ViT-L/16 embedding $\mathbf{e}$~\cite{simeoni2025dinov3}; its 13 current attributes
$\mathbf{o}_0$; its current quality percentile $p_{\mathrm{start}}$; and the requested quality rise, the \emph{lift}
$\delta$. Its output is the \emph{change} $\boldsymbol{\Delta}$ each attribute needs, in buckets. During training the
input is a degraded copy, the target is its original, and the lift is the gap between them,
$\delta=\max(p_{\mathrm{orig}}-p_{\mathrm{start}},0)$, with $p_{\mathrm{orig}}$ the original's percentile.

Many different edits can raise a photograph's quality, so the model predicts a distribution over changes rather than
one change:
\begin{equation}
p_\theta\!\left(\boldsymbol{\Delta} \,\middle|\, \mathbf{e},\,\mathbf{o}_0,\,p_{\mathrm{start}},\,\delta\right)
=\sum_{k=1}^{5}\pi_k\,\mathcal{N}\!\left(\boldsymbol{\Delta}/10;\,\boldsymbol{\mu}_k,\boldsymbol{\Sigma}_k\right),
\label{eq:target}
\end{equation}
The density is a mixture of five Gaussians~\cite{bishop1994mdn}, each with a full covariance over the 13 changes.
Training maximises the likelihood of the true change under this density. The full covariance matters because attributes move
together: raising exposure deepens highlight clipping, and removing haze raises contrast.

Two design choices prevent degenerate solutions. First, the model predicts the change rather than the target. A model asked for the target
with $\mathbf{o}_0$ in view learns to return $\mathbf{o}_0$, because on most training rows the original is already close
to it (supplementary material). Second, one training row in five is an original with $\delta{=}0$ and
$\boldsymbol{\Delta}{=}0$, which teaches the model to leave a photograph alone when no lift is requested.

The network is small: a residual MLP over the \dinov embedding, three blocks of width 2048 and \numPaperParams{}
parameters in all, with the other inputs, $\mathbf{o}_0$, $p_{\mathrm{start}}$ and $\delta$, fed into each block through FiLM
conditioning~\cite{perez2018film}. We train it with
batches of 32{,}768 rows and keep the checkpoint with the lowest negative log-likelihood on a validation set of copies. The
architecture diagram and the training details are in the supplementary material.

\paragraph{Inference and the dial.} For a new photograph we measure $\mathbf{o}_0$ and $p_{\mathrm{start}}$ and ask
for the lift to a target percentile $\tau$: $\delta=\max(\tau-p_{\mathrm{start}},0)$. The dial $\tau$ is the only setting a
user chooses. A photograph already above $\tau$ gets $\delta{=}0$ and is left unchanged. Each attribute's target is
$\mathbf{o}_0$ plus the median of its predicted change, and an attribute whose predicted change is within three
buckets is not changed at all.

\subsection{Acting through an out-of-the-box editor}
\label{sec:editing}

ATune specifies target values for the attributes; reaching them is left to the editor. A photographer would do this
by adjusting a slider, inspecting the result and adjusting again. We automate the same loop as a numerical search over
the editor's own sliders. The search has no learned parameters and consults no quality judge: at each step it renders
the photograph, measures the 13 attributes and compares them with the targets.

\paragraph{Controls.} For each attribute we pick the editor tools a photographer would use to change it. These tools
use very different units, such as exposure in stops and a sharpening radius in pixels. So the search does not set them
directly. Each attribute gets a \emph{control}, a number in $[-1,1]$, and a fixed mapping turns it into tool values.
The search then works on one common scale. In RawTherapee 5.8:
\begin{itemize}
  \item for brightness, contrast and colourfulness, we use the brightness, contrast and saturation sliders;
  \item for temperature and tint, the colour curves;
  \item for shadow depth and highlight level, the shadows and highlights tools;
  \item for shadow and highlight clipping, the black and white points;
  \item for haze and local contrast, the dehaze and local-contrast tools;
  \item for sharpness, sharpening, and for noise, noise reduction.
\end{itemize}
We write the control values into RawTherapee's settings file and render from the command line. Exact settings, and darktable's controls, are in the supplementary material.

\paragraph{Search.} The search sets all controls at once, with one goal per attribute. Attributes the model wants
to change aim for the values it predicted; the rest aim to stay where they are, so the edit leaves them alone. After each render we measure the attributes and compute how far each is from its goal, in buckets. A bounded
Levenberg--Marquardt solver~\cite{levenberg1944,marquardt1963}, with a Jacobian estimated by finite differences and
updated by Broyden's method~\cite{broyden1965}, minimises these distances, with a penalty on any attribute that ends up
further from its goal than it started (exact objective in the supplementary material). The search runs on the
\numSolverFrame\,px frame with at most \numSolverMaxRenders{} renders, then renders at native resolution with pixel radii scaled up, re-measures, and
corrects at most \numSolverCorr. One rule handles a side effect: sharpening raises measured noise, and the denoiser lowers measured
sharpness, so when the model asks for more sharpness the hold on noise is released; otherwise the penalty would block
the sharpening.

The same model and the same search also drive a second editor, darktable~\cite{darktable}, through its own modules,
with a 14th control for requests for more noise (supplementary material).

\paragraph{Calibrating the controls.} The same slider behaves differently in different editors: RawTherapee's sharpening and darktable's
differ in radius, in strength and in how much one step changes the image. For the search to work, each
control must move its attribute by a predictable amount. So, once per editor, we calibrate each control's fixed
settings: the tool it drives, that tool's constants (such as a radius), and how a control value maps to a tool value.
The calibration is a keep-if-better loop following autoresearch~\cite{karpathy2026autoresearch}: an LLM coding agent
proposes one change to the settings at a time, and a change is kept only if the search then reaches more of its goals, where a goal counts as reached when the attribute ends within two buckets of its target. On held-out
photographs the calibrated search reaches \numCalReachRT\% of the model's goals on RawTherapee and
\numCalReachDt\% on darktable (details in the supplementary material).

\section{Experiments}
\label{sec:experiments}

\subsection{Protocol}
\label{sec:setup}

\paragraph{Test sets.} We evaluate on three public sets of different kinds. \textbf{FiveK}~\cite{bychkovsky2011fivek}
is the standard retouching benchmark: we use its test split of 499 photographs, with expert C as the reference edit. \textbf{PPR10K}~\cite{liang2021ppr10k} is a portrait
set: we use 382 portraits from its official validation groups, with the expert's retouched version released by
MMArt~\cite{lin2025jarvisart} as the reference. \textbf{SPAQ}~\cite{fang2020spaq} is a set of smartphone
photographs with no reference edit: we use a fixed list of 1000 of its 11{,}125 photographs.
ATune trains on none of them.

\paragraph{Baselines.} We compare with seven published retouching models and with two frontier vision--language
models that set the sliders of the same editor (\cref{sec:frontier}). CSRNet, AdaInt and NeurOP are supervised on
FiveK's expert edits; AdaInt and NeurOP also come in PPR10K-trained versions, which we use on PPR10K. UEGAN learns
from FiveK's expert edits without pairing them. EnlightenGAN, Zero-DCE and SCI are unpaired low-light enhancers.
CSRNet's authors split FiveK at random, and its released weights trained on \numCsrSeen{} of the 499 test
photographs.

\paragraph{Judges.} A retouch has no single right answer, so we score each render with nine no-reference quality
scores from eight published models: the language-model scorers \qalign~\cite{wu2024qalign} (quality and aesthetics
scores) and DeQA-Score~\cite{you2025deqa}; the CLIP-based \liqe~\cite{zhang2023liqe}, CLIP-IQA~\cite{wang2023clipiqa},
QualiCLIP~\cite{agnolucci2024qualiclip} and LAION aesthetic predictor~\cite{schuhmann2022laion}; and the convolutional
PaQ-2-PiQ~\cite{ying2020paq2piq} and TOPIQ~\cite{chen2024topiq}. Three of the nine are tied to our training: ATune is
conditioned on \liqe{} percentiles, and the corpus was sampled by \qalign, which gives two of the scores. The other six
are \emph{held out}. \qalign{} and DeQA-Score also trained on SPAQ photographs. Judges are imperfect proxies for human judgement: on FiveK renderings, \liqe{} and CLIP-IQA agree with
human pairwise preferences 73\% and 68\% of the time, against 90\% between two raters~\cite{plohotnuk2025dear}. We
therefore report each judge separately.

\paragraph{Statistics.} For each judge we report the mean change in score against the unedited original, with a
95\% bootstrap confidence interval~\cite{efron1994bootstrap} ($10^4$ resamples; PPR10K groups several shots of the same person into a session, so there we
resample sessions rather than photographs). To compare two methods we pair their edits per photograph, test each judge
with Holm's correction~\cite{holm1979} across the nine, and report how many judges prefer ours, show no significant
difference, or prefer the other (ours\,/\,n.s.\,/\,theirs). A count of $k$ means that, with 95\% confidence, at least $k$ judges prefer that side~\cite{dror2017replicability}.

\begin{table*}[t]
\centering
\caption{\textbf{Every judge against the unedited original:} mean per-photograph score change; columns marked $^{\times}$ are scaled by 100 because those judges score on a 0--1
scale.
\textbf{Bold}/\underline{underlined}: best/second; \nsig{grey}: 95\% CI includes 0; \bad{red}: significantly negative. $^\dagger$PPR10K-trained version. $^\ddagger$Saw \numCsrSeen{} of the \numCsrN{} FiveK test photographs in training.}
\label{tab:main}
\setlength{\tabcolsep}{4.2pt}\footnotesize\renewcommand{\arraystretch}{1.02}
\begin{adjustbox}{max width=\textwidth}
\begin{tabular}{@{}lccccccccc@{}}
\toprule
 & \multicolumn{3}{c}{Used in training} & \multicolumn{6}{c}{Held out} \\
\cmidrule(lr){2-4}\cmidrule(l){5-10}
Method & Q-Align & Q-Align\textsubscript{A} & LIQE & PaQ-2-PiQ & TOPIQ$^{\times}$ & LAION & CLIP-IQA$^{\times}$ & QualiCLIP$^{\times}$ & DeQA \\
\midrule
\rowcolor{yellow!40}\multicolumn{10}{l}{\textit{FiveK} ($n{=}499$)} \\
\textbf{ATune} (10M, $\tau{=}0.99$) & \textbf{+0.207} & \textbf{+0.225} & \textbf{+0.71} & +2.03 & +1.99 & \underline{+0.210} & \textbf{+10.48} & \textbf{+4.36} & +0.102 \\
\hdashline
Claude Opus 5, same editor & \underline{+0.200} & \underline{+0.139} & \underline{+0.49} & \underline{+2.27} & \textbf{+2.25} & \textbf{+0.211} & +1.03 & \underline{+2.59} & +0.102 \\
GPT-6 Astra, same editor & +0.145 & +0.085 & +0.35 & +1.55 & +1.54 & +0.147 & \bad{$-$1.49} & +2.08 & +0.083 \\
\hdashline
CSRNet~\cite{he2020csrnet}$^\ddagger$ & \bad{$-$0.138} & \nsig{$-$0.010} & \bad{$-$0.35} & \textbf{+2.70} & \underline{+2.14} & +0.086 & \nsig{$-$0.25} & \bad{$-$5.19} & \bad{$-$0.167} \\
AdaInt~\cite{yang2022adaint}$^\dagger$ & +0.134 & +0.113 & +0.25 & +2.03 & +1.82 & +0.117 & +2.56 & +0.92 & \textbf{+0.127} \\
NeurOP~\cite{wang2022neurop}$^\dagger$ & +0.137 & +0.105 & +0.24 & +1.81 & +1.66 & +0.113 & +2.35 & +1.05 & \underline{+0.122} \\
\hdashline
UEGAN~\cite{ni2020uegan} & +0.130 & +0.106 & +0.25 & +1.66 & +1.48 & +0.109 & \underline{+3.16} & +0.97 & +0.114 \\
EnlightenGAN~\cite{jiang2021enlightengan} & \bad{$-$0.337} & \bad{$-$0.122} & \bad{$-$0.43} & +2.03 & +1.96 & +0.090 & \bad{$-$4.69} & \bad{$-$2.74} & \bad{$-$0.284} \\
Zero-DCE~\cite{guo2020zerodce} & \bad{$-$0.311} & \bad{$-$0.219} & \bad{$-$0.08} & +2.13 & +1.89 & \bad{$-$0.022} & \bad{$-$1.50} & \bad{$-$0.99} & \bad{$-$0.157} \\
SCI~\cite{ma2022sci} & \bad{$-$1.207} & \bad{$-$0.484} & \bad{$-$0.85} & \bad{$-$2.34} & \bad{$-$1.73} & \bad{$-$0.279} & \bad{$-$10.58} & \bad{$-$10.67} & \bad{$-$1.209} \\
\midrule
\rowcolor{yellow!40}\multicolumn{10}{l}{\textit{PPR10K} ($n{=}382$)} \\
\textbf{ATune} (10M, $\tau{=}0.99$) & \textbf{+0.155} & \textbf{+0.148} & \textbf{+0.35} & +1.22 & \textbf{+1.59} & \textbf{+0.124} & \textbf{+10.00} & \textbf{+2.68} & +0.072 \\
\hdashline
Claude Opus 5, same editor & \underline{+0.130} & \textbf{+0.148} & \textbf{+0.35} & +1.42 & \underline{+1.46} & \underline{+0.061} & \underline{+7.74} & \underline{+0.79} & +0.045 \\
GPT-6 Astra, same editor & +0.110 & \underline{+0.080} & \underline{+0.27} & +0.94 & +1.08 & +0.031 & +3.44 & \nsig{+0.35} & +0.085 \\
\hdashline
CSRNet~\cite{he2020csrnet}$^\ddagger$ & \bad{$-$0.508} & \bad{$-$0.303} & \bad{$-$0.65} & \textbf{+2.02} & +0.50 & \bad{$-$0.206} & \bad{$-$1.65} & \bad{$-$9.22} & \bad{$-$0.540} \\
AdaInt~\cite{yang2022adaint}$^\dagger$ & +0.111 & +0.020 & +0.16 & +0.88 & +1.06 & \nsig{+0.008} & +0.97 & \nsig{+0.28} & \textbf{+0.150} \\
NeurOP~\cite{wang2022neurop}$^\dagger$ & +0.074 & \nsig{+0.016} & +0.16 & +0.57 & +0.85 & \nsig{0.000} & \nsig{$-$0.33} & \nsig{$-$0.06} & \underline{+0.118} \\
\hdashline
UEGAN~\cite{ni2020uegan} & +0.032 & +0.030 & +0.03 & +1.09 & +0.78 & \bad{$-$0.028} & +3.43 & \bad{$-$0.53} & \nsig{+0.016} \\
EnlightenGAN~\cite{jiang2021enlightengan} & \bad{$-$0.279} & \bad{$-$0.122} & \bad{$-$0.32} & \underline{+1.50} & +1.23 & \nsig{$-$0.020} & \bad{$-$3.67} & \bad{$-$1.43} & \bad{$-$0.341} \\
Zero-DCE~\cite{guo2020zerodce} & \bad{$-$0.257} & \bad{$-$0.246} & \nsig{+0.01} & +1.44 & +1.34 & \bad{$-$0.102} & \nsig{+0.27} & \bad{$-$1.02} & \bad{$-$0.183} \\
SCI~\cite{ma2022sci} & \bad{$-$1.672} & \bad{$-$0.686} & \bad{$-$0.96} & \bad{$-$2.68} & \bad{$-$3.30} & \bad{$-$0.360} & \bad{$-$9.74} & \bad{$-$19.43} & \bad{$-$1.785} \\
\midrule
\rowcolor{yellow!40}\multicolumn{10}{l}{\textit{SPAQ} ($n{=}1000$)} \\
\textbf{ATune} (10M, $\tau{=}0.99$) & \textbf{+0.125} & \textbf{+0.153} & \textbf{+0.39} & \underline{+3.80} & +1.88 & +0.158 & \textbf{+5.46} & \nsig{\underline{+0.12}} & \textbf{+0.132} \\
\hdashline
Claude Opus 5, same editor & +0.064 & +0.093 & \underline{+0.22} & +3.76 & \underline{+2.05} & \textbf{+0.165} & +2.18 & \bad{$-$1.65} & +0.068 \\
GPT-6 Astra, same editor & \underline{+0.070} & +0.053 & +0.18 & +2.18 & +1.32 & +0.106 & +0.39 & \bad{$-$1.19} & \underline{+0.079} \\
\hdashline
CSRNet~\cite{he2020csrnet}$^\ddagger$ & \bad{$-$0.259} & +0.043 & \bad{$-$0.08} & \textbf{+3.87} & +1.33 & +0.116 & +2.30 & \bad{$-$6.13} & \bad{$-$0.199} \\
AdaInt~\cite{yang2022adaint}$^\dagger$ & \nsig{$-$0.011} & +0.099 & +0.16 & +2.48 & +1.24 & +0.161 & +3.79 & \nsig{+0.07} & +0.021 \\
NeurOP~\cite{wang2022neurop}$^\dagger$ & \nsig{$-$0.009} & \underline{+0.101} & +0.20 & +2.58 & +1.23 & \underline{+0.164} & \underline{+4.07} & \nsig{$-$0.06} & +0.018 \\
\hdashline
UEGAN~\cite{ni2020uegan} & \bad{$-$0.020} & +0.070 & +0.16 & +1.96 & +0.92 & +0.146 & +3.59 & \textbf{+1.34} & \nsig{+0.002} \\
EnlightenGAN~\cite{jiang2021enlightengan} & \bad{$-$0.417} & \bad{$-$0.066} & \bad{$-$0.28} & +2.88 & +1.25 & +0.055 & \nsig{$-$0.03} & \bad{$-$3.37} & \bad{$-$0.363} \\
Zero-DCE~\cite{guo2020zerodce} & \bad{$-$0.337} & \bad{$-$0.136} & \bad{$-$0.10} & +3.18 & \textbf{+2.14} & \nsig{$-$0.009} & +1.38 & \bad{$-$1.25} & \bad{$-$0.220} \\
SCI~\cite{ma2022sci} & \bad{$-$1.133} & \bad{$-$0.273} & \bad{$-$0.53} & \bad{$-$1.13} & \bad{$-$1.78} & \bad{$-$0.179} & \bad{$-$2.32} & \bad{$-$14.44} & \bad{$-$1.029} \\
\bottomrule
\end{tabular}

\end{adjustbox}
\end{table*}

\subsection{Main result}
\label{sec:results}

\begin{table}[t]
\centering
\caption{\textbf{Paired comparison of ATune with each method.} Number of judges that significantly prefer our edit\,/\,show
no significant difference\,/\,prefer the other method's, out of all nine judges and out of the six held out from our
training (Holm-corrected). \good{Green}: more judges prefer ours. \bad{Red}: more judges prefer the other method. $^\dagger$PPR10K-trained version. $^\ddagger$Saw \numCsrSeen{} of the \numCsrN{} FiveK test photographs in training.}
\label{tab:wtl}
\setlength{\tabcolsep}{5pt}\footnotesize\renewcommand{\arraystretch}{1.1}
\begin{adjustbox}{max width=\linewidth}
\begin{tabular}{@{}lcccccc@{}}
\toprule
 & \multicolumn{2}{c}{FiveK} & \multicolumn{2}{c}{PPR10K} & \multicolumn{2}{c}{SPAQ} \\
\cmidrule(lr){2-3}\cmidrule(lr){4-5}\cmidrule(l){6-7}
ATune vs & all 9 & held out & all 9 & held out & all 9 & held out \\
\midrule
Claude Opus 5, same editor & \good{4\,/\,3\,/\,2} & {\scriptsize 2\,/\,2\,/\,2} & \good{3\,/\,5\,/\,1} & {\scriptsize \good{3\,/\,2\,/\,1}} & \good{6\,/\,3\,/\,0} & {\scriptsize \good{3\,/\,3\,/\,0}} \\
GPT-6 Astra, same editor & \good{9\,/\,0\,/\,0} & {\scriptsize \good{6\,/\,0\,/\,0}} & \good{8\,/\,1\,/\,0} & {\scriptsize \good{5\,/\,1\,/\,0}} & \good{9\,/\,0\,/\,0} & {\scriptsize \good{6\,/\,0\,/\,0}} \\
\hdashline
CSRNet~\cite{he2020csrnet}$^\ddagger$ & \good{7\,/\,0\,/\,2} & {\scriptsize \good{4\,/\,0\,/\,2}} & \good{8\,/\,0\,/\,1} & {\scriptsize \good{5\,/\,0\,/\,1}} & \good{8\,/\,1\,/\,0} & {\scriptsize \good{5\,/\,1\,/\,0}} \\
AdaInt~\cite{yang2022adaint}$^\dagger$ & \good{7\,/\,1\,/\,1} & {\scriptsize \good{4\,/\,1\,/\,1}} & \good{8\,/\,0\,/\,1} & {\scriptsize \good{5\,/\,0\,/\,1}} & \good{7\,/\,2\,/\,0} & {\scriptsize \good{4\,/\,2\,/\,0}} \\
NeurOP~\cite{wang2022neurop}$^\dagger$ & \good{8\,/\,0\,/\,1} & {\scriptsize \good{5\,/\,0\,/\,1}} & \good{8\,/\,0\,/\,1} & {\scriptsize \good{5\,/\,0\,/\,1}} & \good{7\,/\,2\,/\,0} & {\scriptsize \good{4\,/\,2\,/\,0}} \\
UEGAN~\cite{ni2020uegan} & \good{8\,/\,1\,/\,0} & {\scriptsize \good{5\,/\,1\,/\,0}} & \good{8\,/\,1\,/\,0} & {\scriptsize \good{5\,/\,1\,/\,0}} & \good{7\,/\,1\,/\,1} & {\scriptsize \good{4\,/\,1\,/\,1}} \\
EnlightenGAN~\cite{jiang2021enlightengan} & \good{7\,/\,2\,/\,0} & {\scriptsize \good{4\,/\,2\,/\,0}} & \good{8\,/\,0\,/\,1} & {\scriptsize \good{5\,/\,0\,/\,1}} & \good{9\,/\,0\,/\,0} & {\scriptsize \good{6\,/\,0\,/\,0}} \\
Zero-DCE~\cite{guo2020zerodce} & \good{7\,/\,2\,/\,0} & {\scriptsize \good{4\,/\,2\,/\,0}} & \good{7\,/\,2\,/\,0} & {\scriptsize \good{4\,/\,2\,/\,0}} & \good{8\,/\,0\,/\,1} & {\scriptsize \good{5\,/\,0\,/\,1}} \\
SCI~\cite{ma2022sci} & \good{9\,/\,0\,/\,0} & {\scriptsize \good{6\,/\,0\,/\,0}} & \good{9\,/\,0\,/\,0} & {\scriptsize \good{6\,/\,0\,/\,0}} & \good{9\,/\,0\,/\,0} & {\scriptsize \good{6\,/\,0\,/\,0}} \\
\bottomrule
\end{tabular}

\end{adjustbox}
\end{table}

ATune improves the photographs under the judges: the mean score change against the unedited original is
significantly positive in \numOursPosCells{} of the 27 judge--set cells (9 judges $\times$ 3 test sets) and negative
in none (\cref{tab:main}). In the paired
comparison of \cref{tab:wtl}, at least \numWinMinRetouch{} of the nine judges prefer our edit to each retouching
model's on every test set, and at most \numLossMaxRetouch{} prefer theirs. This holds against AdaInt and NeurOP on
FiveK, which they were trained on (\numWTLAdaIntFiveK{} and \numWTLNeurOPFiveK). The one dissenting judge is
DeQA-Score, which prefers AdaInt and NeurOP to ours on FiveK and PPR10K.

The supervised models do not carry over to other photographs. On SPAQ, AdaInt and NeurOP leave \qalign{} quality
unchanged (\numMeanAdaIntQASPAQ{} and \numMeanNeurOPQASPAQ, against \numMeanOursQASPAQ{} for ours), whereas on FiveK
they raise it (\numMeanAdaIntQAFiveK{} and \numMeanNeurOPQAFiveK). Even on PPR10K, with their PPR10K-trained
versions, AdaInt improves the photograph under \numAdaIntUpPPR{} judges and NeurOP under \numNeurOPUpPPR, against
\numOursUpPPR{} for ours. The unpaired low-light enhancers lower \qalign{} quality on every set: they brighten
photographs that are already well exposed.

\subsection{Against frontier vision--language models}
\label{sec:frontier}

A frontier vision--language model can set an editor's sliders too, so we ask Claude Opus 5 and GPT-6 Astra to make
the same decision ATune makes. Each receives the same 1024\,px frame and one RawTherapee slider for each of the 13 attributes, with the
minimum and maximum the editor allows, plus the current white balance. The prompt adds no system message,
hints or example values (full prompt in the supplementary material). The model answers once, with no feedback, and
its slider values are written into the editor and rendered at native resolution, as ours are.

Against GPT-6 Astra, at least \numWinMinGPT{} of the nine judges prefer our edits on every test set
(\numWTLGPTFiveK, \numWTLGPTPPR{} and \numWTLGPTSPAQ; \cref{tab:wtl}), although GPT-6 Astra's own edits improve the
photograph in \numGPTPosCells{} of the 27 judge--set cells. Against Claude Opus 5 the margin is narrower:
\numWTLClaudeFiveK{} on FiveK, \numWTLClaudePPR{} on PPR10K and \numWTLClaudeSPAQ{} on SPAQ, and among the six
held-out judges \numWTLHeldClaudeFiveK, \numWTLHeldClaudePPR{} and \numWTLHeldClaudeSPAQ (examples in \cref{fig:qual}).

\textbf{A second editor.} Because ATune never sees the editor, we also ran it through darktable
3.8~\cite{darktable}, calibrated as for RawTherapee (\cref{sec:editing}), with the model, the dial and the search
unchanged. Its darktable edits improve the photograph in \numDtPosCells{} of the 27 cells and lower it in none. The
judges prefer the RawTherapee edits (\numVEditorDtFiveK, \numVEditorDtPPR{} and \numVEditorDtSPAQ;
\cref{tab:ablation}c), which remain our main result. Given darktable's own sliders, Claude Opus 5 and GPT-6 Astra
fall behind ATune there too: at least \numDtWinMinFrontier{} judges prefer our edits on every test set, and at most
\numDtLossMaxFrontier{} prefer theirs (supplementary material).

\subsection{Human raters}
\label{sec:human}

\begin{table}[t]
\centering
\caption{\textbf{Human preference for ATune's edits.} Percentage of pairwise votes in favour of our edit against each
method (95\% confidence interval in parentheses). Pooled results weight the three test sets equally. \good{Green}: significantly
above 50\% (Holm-corrected). $^\dagger$NeurOP on FiveK,
AdaInt on PPR10K.}
\label{tab:human}
\setlength{\tabcolsep}{4pt}\footnotesize\renewcommand{\arraystretch}{1.05}
\begin{adjustbox}{max width=\linewidth}
\begin{tabular}{@{}lcccc@{}}
\toprule
ATune vs & FiveK & PPR10K & SPAQ & Pooled \\
\midrule
Claude Opus 5 & 44.2 {\scriptsize(38.4--49.8)} & 56.1 {\scriptsize(49.3--62.8)} & 51.2 {\scriptsize(45.3--57.1)} & 50.5 {\scriptsize(46.8--54.2)} \\
GPT-6 Astra & 54.0 {\scriptsize(48.1--59.7)} & 54.1 {\scriptsize(47.0--61.3)} & 55.0 {\scriptsize(48.9--60.7)} & \good{54.3} {\scriptsize(50.3--58.3)} \\
NeurOP\,/\,AdaInt$^\dagger$ & 47.2 {\scriptsize(41.5--53.0)} & 45.3 {\scriptsize(38.3--52.6)} & -- & 46.3 {\scriptsize(41.5--51.2)} \\
\bottomrule
\end{tabular}

\end{adjustbox}
\end{table}

We ran a pairwise preference study on a crowdsourcing platform. Each rater saw our edit and another method's edit
of the same photograph side by side, in random order, and answered ``Which image is better?'' with Image~1, Image~2 or About the same. The photographs
are random samples of the three test sets. \numHumRaters{} raters gave \numHumVotes{} answers on \numHumPairs{}
pairs, at least five per pair; \numHumTie\% of answers were ties and are excluded. \Cref{tab:human} reports the
share of the remaining votes for our edit, with confidence intervals from a bootstrap over raters and
photographs~\cite{owen2007pigeonhole}.

Three results stand out (\cref{tab:human}). \textbf{First}, raters prefer our edits to GPT-6 Astra's: \numHumGPTAll\% of
votes, significant after Holm's correction ($\numHumGPTp$). \textbf{Second}, our edits and Claude Opus 5's are rated as equal:
\numHumClaudeAll\% of votes, with a 95\% CI of \numHumClaudeAllCI. Per set, Claude led on FiveK and ours led on PPR10K, by a difference of
\numHumClaudeGap{} points (95\% CI \numHumClaudeGapCI). \textbf{Third}, against NeurOP and AdaInt on their own test sets,
raters gave our edits \numHumSupAll\% of votes, even though \numWinMinSupOwn{} of the nine judges prefer ours there.

How often does a judge pick the same edit as the raters? \qalign's quality score matches the raters' majority on
\numHumAccQA\% of pairs, the most of any judge; PaQ-2-PiQ matches on \numHumAccPaq\%, the least. For comparison, a single rater
matches the majority of the other raters on the same pair \numHumAccRater\% of the time, so raters themselves disagree
on the rest.

\begin{figure*}[t]
\centering
\setlength{\tabcolsep}{1.5pt}
\begin{tabular}{@{}m{0.035\linewidth}m{0.186\linewidth}m{0.186\linewidth}m{0.186\linewidth}m{0.186\linewidth}m{0.186\linewidth}@{}}
 & \centering\small Original & \centering\small ATune & \centering\small Claude Opus 5 & \centering\small GPT-6 Astra & \centering\small Expert \tabularnewline[2pt]
\rotatebox{90}{\parbox{2.45cm}{\centering\scriptsize Prefer ours\\(90th pct.)}} & \centering \adjustbox{cframe=black!25 0.3pt}{\includegraphics[width=\dimexpr\linewidth-0.6pt\relax,height=2.45cm,keepaspectratio]{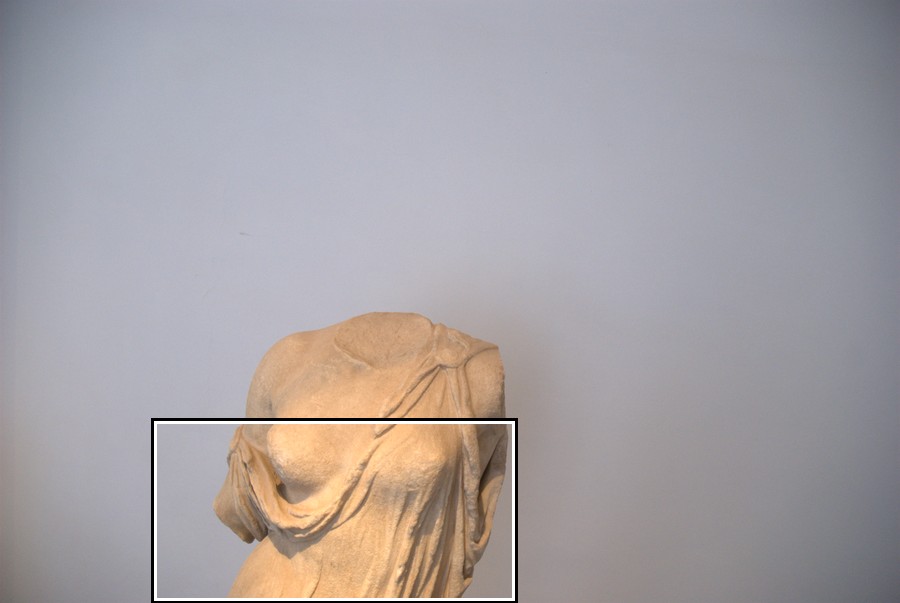}}\\[1.5pt]\adjustbox{cframe=black!25 0.3pt}{\includegraphics[width=\dimexpr\linewidth-0.6pt\relax]{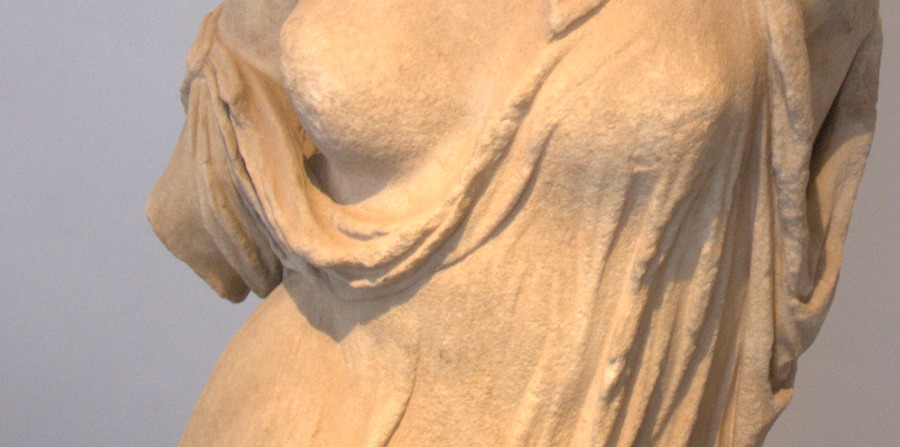}} & \centering \begin{tikzpicture}[baseline=(im.center)]\node[inner sep=0] (im) {\adjustbox{cframe=black!25 0.3pt}{\includegraphics[width=\dimexpr\linewidth-0.6pt\relax,height=2.45cm,keepaspectratio]{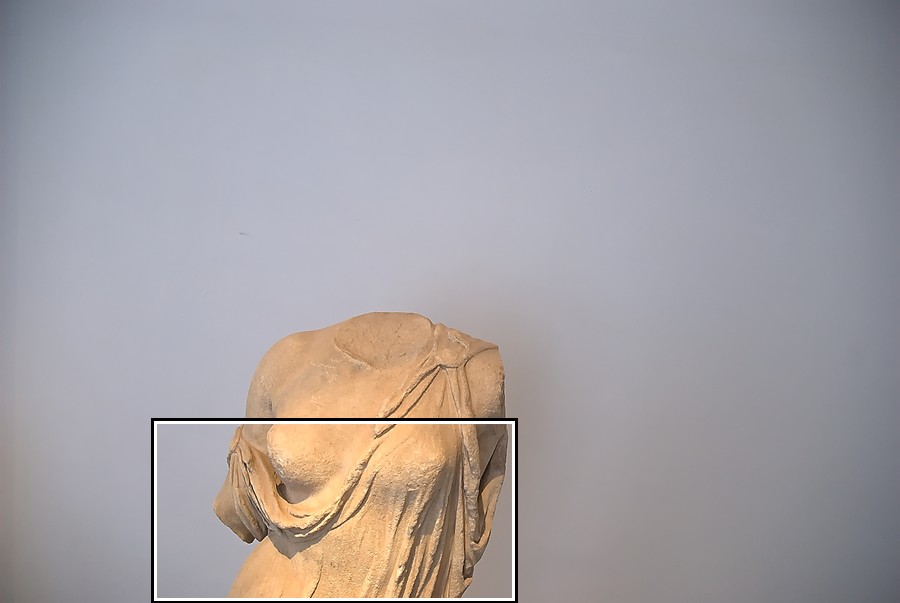}}};\node[anchor=south east, inner sep=1.2pt, fill=posgreen, text=white, font=\scriptsize\bfseries, rounded corners=0.8pt] at ([xshift=-2pt, yshift=2pt]im.south east) {8/9 judges};\end{tikzpicture}\\[1.5pt]\adjustbox{cframe=black!25 0.3pt}{\includegraphics[width=\dimexpr\linewidth-0.6pt\relax]{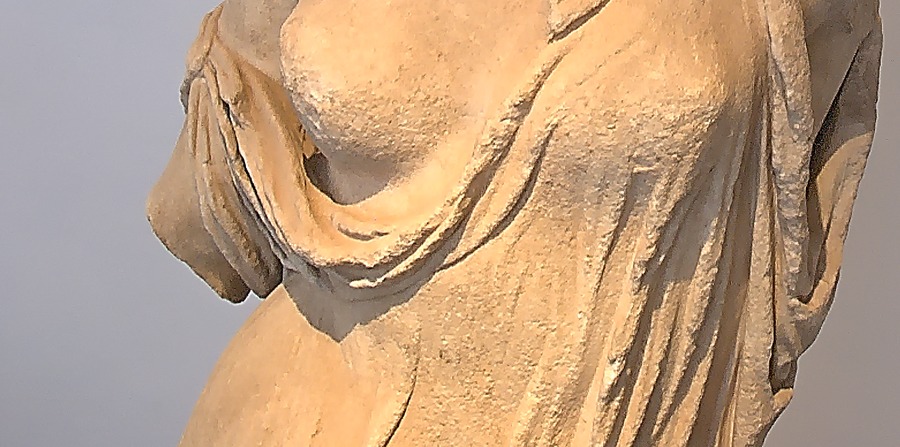}} & \centering \begin{tikzpicture}[baseline=(im.center)]\node[inner sep=0] (im) {\adjustbox{cframe=black!25 0.3pt}{\includegraphics[width=\dimexpr\linewidth-0.6pt\relax,height=2.45cm,keepaspectratio]{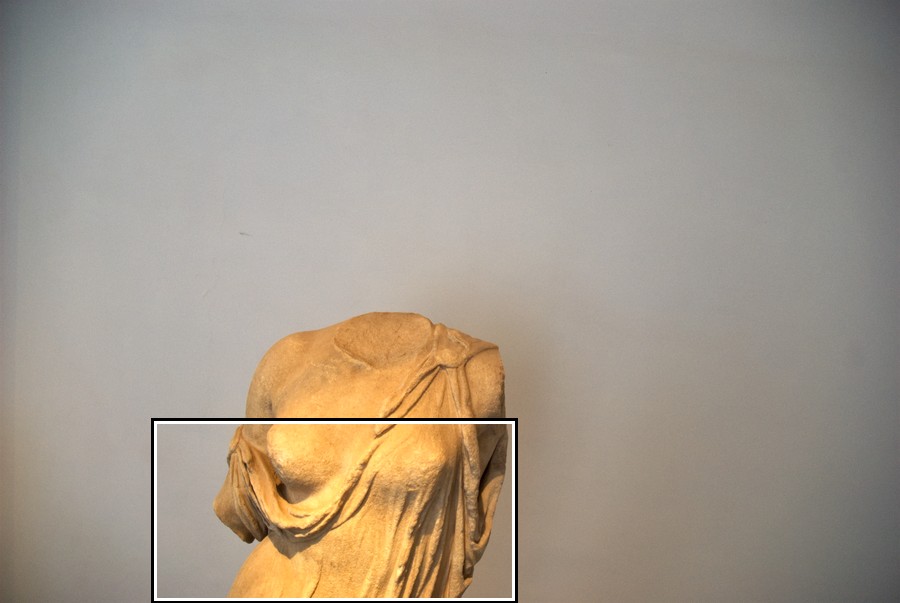}}};\node[anchor=south east, inner sep=1.2pt, fill=black!55, text=white, font=\scriptsize\bfseries, rounded corners=0.8pt] at ([xshift=-2pt, yshift=2pt]im.south east) {1/9 judges};\end{tikzpicture}\\[1.5pt]\adjustbox{cframe=black!25 0.3pt}{\includegraphics[width=\dimexpr\linewidth-0.6pt\relax]{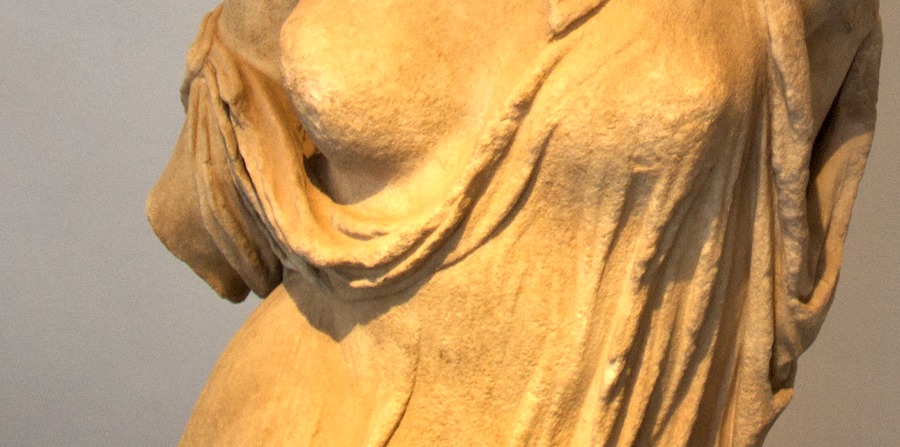}} & \centering \adjustbox{cframe=black!25 0.3pt}{\includegraphics[width=\dimexpr\linewidth-0.6pt\relax,height=2.45cm,keepaspectratio]{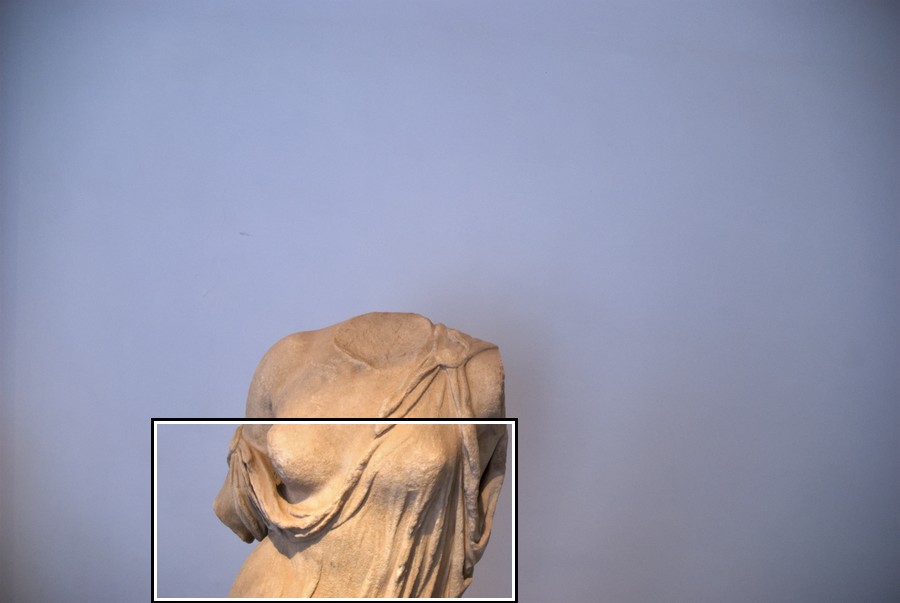}}\\[1.5pt]\adjustbox{cframe=black!25 0.3pt}{\includegraphics[width=\dimexpr\linewidth-0.6pt\relax]{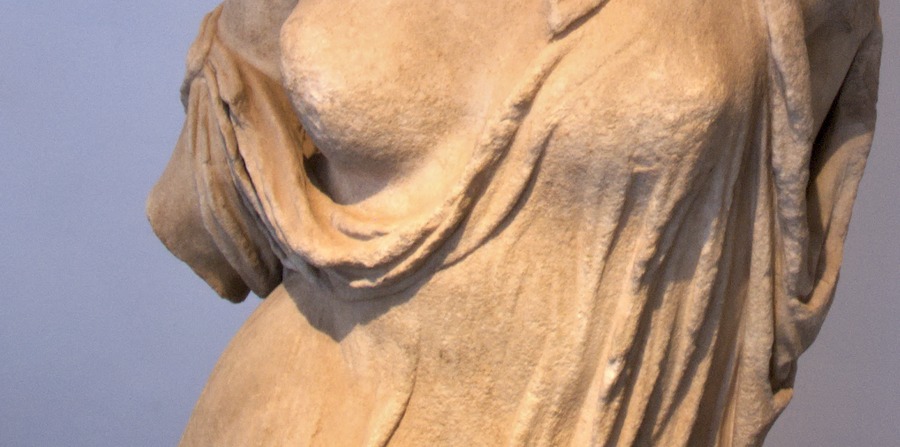}} & \centering \adjustbox{cframe=black!25 0.3pt}{\includegraphics[width=\dimexpr\linewidth-0.6pt\relax,height=2.45cm,keepaspectratio]{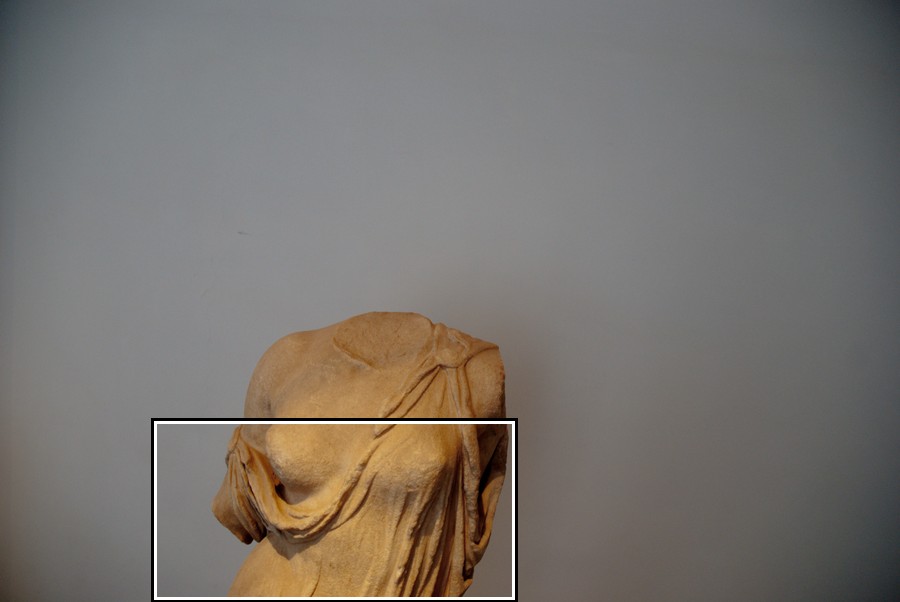}}\\[1.5pt]\adjustbox{cframe=black!25 0.3pt}{\includegraphics[width=\dimexpr\linewidth-0.6pt\relax]{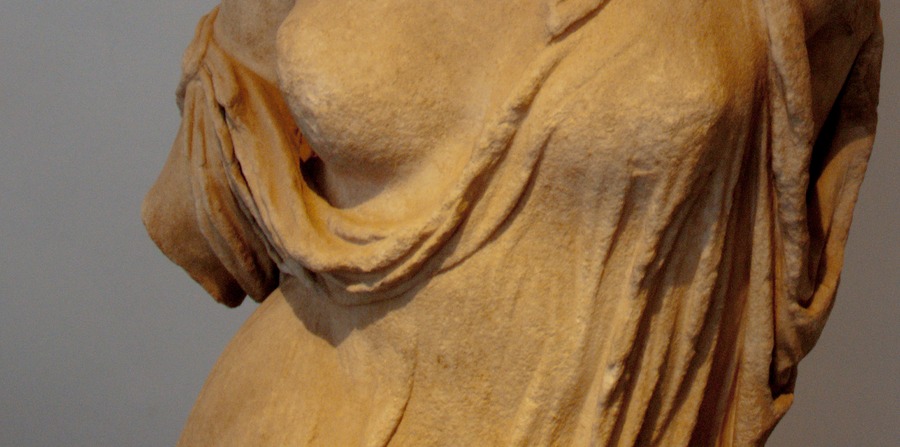}} \tabularnewline[7pt]
\rotatebox{90}{\parbox{2.45cm}{\centering\scriptsize Typical\\(median)}} & \centering \adjustbox{cframe=black!25 0.3pt}{\includegraphics[width=\dimexpr\linewidth-0.6pt\relax,height=2.45cm,keepaspectratio]{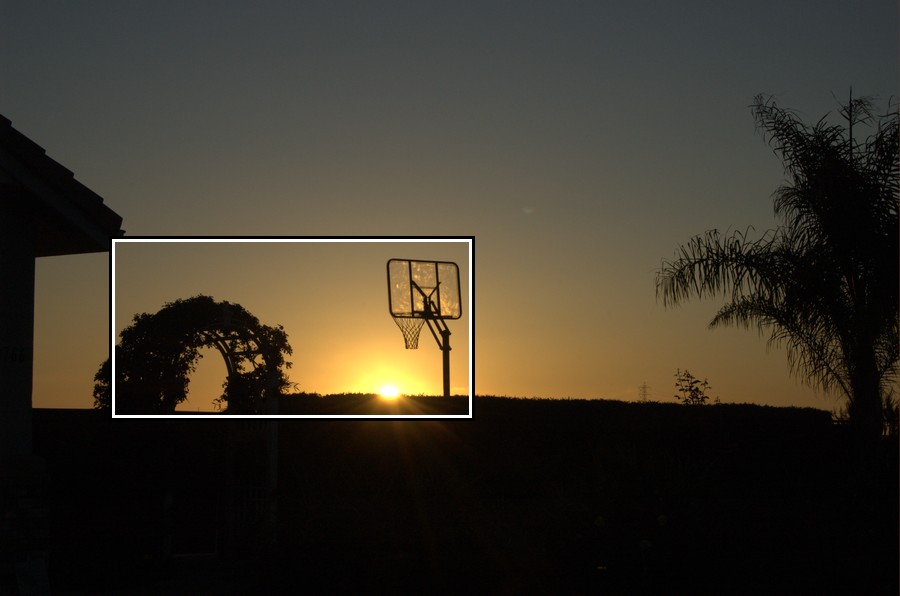}}\\[1.5pt]\adjustbox{cframe=black!25 0.3pt}{\includegraphics[width=\dimexpr\linewidth-0.6pt\relax]{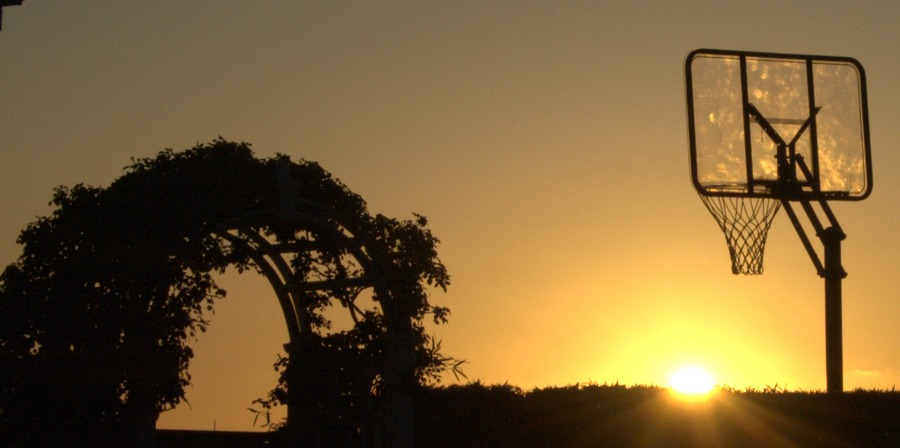}} & \centering \begin{tikzpicture}[baseline=(im.center)]\node[inner sep=0] (im) {\adjustbox{cframe=black!25 0.3pt}{\includegraphics[width=\dimexpr\linewidth-0.6pt\relax,height=2.45cm,keepaspectratio]{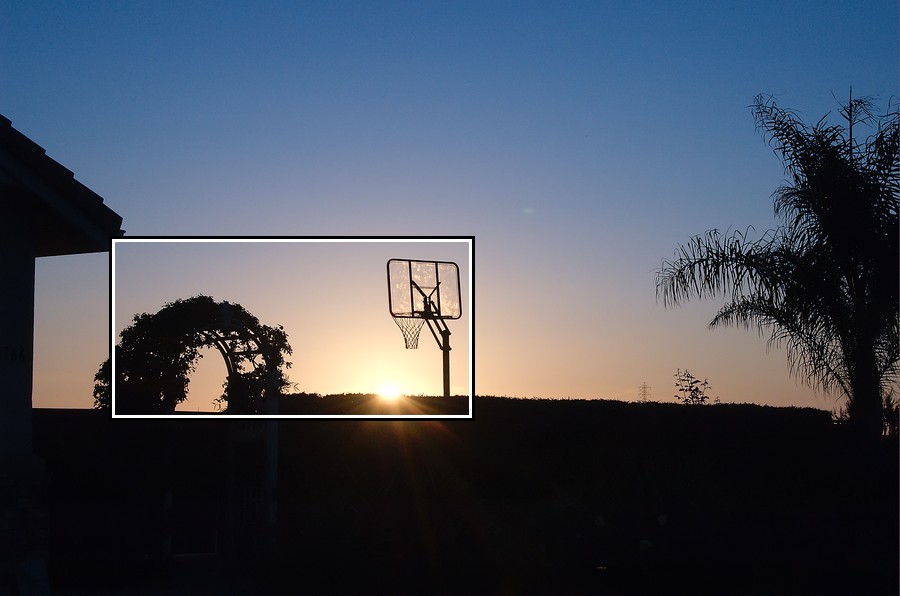}}};\node[anchor=south east, inner sep=1.2pt, fill=posgreen, text=white, font=\scriptsize\bfseries, rounded corners=0.8pt] at ([xshift=-2pt, yshift=2pt]im.south east) {7/9 judges};\end{tikzpicture}\\[1.5pt]\adjustbox{cframe=black!25 0.3pt}{\includegraphics[width=\dimexpr\linewidth-0.6pt\relax]{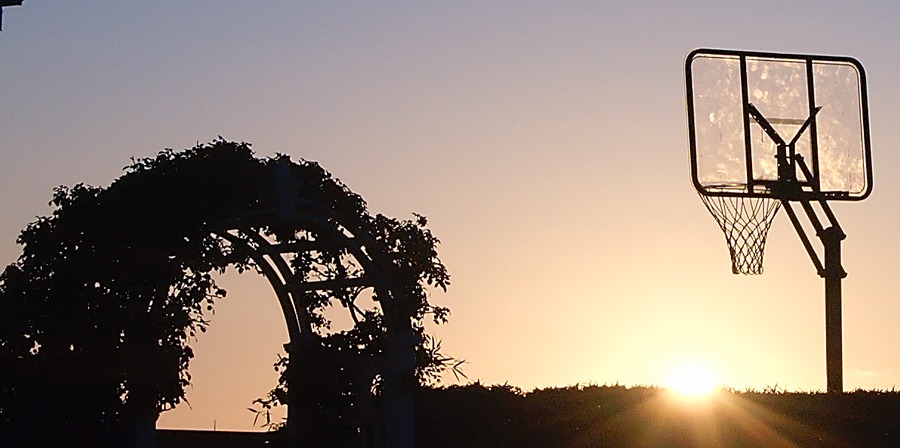}} & \centering \begin{tikzpicture}[baseline=(im.center)]\node[inner sep=0] (im) {\adjustbox{cframe=black!25 0.3pt}{\includegraphics[width=\dimexpr\linewidth-0.6pt\relax,height=2.45cm,keepaspectratio]{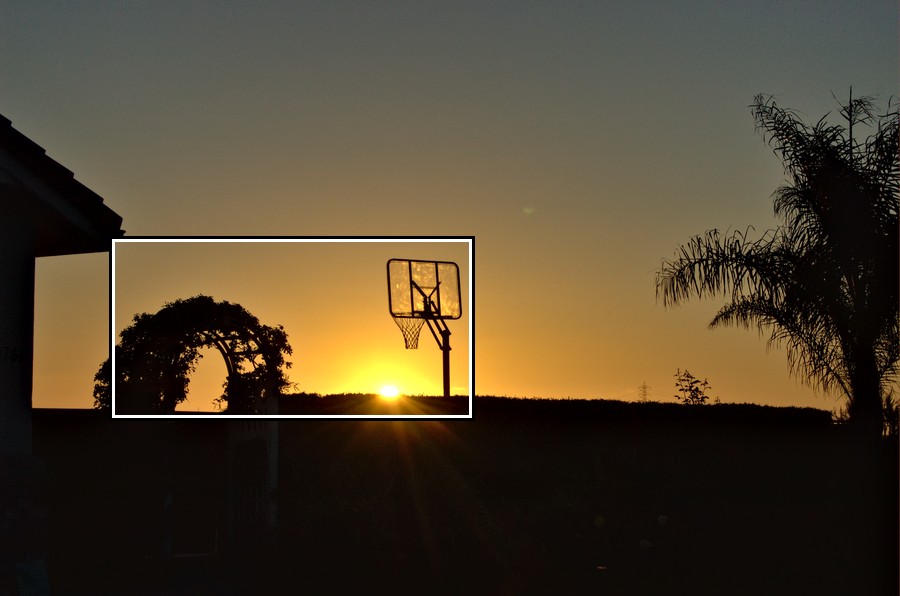}}};\node[anchor=south east, inner sep=1.2pt, fill=black!55, text=white, font=\scriptsize\bfseries, rounded corners=0.8pt] at ([xshift=-2pt, yshift=2pt]im.south east) {2/9 judges};\end{tikzpicture}\\[1.5pt]\adjustbox{cframe=black!25 0.3pt}{\includegraphics[width=\dimexpr\linewidth-0.6pt\relax]{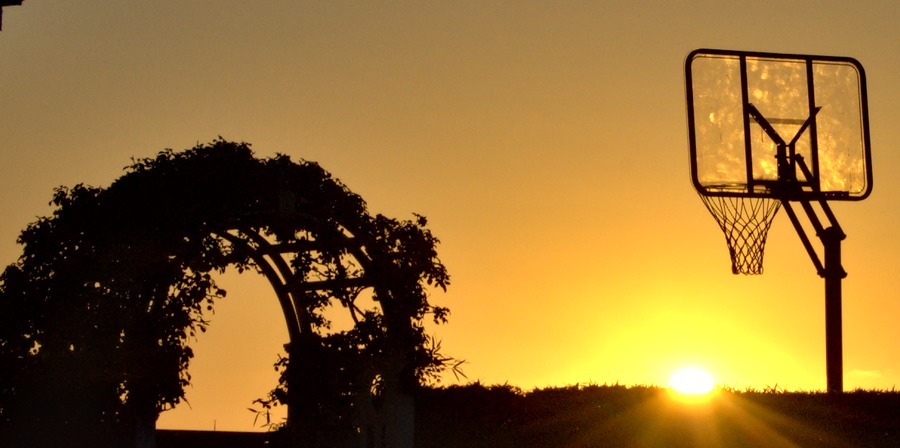}} & \centering \adjustbox{cframe=black!25 0.3pt}{\includegraphics[width=\dimexpr\linewidth-0.6pt\relax,height=2.45cm,keepaspectratio]{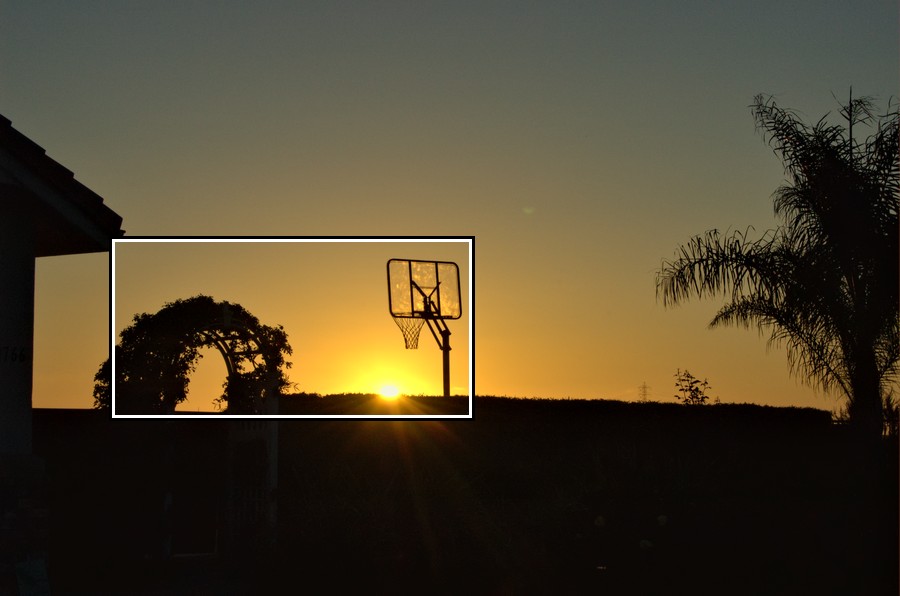}}\\[1.5pt]\adjustbox{cframe=black!25 0.3pt}{\includegraphics[width=\dimexpr\linewidth-0.6pt\relax]{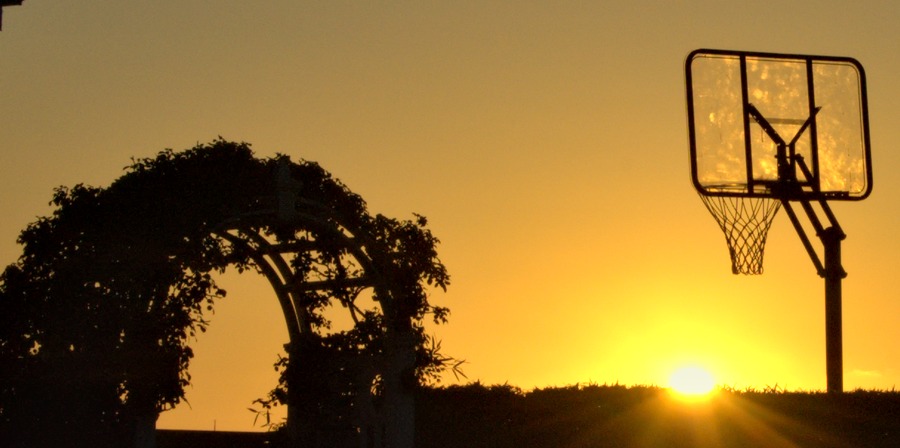}} & \centering \adjustbox{cframe=black!25 0.3pt}{\includegraphics[width=\dimexpr\linewidth-0.6pt\relax,height=2.45cm,keepaspectratio]{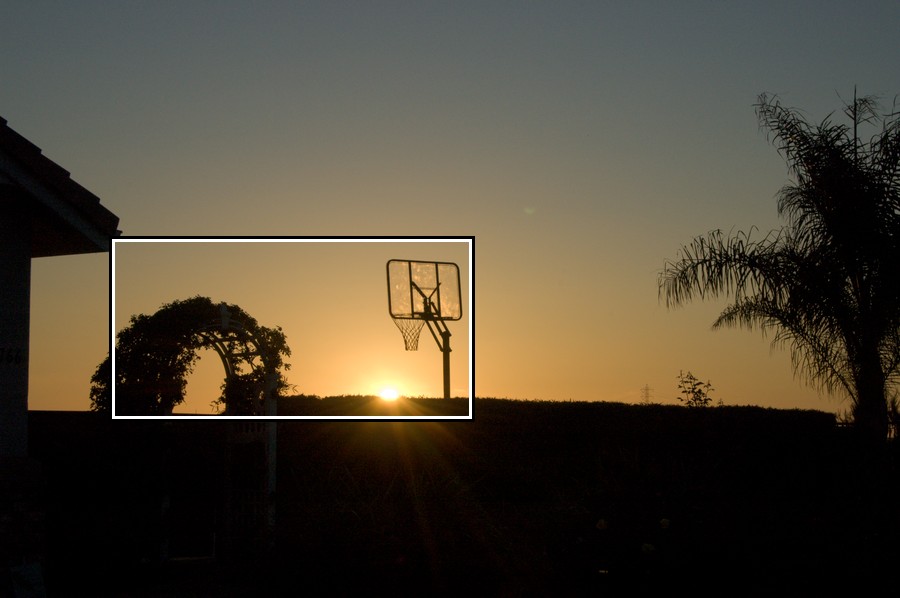}}\\[1.5pt]\adjustbox{cframe=black!25 0.3pt}{\includegraphics[width=\dimexpr\linewidth-0.6pt\relax]{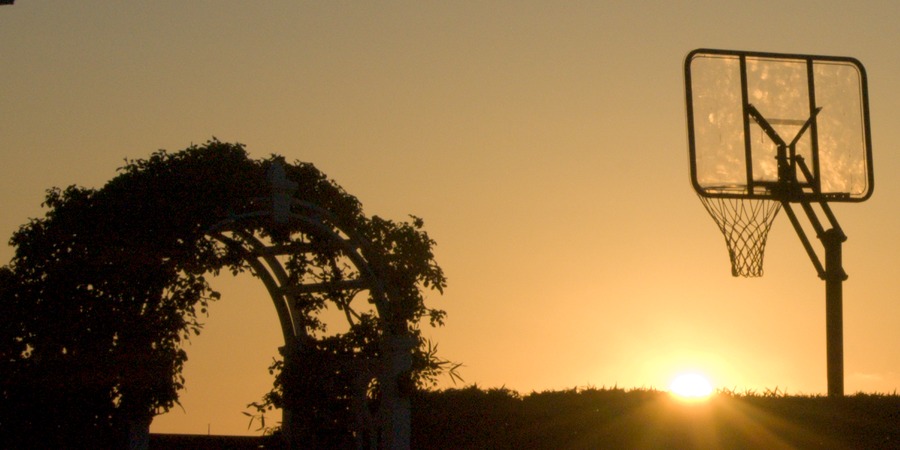}} \tabularnewline[7pt]
\rotatebox{90}{\parbox{2.45cm}{\centering\scriptsize Prefer Claude\\(10th pct.)}} & \centering \adjustbox{cframe=black!25 0.3pt}{\includegraphics[width=\dimexpr\linewidth-0.6pt\relax,height=2.45cm,keepaspectratio]{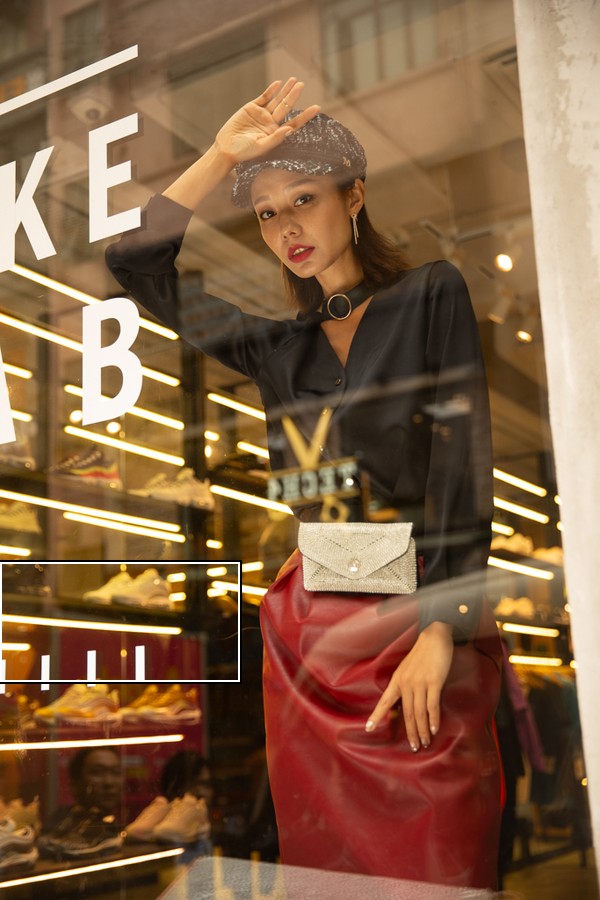}}\\[1.5pt]\adjustbox{cframe=black!25 0.3pt}{\includegraphics[width=\dimexpr\linewidth-0.6pt\relax]{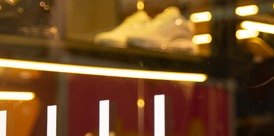}} & \centering \begin{tikzpicture}[baseline=(im.center)]\node[inner sep=0] (im) {\adjustbox{cframe=black!25 0.3pt}{\includegraphics[width=\dimexpr\linewidth-0.6pt\relax,height=2.45cm,keepaspectratio]{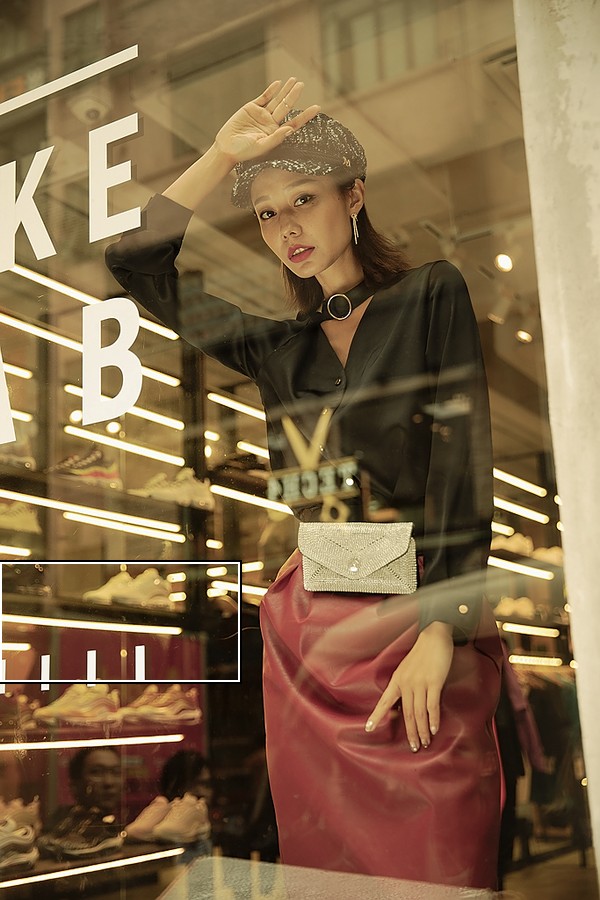}}};\node[anchor=south east, inner sep=1.2pt, fill=black!55, text=white, font=\scriptsize\bfseries, rounded corners=0.8pt] at ([xshift=-2pt, yshift=2pt]im.south east) {3/9 judges};\end{tikzpicture}\\[1.5pt]\adjustbox{cframe=black!25 0.3pt}{\includegraphics[width=\dimexpr\linewidth-0.6pt\relax]{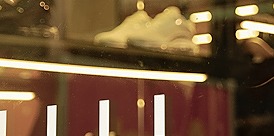}} & \centering \begin{tikzpicture}[baseline=(im.center)]\node[inner sep=0] (im) {\adjustbox{cframe=black!25 0.3pt}{\includegraphics[width=\dimexpr\linewidth-0.6pt\relax,height=2.45cm,keepaspectratio]{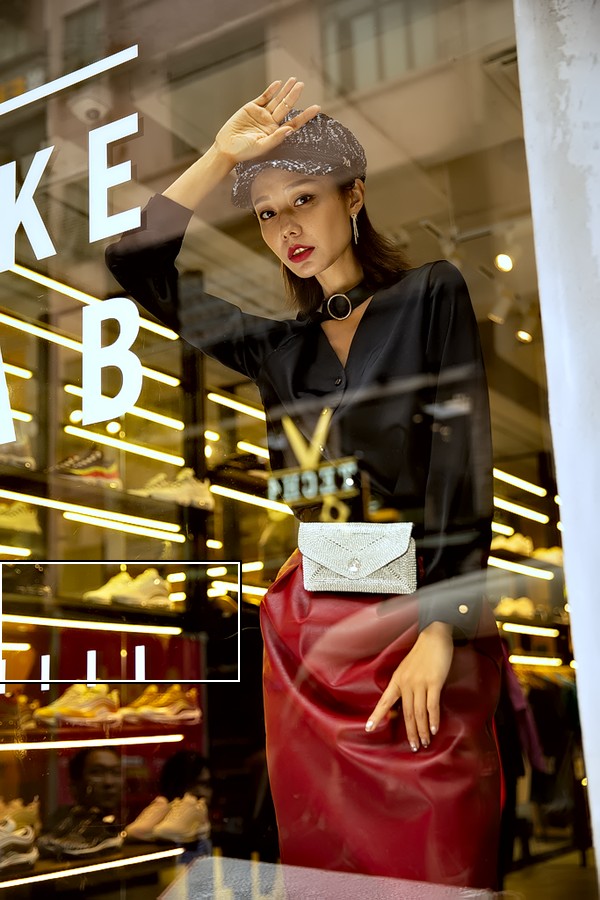}}};\node[anchor=south east, inner sep=1.2pt, fill=posgreen, text=white, font=\scriptsize\bfseries, rounded corners=0.8pt] at ([xshift=-2pt, yshift=2pt]im.south east) {6/9 judges};\end{tikzpicture}\\[1.5pt]\adjustbox{cframe=black!25 0.3pt}{\includegraphics[width=\dimexpr\linewidth-0.6pt\relax]{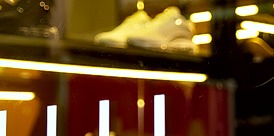}} & \centering \adjustbox{cframe=black!25 0.3pt}{\includegraphics[width=\dimexpr\linewidth-0.6pt\relax,height=2.45cm,keepaspectratio]{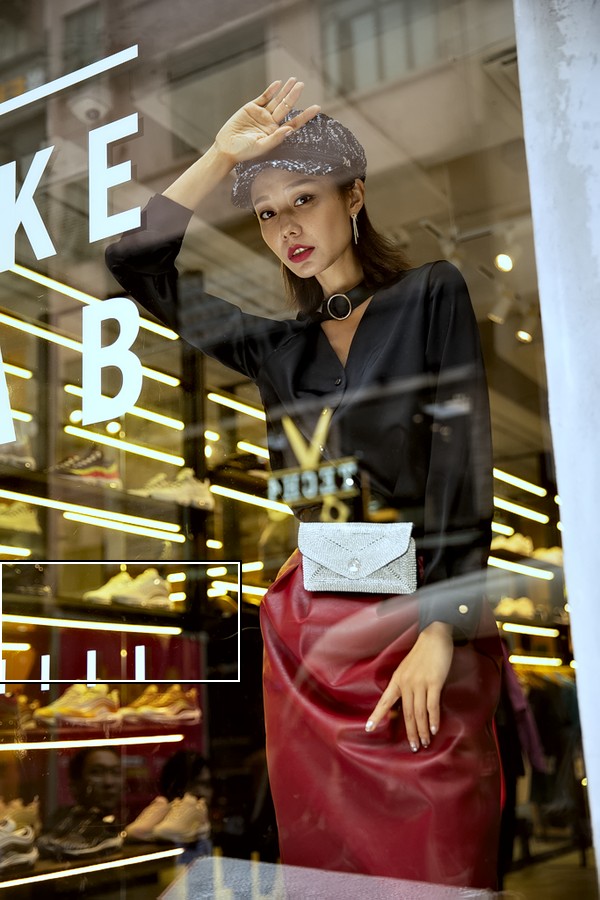}}\\[1.5pt]\adjustbox{cframe=black!25 0.3pt}{\includegraphics[width=\dimexpr\linewidth-0.6pt\relax]{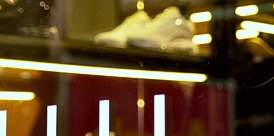}} & \centering \adjustbox{cframe=black!25 0.3pt}{\includegraphics[width=\dimexpr\linewidth-0.6pt\relax,height=2.45cm,keepaspectratio]{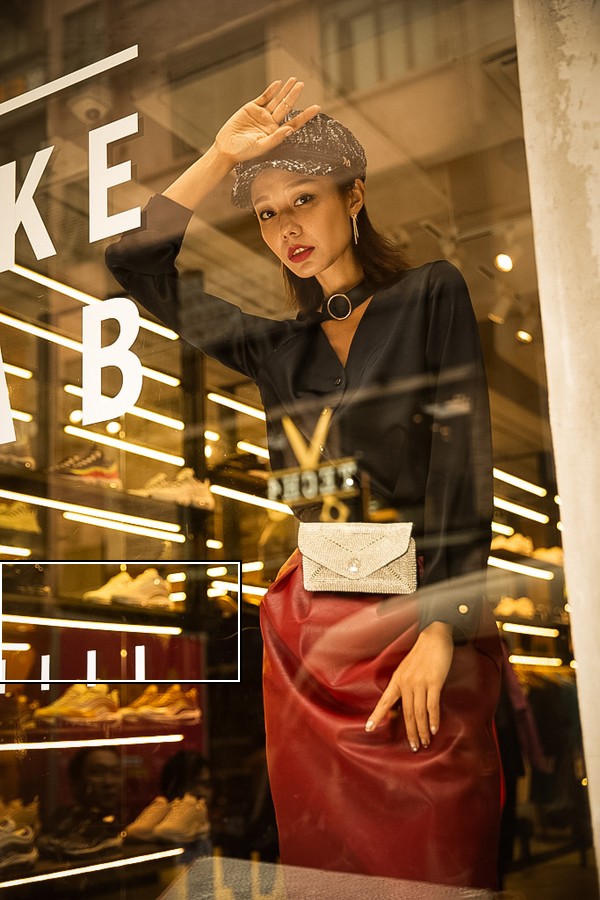}}\\[1.5pt]\adjustbox{cframe=black!25 0.3pt}{\includegraphics[width=\dimexpr\linewidth-0.6pt\relax]{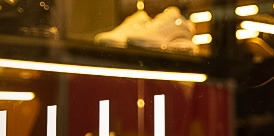}} \tabularnewline[7pt]
\end{tabular}
\caption{\textbf{Three example edits.} We score every FiveK and PPR10K photograph by how much the nine judges prefer our edit to Claude's, and show one photograph the judges clearly give to ours (top), a typical one (middle) and one they clearly give to Claude (bottom). The badge on each edit counts the judges that score it above the other. Below each image: the boxed region enlarged.}
\label{fig:qual}
\end{figure*}

\subsection{What drives the result}
\label{sec:ablations}

\begin{table}[t]
\centering
\caption{\textbf{Paired comparisons:} judges preferring the first\,/\,neither\,/\,the second (Holm over nine
judges); \good{green}/\bad{red}: more judges favour the first/second. (a) Training-set size. (b) The dial at 4M.
(c) ATune on darktable against itself on RawTherapee and against the frontier models on darktable.}
\label{tab:ablation}
\setlength{\tabcolsep}{4pt}\footnotesize\renewcommand{\arraystretch}{1.08}
\begin{adjustbox}{max width=\linewidth}
\begin{tabular}{@{}lccc@{}}
\toprule
First vs second & FiveK & PPR10K & SPAQ \\
\midrule
\multicolumn{4}{l}{\textit{(a) Corpus size}} \\
\quad ATune (10M) vs 8M & 0\,/\,9\,/\,0 & \good{2\,/\,7\,/\,0} & 0\,/\,9\,/\,0 \\
\quad 8M vs 6M & \good{7\,/\,2\,/\,0} & \good{3\,/\,6\,/\,0} & 0\,/\,9\,/\,0 \\
\quad 6M vs 4M & \good{1\,/\,8\,/\,0} & \good{2\,/\,7\,/\,0} & \good{3\,/\,6\,/\,0} \\
\quad 6M vs 2M & \good{7\,/\,2\,/\,0} & \good{5\,/\,4\,/\,0} & \good{3\,/\,6\,/\,0} \\
\quad 6M vs 1M & \good{3\,/\,6\,/\,0} & \good{8\,/\,1\,/\,0} & \good{4\,/\,5\,/\,0} \\
\hdashline
\multicolumn{4}{l}{\textit{(b) Dial $\tau$, 4M}} \\
\quad $0.99$ vs $0.97$ & \good{9\,/\,0\,/\,0} & \good{7\,/\,2\,/\,0} & \good{3\,/\,6\,/\,0} \\
\quad $0.97$ vs $0.90$ & \good{9\,/\,0\,/\,0} & \good{8\,/\,1\,/\,0} & \good{6\,/\,3\,/\,0} \\
\hdashline
\multicolumn{4}{l}{\textit{(c) Editor, 10M}} \\
\quad darktable vs RawTherapee & \bad{1\,/\,6\,/\,2} & \bad{0\,/\,4\,/\,5} & \bad{2\,/\,4\,/\,3} \\
\quad ATune vs Claude, both darktable & \good{5\,/\,2\,/\,2} & \good{7\,/\,2\,/\,0} & \good{8\,/\,1\,/\,0} \\
\quad ATune vs GPT-6, both darktable & \good{8\,/\,1\,/\,0} & \good{7\,/\,2\,/\,0} & \good{9\,/\,0\,/\,0} \\
\bottomrule
\end{tabular}

\end{adjustbox}
\end{table}

\begin{figure}[H]
\centering
\includegraphics[width=\linewidth]{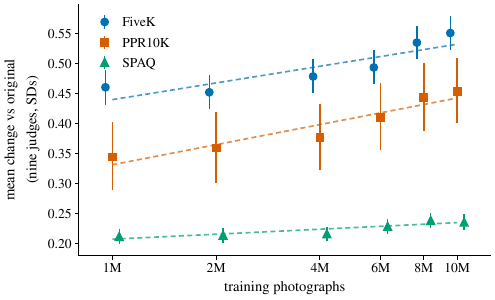}
\caption{\textbf{More training photographs, better edits.} Mean change against the unedited original, averaged over the
nine judges.}
\label{fig:scaling-main}
\end{figure}

\textbf{Corpus size.} More photographs help. We trained ATune on 1M, 2M, 4M, 6M, 8M and \numPaperSize{} photographs
(\cref{fig:scaling-main}). Each larger corpus wins or ties, and no judge on any test set prefers a smaller one
(\cref{tab:ablation}a). The gains flatten at the top. We trained one model per size, so we report the trend and fit
no scaling law.

\textbf{The dial.} The dial $\tau$ is the target quality percentile: $\tau{=}0.99$ asks for a photograph as good, by
\liqe, as the corpus's 99th-percentile photograph (\cref{sec:method-target}). A higher target gives better edits: at 4M, $\tau{=}0.99$ beats $0.97$ and $0.97$ beats $0.90$
under every test set, and no judge prefers the lower target (\cref{tab:ablation}b).

\section{Limitations}
\label{sec:limitations}

\textbf{Global edits.} All 13 attributes are image-wide; faces, skies and other regions cannot be treated separately.
\textbf{Speed.} The search renders repeatedly: a median of \numSolverSec\,s per photograph, against \numClaudeSec\,s
for a Claude call.
\textbf{One shot.} Each frontier model decided once, without seeing its result, and so did ATune; either could
iterate on its own output and do better.

\section{Conclusion}
\label{sec:conclusion}

A large corpus of good photographs is enough to teach a model photo retouching, stated declaratively as targets for 13
measured attributes that a search reaches with an editor's own sliders. One model drives two editors, every edit is a
slider setting a user can read and undo, and a \numPaperParams-parameter model trained without a single expert edit
matches Claude Opus 5 and beats GPT-6 Astra with human raters. Edit quality appears to be capped not by the number of
expert edits but by the quality of the corpus. \textbf{Future work:} a higher-quality corpus, to see how far better photographs alone go, and
a network that maps attributes straight to slider values, cutting search time to milliseconds.

{
    \small
    \bibliographystyle{ieeenat_fullname}
    \bibliography{main}
}
\clearpage
\appendix

\newcommand{\tocgroup}[1]{\vspace{4pt}\noindent\textbf{#1}\par\vspace{1pt}}
\newcommand{\tocline}[2]{\noindent\makebox[1.6em][l]{\ref{#1}}#2\dotfill\makebox[1.2em][r]{\pageref{#1}}\par}
\begin{center}\textbf{\large Contents}\end{center}
\vspace{-6pt}
{\small
\tocgroup{Method}
\tocline{app:attrsel}{Attribute measures}
\tocline{app:distortions}{Training defects}
\tocline{app:training}{Training ATune}
\tocline{app:solver}{Solver}
\tocline{app:darktable}{A second editor: darktable}
\tocgroup{Comparisons}
\tocline{app:claude}{The frontier models}
\tocline{app:baselines}{Baselines}
\tocline{app:sets}{Test sets}
\tocline{app:expertfull}{Against the expert: all methods}
\tocgroup{Human study}
\tocline{app:human}{Human study}
\tocgroup{More results}
\tocline{app:analysis}{Two exploratory analyses}
\tocline{app:scaling}{Training-set size: every judge}
\tocline{app:robust}{Robustness: judges given the full-size image}
\tocline{app:qual}{Qualitative examples}
\tocgroup{Reproducibility}
\tocline{app:repro}{Reproducibility, data and impact}
}
\vspace{8pt}

\section{Attribute measures}
\label{app:attrsel}

\paragraph{The attributes.} Each global tool in a photo editor changes one quality of the whole photograph. We went
through the global tools of RawTherapee and darktable and wrote down what each one changes. This gives the 13
attributes of Tab.~1 in the main paper:
\begin{itemize}[leftmargin=*,itemsep=0pt,topsep=2pt]
  \item \textbf{Tone.} Exposure changes brightness. The contrast tool changes contrast. The shadows and highlights
  tools change shadow depth and highlight level. Clarity changes local contrast. Dehaze changes haze. The black and
  white points change clipping at black and at white.
  \item \textbf{Colour.} Saturation changes colourfulness. White balance changes colour temperature and tint.
  \item \textbf{Detail.} Sharpening changes sharpness. Noise reduction changes noise.
\end{itemize}
Local tools, such as masks, crops and retouching, are out of scope.

\paragraph{The tests.} Each attribute needs a measure: a function that turns the pixels into a number. A good measure
changes when its attribute changes, and stays the same otherwise. We therefore check three things:
\begin{itemize}[leftmargin=*,itemsep=0pt,topsep=2pt]
  \item \textbf{Response.} Does the number follow its own attribute? KADID-10k~\cite{lin2019kadid} has 81
  photographs, each distorted 25 ways at five strengths. Several distortions change one of our attributes, such as
  brightening, blur, sharpening, added noise and denoising. KADID has no white-balance or haze distortion, so we add
  our own: temperature and tint shifts, and a haze veil. For each photograph and distortion, we compute Spearman's $\rho$
  between the five strengths and the measure's five values. We then take the mean $\rho$ over the 81 photographs. Some attributes are changed by more than one
  distortion; brightness, for example, by brightening, darkening and a mean shift. Response is the lowest $|\rho|$
  among them. Higher is better.
  \item \textbf{Specificity.} Does the number ignore the other attributes? This is the mean $|\rho|$ over the other
  attributes' distortions. Lower is better.
  \item \textbf{Stability.} Does the number ignore changes that leave the photograph looking the same? We take
  \numRobustN{} SPAQ photographs~\cite{fang2020spaq} and change each in five ways: resize to 512 or 768\,px, save
  as JPEG at quality 75 or 90, and crop 3\%. For each change we measure every photograph before and after, and correlate the two lists (Spearman's $\rho$). A
  $\rho$ near 1 means the change did not reorder the photographs. Stability is the lowest $\rho$ of the five
  changes. Higher is better.
\end{itemize}

\begin{table}[t]
\centering
\caption{\textbf{Candidate measures.} Own: response. Others: specificity. Bold: the measure kept.}
\label{tab:candidates}
\footnotesize
\begin{adjustbox}{max width=\linewidth}
\begin{tabular}{@{}llccc@{}}
\toprule
Attribute & Measure & Own $|\rho|$ & Others $|\rho|$ & Stability $\rho$ \\
\midrule
Brightness & mean grey level & 1.00 & 0.28 & 1.00 \\
 & mean $L^*$ & 1.00 & 0.43 & 1.00 \\
 & log key & 1.00 & 0.51 & 1.00 \\
 & \textbf{median log intensity} & \textbf{1.00} & \textbf{0.17} & \textbf{1.00} \\
\addlinespace[2pt]
Contrast & RMS contrast & 0.98 & 0.64 & 1.00 \\
 & Michelson contrast & 1.00 & 0.57 & 1.00 \\
 & global contrast factor (multi-resolution) & 0.98 & 0.77 & 0.98 \\
 & \textbf{$L^*$ interquartile range} & \textbf{0.95} & \textbf{0.38} & \textbf{0.99} \\
\addlinespace[2pt]
Sharpness & Laplacian variance & 0.98 & 0.58 & 0.95 \\
 & Tenengrad & 0.99 & 0.60 & 0.96 \\
 & \textbf{maximum local variation} & \textbf{0.99} & \textbf{0.63} & \textbf{0.97} \\
 & CPBD & 0.78 & 0.60 & 0.94 \\
\addlinespace[2pt]
Noise & Immerk\ae r residual, whole image & 0.91 & 0.69 & 0.87 \\
 & wavelet MAD & 0.97 & 0.71 & 0.81 \\
 & weak-texture PCA & 0.99 & 0.60 & 0.58 \\
 & high-pass texture energy & 0.33 & 0.70 & 0.96 \\
 & \textbf{block residual at two scales} & \textbf{0.93} & \textbf{0.59} & \textbf{0.91} \\
\addlinespace[2pt]
Colourfulness & \textbf{Hasler and S\"usstrunk} & \textbf{0.99} & \textbf{0.45} & \textbf{1.00} \\
 & mean saturation & 1.00 & 0.54 & 1.00 \\
 & mean chroma & 0.94 & 0.51 & 1.00 \\
 & spread of $a^*$, $b^*$ & 0.93 & 0.56 & 1.00 \\
\addlinespace[2pt]
Colour temperature, tint & \textbf{CCT of the mean colour (temperature)} & \textbf{0.99} & \textbf{0.20} & \textbf{1.00} \\
 & \textbf{mean $a^*$ (tint)} & \textbf{1.00} & \textbf{0.24} & \textbf{1.00} \\
 & mean $b^*$ & 1.00 & 0.26 & 1.00 \\
 & cast magnitude & 0.47 & 0.60 & 1.00 \\
\addlinespace[2pt]
Shadow, highlight clipping & \textbf{signed clipping at black (shadow)} & \textbf{1.00} & \textbf{0.56} & \textbf{1.00} \\
 & \textbf{signed clipping at white (highlight)} & \textbf{1.00} & \textbf{0.59} & \textbf{0.99} \\
\addlinespace[2pt]
Shadow depth & \textbf{$L^*$ at the 5th percentile} & \textbf{0.99} & \textbf{0.66} & \textbf{1.00} \\
 & mean $L^*$ of the darkest decile & 0.99 & 0.70 & 1.00 \\
\addlinespace[2pt]
Highlight level & \textbf{$L^*$ at the 95th percentile} & \textbf{0.98} & \textbf{0.54} & \textbf{0.99} \\
 & mean $L^*$ of the brightest decile & 0.99 & 0.61 & 0.99 \\
\addlinespace[2pt]
Local contrast & \textbf{midtone-weighted unsharp-mask energy} & \textbf{0.99} & \textbf{0.47} & \textbf{0.95} \\
 & locally normalised standard deviation & 0.99 & 0.52 & 0.97 \\
\addlinespace[2pt]
Haze & \textbf{dark channel prior} & \textbf{1.00} & \textbf{0.74} & \textbf{0.99} \\
 & colour attenuation prior & 1.00 & 0.55 & 1.00 \\
\bottomrule
\end{tabular}

\end{adjustbox}
\end{table}

\begin{figure}[H]
\centering
\includegraphics[width=\linewidth]{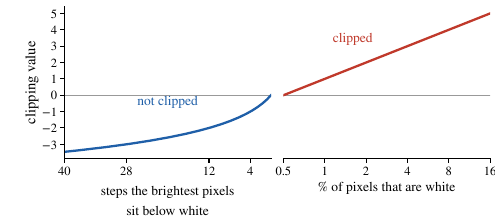}
\caption{\textbf{The highlight clipping value.} Left: not clipped, the value falls as the brightest pixels move away
from white. Right: clipped, it rises by one each time the white area doubles.}
\label{fig:clipping}
\end{figure}

\paragraph{The choice.} \Cref{tab:candidates} scores \numAttrCandN{} candidates for the eleven attributes that have
more than one established measure. Clipping has only one, the share of pixels at black or at white; we use its signed form (\cref{fig:clipping}). Among candidates
with response and stability of at least 0.9, we keep the one that reacts least to the other attributes. This picks every bold row, with two exceptions:
\begin{itemize}[leftmargin=*,itemsep=0pt,topsep=2pt]
  \item \textbf{Sharpness.} Averaged over the other attributes, Laplacian variance and Tenengrad react less
  ($\numLapOthers$ and $\numTenOthers$, against $\numMlvOthers$). The average hides one attribute, noise: they rise
  with added noise as much as with sharpening ($\rho{=}{\numLapNoiseLeak}$ and $\numTenNoiseLeak$). Maximum local
  variation reacts to noise with $\rho{=}{\numMlvNoiseLeak}$, so we kept it.
  \item \textbf{Haze.} The colour attenuation prior~\cite{zhu2015cap} reacts less to the other attributes
  ($\numCapOthers$, against $\numDcpOthers$), but we kept the dark channel
  prior~\cite{he2009haze} because the dehaze tools of both editors use it.
\end{itemize}
Exposure is not in the table because it is the same as brightness ($\rho{=}\numExposureBrightRho$). Some
measures react to other attributes by design. Clipping moves whenever pixels are pushed to black or white, and shadow
depth and highlight level move with contrast.

\paragraph{Signed clipping.} Pixel values run from 0 (black) to 255 (white). The obvious way to measure
highlight clipping is the share of white pixels. But that share is zero for every photograph with no white pixels, whether its brightest pixels are
at 254 or at 150. So it cannot say how close a photograph is to clipping. We fix this with a signed value
(\cref{fig:clipping}):
\begin{itemize}[leftmargin=*,itemsep=0pt,topsep=2pt]
  \item \textbf{Clipped}, when at least $\numClipMZeroPct\%$ of the pixels are white: the value is positive and rises by one each
  time the white area doubles.
  \item \textbf{Not clipped}: the value is negative and falls as the brightest pixels move further below white.
\end{itemize}
Let $s$ be the share of white pixels (above $T{=}\numClipT$), $m_0{=}\numClipMZeroPct\%$, and $q$ the value
of the brightest $\numClipMZeroPct\%$ of pixels, so that $T{-}q$ is how many steps they sit below white. Then
\begin{equation*}
c \;=\;
\begin{cases}
\log_2\!\left(\dfrac{s}{m_0}\right) & \text{if } s \ge m_0 \quad \text{(clipped)},\\[8pt]
-\log_2\!\left(1 + \dfrac{T - q}{\numClipK}\right) & \text{otherwise} \quad \text{(not clipped)}.
\end{cases}
\end{equation*}
For example, when the brightest pixels sit $\numClipK$ steps below white, $c{=}{-1}$. Shadows use the same formula with black in place
of white.

\section{Training defects}
\label{app:distortions}

\paragraph{Why defects.} The corpus holds only finished photographs. To learn how a photograph should change, ATune
needs pairs: a worse version and its finished original. We make the worse version by adding defects to the original.
The defects should look like real mistakes a camera or a quick edit makes, at a realistic size. We chose defects that an editor tool can
reverse, for example a wrong exposure with the exposure slider or noise with noise reduction. The one exception is
JPEG compression, which no tool can reverse. The model learns to undo these defects, and an editor can carry out
every fix.

\begin{table}[h]
\centering
\caption{\textbf{The training defects} at the weakest and strongest strength $s$ a copy can receive. $\pm$ and $\times$/$\div$: the defect goes either way. $\sigma$: standard deviation of the noise. Shot noise varies with brightness, so its $\sigma$ is given for a mid-grey area.}
\label{tab:distortions}
\footnotesize
\begin{tabular}{@{}lll@{}}
\toprule
Defect & Weakest ($s{=}0.3$) & Strongest ($s{=}1.15$) \\
\midrule
\multicolumn{3}{@{}l}{\emph{Tone}} \\
\quad Exposure & $\pm$0.6 EV & $\pm$2.2 EV \\
\quad Black point & $\pm$5\% of range & $\pm$21\% of range \\
\quad Highlight clipping & white above 86\% & white above 70\% \\
\quad Highlight compression & white at 78\% & white at 14\% \\
\quad Contrast & $\times$/$\div$\,1.3 & $\times$/$\div$\,3.1 \\
\quad Haze & 16\% opacity & 63\% opacity \\
\quad Local contrast loss & 24\% & 92\% \\
\midrule
\multicolumn{3}{@{}l}{\emph{Colour}} \\
\quad Temperature & $\pm$12\% red/blue gain & $\pm$47\% red/blue gain \\
\quad Tint & $\pm$7\% green gain & $\pm$25\% green gain \\
\quad Saturation & $\times$/$\div$\,1.5 chroma & $\times$/$\div$\,4.2 chroma \\
\midrule
\multicolumn{3}{@{}l}{\emph{Detail}} \\
\quad Blur & $\sigma$\,=\,0.9\,px & $\sigma$\,=\,3.4\,px \\
\quad Oversharpening & amount 0.9 & amount 3.4 \\
\midrule
\multicolumn{3}{@{}l}{\emph{Noise}} \\
\quad Luminance noise & $\sigma$\,=\,4/255 & $\sigma$\,=\,17/255 \\
\quad Shot noise & $\sigma$\,=\,5/255 & $\sigma$\,=\,9/255 \\
\quad Chroma noise & $\sigma$\,=\,4 Lab units & $\sigma$\,=\,14 Lab units \\
\midrule
\multicolumn{3}{@{}l}{\emph{Compression}} \\
\quad JPEG & quality 66 & quality 10 \\
\bottomrule
\end{tabular}

\end{table}

\paragraph{Making a copy.}
\begin{itemize}[leftmargin=*,itemsep=0pt,topsep=2pt]
  \item Each photograph gets \numDefCopies{} copies.
  \item Each copy gets \numDefOpsLo{} or \numDefOpsHi{} different defects from \cref{tab:distortions}. They are applied in a fixed
  order: tone, colour, blur or sharpening, noise, and JPEG. Like Real-ESRGAN~\cite{wang2021realesrgan}, we add noise
  after blur and compress last.
  \item Each defect gets a strength between \numDefSMin{} and a maximum set by the original's \liqe{} score:
  \numDefMaxLo{} below \numDefLiqeLo, \numDefMaxMid{} below \numDefLiqeHi{} and \numDefMaxHi{} above. Weaker photographs get milder defects, so they stay recognisable. Where a defect can
  go either way, such as brighter or darker, the direction is random.
  \item Two safeguards. In a \numDefOpsHi-defect copy, one defect is capped at \numDefThirdCap. A copy whose share of
  clipped pixels rises by more than \numDefClipRise{} points is drawn again.
\end{itemize}
We manually checked copies from this recipe to make sure the defects look reasonable.

\section{Training ATune}
\label{app:training}

\providecommand{\attrsword}{attributes}
\begin{figure}[t]
\centering
\resizebox{0.62\linewidth}{!}{%
\begin{tikzpicture}[
  font=\sffamily\scriptsize,
  layer/.style={rounded corners=3.5pt, draw=black!25, line width=0.5pt, align=center,
                minimum width=2.75cm, minimum height=5.2mm, inner ysep=1.5pt},
  lnorm/.style={layer, fill=black!7},
  linear/.style={layer, fill=blue!12},
  act/.style={layer, fill=green!55!black!12},
  filmst/.style={layer, fill=orange!22},
  headst/.style={layer, fill=violet!12},
  arr/.style={-{Stealth[length=1.8mm]}, line width=0.7pt, draw=black!50},
  carr/.style={-{Stealth[length=1.8mm]}, line width=0.7pt, draw=orange!70!black!70},
  lab/.style={font=\sffamily\scriptsize, text=black!55, align=center}
]
\def\pitch{8.2mm}
\node[layer, fill=black!12] (dino) at (0,0) {\dinov\ {\color{black!55}(frozen)}};
\node[lnorm, above=3.2mm of dino] (ln0) {LayerNorm};
\node[linear, above=2.6mm of ln0] (proj) {Linear $1024 \to 2048$};
\node[lnorm, above=7.5mm of proj] (bln) {LayerNorm};
\node[filmst, above=2.6mm of bln] (film) {FiLM\quad$(1{+}\gamma)\odot h+\beta$};
\node[linear, above=2.6mm of film] (fc1) {Linear $2048 \to 4096$};
\node[act, above=2.6mm of fc1] (gelu) {GELU};
\node[linear, above=2.6mm of gelu] (fc2) {Linear $4096 \to 2048$};
\node[circle, draw=black!40, line width=0.6pt, fill=white, inner sep=1.2pt, font=\sffamily\small, above=3mm of fc2] (add) {$+$};
\begin{scope}[on background layer]
\node[rounded corners=6pt, fill=blue!5, draw=black!25, line width=0.6pt, inner xsep=5pt, inner ysep=5pt,
      fit=(bln)(film)(fc1)(gelu)(fc2)(add)] (block) {};
\end{scope}
\node[font=\sffamily\footnotesize\bfseries, rotate=90, anchor=south] at ([xshift=-1.5mm]block.west) {$\times3$};
\node[lnorm, above=7.5mm of add] (lnf) {LayerNorm};
\node[headst, above=2.6mm of lnf] (head) {Linear $2048 \to$ mixture head};
\begin{scope}[shift={($(head.north)+(-1.62,0.62)$)}]
  \draw[black!30, line width=0.4pt] (0,0) -- (1.30,0);
  \draw[black!30, line width=0.4pt] (0,0) -- (0,0.92);
  \node[font=\sffamily\tiny, text=black!50, anchor=north east] at (1.30,0) {$\Delta$ attr.\ $j$};
  \foreach \cx/\cy/\ang/\rx/\ry/\op in {0.50/0.44/32/0.30/0.12/0.55, 0.70/0.60/24/0.24/0.10/0.42, 0.58/0.50/38/0.34/0.14/0.30, 0.34/0.66/-12/0.15/0.08/0.22, 0.95/0.30/12/0.12/0.07/0.18} {
    \fill[blue!60, fill opacity=\op, rotate around={\ang:(\cx,\cy)}] (\cx,\cy) ellipse ({\rx} and {\ry});
  }
  \draw[blue!70!black, line width=0.6pt, smooth] plot coordinates {(0.10,0.00) (0.25,0.05) (0.40,0.13) (0.55,0.17) (0.70,0.13) (0.85,0.07) (1.00,0.04) (1.15,0.01)};
  \draw[red!75, line width=0.9pt] (0.58,-0.04) -- (0.58,0.92);
\end{scope}
\node[lab, anchor=west] at ($(head.north)+(0.72,0.95)$) {\shortstack[l]{5 Gaussians over the\\ changes of 13 \attrsword;\\ target $=\mathbf{o}_0+$ marginal\\ \textcolor{red!75}{median} change}};
\node[lab, below=3mm of dino] (img) {image};
\draw[arr] (img) -- (dino);
\draw[arr] (dino) -- (ln0) node[lab, midway, right=1mm] {$\mathbf{e}\in\mathbb{R}^{1024}$};
\draw[arr] (ln0) -- (proj);
\draw[arr] (proj) -- (bln);
\draw[arr] (bln) -- (film); \draw[arr] (film) -- (fc1); \draw[arr] (fc1) -- (gelu);
\draw[arr] (gelu) -- (fc2); \draw[arr] (fc2) -- (add);
\coordinate (rsrc) at ($(proj.north)!0.45!(bln.south)$);
\draw[arr, rounded corners=3pt] (rsrc) -- ++(-1.72,0) |- (add.west);
\draw[arr] (add) -- (lnf);
\draw[arr] (lnf) -- (head);
\draw[arr] (head.north) -- ++(0,0.42);
\node[rounded corners=4pt, draw=black!25, line width=0.5pt, fill=orange!10, align=center, anchor=west]
  (cin) at ($(bln.east)+(0.62,-0.72)$) {$\mathbf{o}_0$ (13 \attrsword),\\ $p_{\mathrm{start}}$,\ $\delta{=}\max(\tau{-}p_{\mathrm{start}},0)$};
\node[rounded corners=4pt, draw=black!25, line width=0.5pt, fill=orange!10, above=3mm of cin]
  (cmlp) {MLP $\to 64$};
\draw[carr] (cin) -- (cmlp);
\draw[carr, rounded corners=3pt] (cmlp.north) |- (film.east)
  node[lab, pos=0.75, above=0.6mm, xshift=3mm] {$\gamma,\beta$};
\node[lab, anchor=north, text width=2.2cm] at ([yshift=-1.5mm]cin.south) {zero-initialised:\\ starts unconditional};
\end{tikzpicture}}
\providecommand{\archcaption}{\textbf{ATune} (\mainsec{sec:method-target}{3.3}; \numPaperParams{} parameters): a FiLM-conditioned residual MLP over the frozen \dinov embedding. The conditioning carries the photo's measured attributes $\mathbf{o}_0$, its quality percentile $p_{\mathrm{start}}$ and the requested lift $\delta$; the head emits five full-covariance Gaussians over the \emph{changes} of the 13 attributes, and each attribute's target is its current value plus the median of its marginal change.}
\caption{\archcaption}
\label{fig:arch}
\end{figure}

\paragraph{Training data.} Each training example is one copy (\cref{app:distortions}), described by:
\begin{itemize}[leftmargin=*,itemsep=0pt,topsep=2pt]
  \item its content, as a \dinov embedding;
  \item its 13 attributes, in buckets, with a flag where a value is missing;
  \item its quality, as a \liqe{} percentile;
  \item the lift: how much higher its original's percentile is.
\end{itemize}
The target is the change that takes the copy's attributes to its original's. One example in five is an untouched
original, with lift and change both zero, so the model also learns when to leave a photograph alone.

\paragraph{Why predict the change.} If the model predicted the target attributes directly, it could lower its loss by
copying the current ones, which it can see. It would learn to leave photographs as they are. Predicting the change
removes this shortcut, and the model still sees where the photograph starts.

\paragraph{Corpora.}
\begin{itemize}[leftmargin=*,itemsep=0pt,topsep=2pt]
  \item \textbf{1M to 6M.} Sampled evenly over \qalign{} score bands of width 0.1, from 2 to 5. A band with too few
  photographs passes the shortfall to the next band up. 41\% of the photographs score at least 4.
  \item \textbf{8M.} 60\% of the photographs score at least 4. Trained for up to \numEightMSteps{} steps.
  \item \textbf{\numPaperSize{} (ATune).} \numPaperRows{} photographs, \numPaperHiPct\% scoring at least 4. Trained for
  up to \numPaperSteps{} steps; we keep step \numPaperBest.
\end{itemize}

\paragraph{Optimisation.}
\begin{itemize}[leftmargin=*,itemsep=0pt,topsep=2pt]
  \item \textbf{Loss.} The negative log-likelihood of the mixture, Eq.~(1) of the main paper.
  \item \textbf{Optimiser.} AdamW with learning rate $\numTrainLR$ and weight decay \numTrainWD. Linear warm-up
  over the first \numTrainWarmPct\% of steps, then cosine decay to $\numTrainLRFloor$.
  \item \textbf{Batch.} \numTrainBatch{} examples, \numTrainBatchGPU{} per GPU on \numTrainGPUs{} H100s. The photographs are measured once; training the
  \numPaperSize{} model then takes \numPaperTrainMin{} minutes.
  \item \textbf{Checkpoint.} The one with the lowest validation loss, checked every \numTrainEvalEvery{} steps after
  the first \numTrainSelFromPct\% of training.
\end{itemize}

\paragraph{Lifts at test time.} In training, the lift is the quality gap the defects create: median
$\numTrainLiftMed$, with 99\% of lifts below $\numTrainLiftHi$. At test time with $\tau{=}\numSelTau$, photographs ask for
larger lifts:
\begin{center}\footnotesize
\begin{tabular}{@{}lccc@{}}
\toprule
 & FiveK & PPR10K & SPAQ \\
\midrule
median lift & $\numTestLiftMedFiveK$ & $\numTestLiftMedPPR$ & $\numTestLiftMedSPAQ$ \\
lift above $\numTrainLiftHi$ & \numTestLiftAboveFiveK\% & \numTestLiftAbovePPR\% & \numTestLiftAboveSPAQ\% \\
\bottomrule
\end{tabular}
\end{center}
For lifts above $\numTrainLiftHi$, the model goes beyond what it saw in training. Even so, a higher $\tau$
gives better edits (\mainsec{sec:ablations}{4.5}).

\paragraph{Choosing $\tau$.} To avoid tuning on the test sets, we took \numSelSetN{} photographs from FiveK's training split.
We edited them with the 4M model at $\tau$ set to \numSelTauGrid{} and scored the edits with five held-out judges. $\tau{=}\numSelTau$
gave the highest quality.

\section{Solver}
\label{app:solver}

\input{fig/loop}

\begin{table*}[t]
\centering
\caption{\textbf{Our controls} and the editor tools each one sets. Each control is named after the attribute it changes. darktable has a 14th control, used only when the
model asks for more noise.}
\label{tab:sliders}
\footnotesize
\providecommand{\Attrword}{Control (attribute)}
\begin{tabular}{@{}lp{0.38\linewidth}p{0.36\linewidth}@{}}
\toprule
\Attrword{} & RawTherapee 5.8 & darktable 3.8 \\
\midrule
temperature, tint & RGB curves, mid-tone point per channel & white balance, channel coefficients \\
brightness & exposure: brightness & exposure \\
contrast & exposure: contrast & contrast brightness saturation: contrast \\
colourfulness & exposure: saturation & contrast brightness saturation: saturation \\
shadow depth & shadows \& highlights: shadows (lift); tone curve: shadows (deepen) & shadows and highlights: shadows \\
highlight level & tone curve: highlights & shadows and highlights: highlights \\
shadow clipping & exposure: black & exposure: black level correction \\
highlight clipping & tone curve: white point & shadows and highlights: white point adjustment \\
haze & haze removal: strength & haze removal: strength \\
local contrast & local contrast, radius 8\,px & local contrast: detail \\
sharpness & sharpening (unsharp mask, radius 0.5\,px) & sharpen (radius 2\,px) \\
noise (lower) & noise reduction: luminance & denoise (non-local means): luma \\
noise (raise) & -- & sharpen, low threshold (radius 1\,px) \\
\bottomrule
\end{tabular}

\end{table*}

For each attribute, we pick the editor tools a photographer would use to change it (\mainsec{sec:editing}{3.4}).
These tools use very different units, such as exposure in stops and a sharpening radius in pixels. So the solver does
not set them directly. Instead, each attribute gets a \emph{control}: one number between $-1$ and $1$. A fixed mapping
turns the control into tool settings. \Cref{fig:loop} shows how attribute, measure, control and tool fit together, and \cref{tab:sliders} lists the controls and the tools they set. The solver chooses the
control values. It has two aims: bring each attribute close to its goal, and leave no attribute worse than it was.

\paragraph{Objective.} For attribute $i$, $r_i$ is its value after the edit minus its goal, in buckets.
$r_i^0$ is the same before the edit. The solver minimises
\begin{align*}
L \;=\;& \sum_i w_i\,h\big(\max(|r_i|-b,\,0)\big)\\
&+\;\numSolverHarm\sum_i \max\!\big(|r_i|-\max(|r_i^0|,\,\numSolverTol+s_i),\,0\big)^2 .
\end{align*}
\begin{itemize}[leftmargin=*,itemsep=0pt,topsep=2pt]
  \item \textbf{First term: reach the goal.} $b$ is a tolerance: an attribute within $b$ buckets of its goal counts as on target and costs nothing. $h$ is the Huber loss~\cite{huber1964}
  (knee at $\numSolverHuber$), so the cost grows only linearly far from the goal and an unreachable goal cannot dominate. Each attribute's term has a weight $w_i$.
  \item \textbf{Second term: do no harm.} It penalises an attribute that ends further from its goal than it started.
  An attribute may still move within the success tolerance of \numSolverTol{} buckets, adjusted by a slack $s_i$.
\end{itemize}

\paragraph{Search.} The solver adjusts all controls together. At each step it renders the photograph, measures the
13 attributes and moves the controls to lower $L$. The step is a Levenberg--Marquardt step~\cite{levenberg1944,marquardt1963} within the controls' bounds,
with the Huber loss handled by iteratively reweighted least squares~\cite{holland1977irls}. How each control moves each attribute (the Jacobian) is estimated
by finite differences and then updated cheaply with Broyden's method~\cite{broyden1965}. The search stops after at most \numSolverMaxRenders{} renders on
the \numSolverFrame\,px frame. The final controls are then applied at full resolution; the result is measured again and corrected
at most \numSolverCorr.

\paragraph{Calibrating the controls.} Each editor needs a mapping from our controls to its tools: which tools a
control sets, how strongly per unit, and the solver's $b$, $w_i$ and $s_i$. One file per editor holds this mapping. An LLM coding agent
edits it in a keep-if-better loop, following autoresearch~\cite{karpathy2026autoresearch}: it proposes one change at a
time, and a fixed program scores it. A candidate mapping is scored by one number: the share of the model's goals that the solver reaches within \numSolverTol{} buckets. A change to the
mapping is kept only if it raises the score.
The calibrated solver settings are:
\begin{itemize}[leftmargin=*,itemsep=0pt,topsep=2pt]
  \item \textbf{RawTherapee.} \numSolverRT
  \item \textbf{darktable.} \numSolverDt
\end{itemize}

\paragraph{Result.} On \numCalN{} held-out photographs, the solver reaches \numCalReachRT\% of the goals on RawTherapee and
\numCalReachDt\% on darktable.

\section{A second editor: darktable}
\label{app:darktable}

\paragraph{Mapping.} To show that ATune is not tied to one editor, we also run it through darktable
3.8~\cite{darktable}. ATune itself is unchanged; we only write a new mapping from our controls to darktable's modules
(\cref{tab:sliders}):
\begin{itemize}[leftmargin=*,itemsep=0pt,topsep=2pt]
  \item \textbf{Shared modules.} Temperature and tint both set the white-balance module. Contrast and colourfulness
  both set the contrast-brightness-saturation module.
  \item \textbf{Extra control.} When the model asks for more noise, a 14th control applies a low-threshold sharpen,
  which strengthens the photograph's own fine noise.
  \item \textbf{Strength.} In RawTherapee, a tool's setting is mostly the control value times a constant. darktable's
  modules do not respond that evenly, so each control maps to a module setting through a few fixed points joined by
  straight lines. For haze, control values \numDtHazeT{} set the dehaze strength to \numDtHazeV. The points are
  calibrated by the same autoresearch loop as RawTherapee's mapping (\cref{app:solver}).
  \item \textbf{Defaults.} Every other module stays at its default, with automatic sharpening off. We render with
  darktable-cli.
\end{itemize}
Nine judges on three test sets give 27 results. On darktable, ATune's edits significantly improve the photograph in
\numDtPosCells{} of them and worsen it in none (\cref{tab:dtmain}). At least \numDtWinMinFrontier{} of the nine judges prefer them to Claude Opus 5's and
GPT-6 Astra's on every test set, and at most \numDtLossMaxFrontier{} prefer theirs (\cref{tab:dtwtl}). Against ATune's own
RawTherapee edits, the judges split \numVEditorDtFiveK{} on FiveK, \numVEditorDtPPR{} on PPR10K and \numVEditorDtSPAQ{}
on SPAQ (darktable / neither / RawTherapee; first row of \cref{tab:dtwtl}), so RawTherapee remains our main editor.

\begin{table*}[t]
\centering
\caption{\textbf{ATune on both editors, every judge against the unedited original} (mean per-photograph
change, as in \maintab{tab:main}{2}).}
\label{tab:dtmain}
\setlength{\tabcolsep}{4pt}\footnotesize
\begin{tabular}{@{}lccccccccc@{}}
\toprule
 & \multicolumn{3}{c}{Used in training} & \multicolumn{6}{c}{Held out} \\
\cmidrule(lr){2-4}\cmidrule(l){5-10}
Method & Q-Align & Q-Align\textsubscript{A} & LIQE & PaQ-2-PiQ & TOPIQ$^{\times}$ & LAION & CLIP-IQA$^{\times}$ & QualiCLIP$^{\times}$ & DeQA \\
\midrule
\multicolumn{10}{l}{\textit{FiveK}} \\
ATune on RawTherapee & +0.207 & +0.225 & +0.71 & +2.03 & +1.99 & +0.210 & +10.48 & +4.36 & +0.102 \\
ATune on darktable & +0.207 & +0.237 & +0.60 & +2.06 & +1.98 & +0.205 & +9.17 & +5.78 & +0.102 \\
\midrule
\multicolumn{10}{l}{\textit{PPR10K}} \\
ATune on RawTherapee & +0.155 & +0.148 & +0.35 & +1.22 & +1.59 & +0.124 & +10.00 & +2.68 & +0.072 \\
ATune on darktable & +0.103 & +0.138 & +0.22 & +1.16 & +1.59 & +0.113 & +6.50 & +1.59 & \nsig{$-$0.018} \\
\midrule
\multicolumn{10}{l}{\textit{SPAQ}} \\
ATune on RawTherapee & +0.125 & +0.153 & +0.39 & +3.80 & +1.88 & +0.158 & +5.46 & \nsig{+0.12} & +0.132 \\
ATune on darktable & +0.116 & +0.179 & +0.35 & +3.78 & +1.77 & +0.162 & +3.96 & +1.94 & +0.107 \\
\bottomrule
\end{tabular}

\end{table*}

\begin{table*}[t]
\centering
\caption{\textbf{ATune on darktable against each method}, as in \maintab{tab:wtl}{3}; the first row
compares the two editors.}
\label{tab:dtwtl}
\setlength{\tabcolsep}{4pt}\footnotesize
\begin{tabular}{@{}lcccccc@{}}
\toprule
 & \multicolumn{2}{c}{FiveK} & \multicolumn{2}{c}{PPR10K} & \multicolumn{2}{c}{SPAQ} \\
\cmidrule(lr){2-3}\cmidrule(lr){4-5}\cmidrule(l){6-7}
ATune on darktable vs & all 9 & held out & all 9 & held out & all 9 & held out \\
\midrule
ATune on RawTherapee & \bad{1\,/\,6\,/\,2} & {\scriptsize 1\,/\,4\,/\,1} & \bad{0\,/\,4\,/\,5} & {\scriptsize \bad{0\,/\,3\,/\,3}} & \bad{2\,/\,4\,/\,3} & {\scriptsize \bad{1\,/\,3\,/\,2}} \\
\hdashline
Claude Opus 5, same editor & \good{5\,/\,2\,/\,2} & {\scriptsize \good{3\,/\,1\,/\,2}} & \good{7\,/\,2\,/\,0} & {\scriptsize \good{4\,/\,2\,/\,0}} & \good{8\,/\,1\,/\,0} & {\scriptsize \good{5\,/\,1\,/\,0}} \\
GPT-6 Astra, same editor & \good{8\,/\,1\,/\,0} & {\scriptsize \good{5\,/\,1\,/\,0}} & \good{7\,/\,2\,/\,0} & {\scriptsize \good{4\,/\,2\,/\,0}} & \good{9\,/\,0\,/\,0} & {\scriptsize \good{6\,/\,0\,/\,0}} \\
\hdashline
CSRNet~\cite{he2020csrnet}$^\ddagger$ & \good{7\,/\,0\,/\,2} & {\scriptsize \good{4\,/\,0\,/\,2}} & \good{8\,/\,0\,/\,1} & {\scriptsize \good{5\,/\,0\,/\,1}} & \good{8\,/\,1\,/\,0} & {\scriptsize \good{5\,/\,1\,/\,0}} \\
AdaInt~\cite{yang2022adaint}$^\dagger$ & \good{7\,/\,1\,/\,1} & {\scriptsize \good{4\,/\,1\,/\,1}} & \good{7\,/\,1\,/\,1} & {\scriptsize \good{5\,/\,0\,/\,1}} & \good{7\,/\,2\,/\,0} & {\scriptsize \good{4\,/\,2\,/\,0}} \\
NeurOP~\cite{wang2022neurop}$^\dagger$ & \good{8\,/\,0\,/\,1} & {\scriptsize \good{5\,/\,0\,/\,1}} & \good{8\,/\,0\,/\,1} & {\scriptsize \good{5\,/\,0\,/\,1}} & \good{7\,/\,2\,/\,0} & {\scriptsize \good{4\,/\,2\,/\,0}} \\
UEGAN~\cite{ni2020uegan} & \good{8\,/\,1\,/\,0} & {\scriptsize \good{5\,/\,1\,/\,0}} & \good{7\,/\,1\,/\,1} & {\scriptsize \good{4\,/\,1\,/\,1}} & \good{8\,/\,1\,/\,0} & {\scriptsize \good{5\,/\,1\,/\,0}} \\
EnlightenGAN~\cite{jiang2021enlightengan} & \good{7\,/\,2\,/\,0} & {\scriptsize \good{4\,/\,2\,/\,0}} & \good{8\,/\,0\,/\,1} & {\scriptsize \good{5\,/\,0\,/\,1}} & \good{9\,/\,0\,/\,0} & {\scriptsize \good{6\,/\,0\,/\,0}} \\
Zero-DCE~\cite{guo2020zerodce} & \good{7\,/\,2\,/\,0} & {\scriptsize \good{4\,/\,2\,/\,0}} & \good{7\,/\,2\,/\,0} & {\scriptsize \good{4\,/\,2\,/\,0}} & \good{8\,/\,0\,/\,1} & {\scriptsize \good{5\,/\,0\,/\,1}} \\
SCI~\cite{ma2022sci} & \good{9\,/\,0\,/\,0} & {\scriptsize \good{6\,/\,0\,/\,0}} & \good{9\,/\,0\,/\,0} & {\scriptsize \good{6\,/\,0\,/\,0}} & \good{9\,/\,0\,/\,0} & {\scriptsize \good{6\,/\,0\,/\,0}} \\
\bottomrule
\end{tabular}

\end{table*}

\section{The frontier models}
\label{app:claude}

We compare ATune with two frontier vision-language models, Claude Opus 5 and GPT-6 Astra. Each model looks at the
photograph and sets the editor's sliders itself, in one answer. \Cref{tab:h2h} compares ATune with Claude judge by
judge.

\begin{table}[h]
\centering
\caption{\textbf{ATune vs Claude Opus 5, judge by judge}, paired per photograph. Diff.: mean score of ATune's edit
minus Claude's, in each judge's own units ($\times$100 where marked). \good{Green}/\bad{red}: significant in ATune's/Claude's
favour after Holm's correction over the nine judges; grey: not significant. W: share of photographs where ATune's edit scores
higher. Last row: the counts of \maintab{tab:wtl}{3}.}
\label{tab:h2h}
\setlength{\tabcolsep}{3.4pt}\footnotesize\renewcommand{\arraystretch}{1.1}
\begin{adjustbox}{max width=\linewidth}
\begin{tabular}{@{}lccIccIcc@{}}
\toprule
 & \multicolumn{2}{c}{FiveK} & \multicolumn{2}{c}{PPR10K} & \multicolumn{2}{c}{SPAQ} \\
Judge & Diff. & W\,\% & Diff. & W\,\% & Diff. & W\,\% \\
\midrule
Q-Align & \nsig{+0.007} & 51 & \nsig{+0.025} & 55 & \good{+0.062} & 59 \\
Q-Align\textsubscript{A} & \good{+0.086} & 69 & \nsig{0.000} & 50 & \good{+0.060} & 63 \\
LIQE & \good{+0.22} & 69 & \nsig{0.00} & 50 & \good{+0.17} & 62 \\
PaQ-2-PiQ & \bad{$-$0.24} & 43 & \bad{$-$0.21} & 42 & \nsig{+0.04} & 47 \\
TOPIQ\,$\times$100 & \bad{$-$0.26} & 41 & \nsig{+0.13} & 51 & \nsig{$-$0.17} & 42 \\
LAION & \nsig{$-$0.001} & 50 & \good{+0.064} & 65 & \nsig{$-$0.007} & 47 \\
CLIP-IQA\,$\times$100 & \good{+9.44} & 89 & \good{+2.26} & 66 & \good{+3.28} & 70 \\
QualiCLIP\,$\times$100 & \good{+1.77} & 65 & \good{+1.89} & 71 & \good{+1.77} & 61 \\
DeQA & \nsig{0.000} & 55 & \nsig{+0.027} & 59 & \good{+0.063} & 62 \\
\midrule
Judges for ATune / neither / Claude & \multicolumn{2}{c}{4 / 3 / 2} & \multicolumn{2}{c}{3 / 5 / 1} & \multicolumn{2}{c}{6 / 3 / 0} \\
\bottomrule
\end{tabular}

\end{adjustbox}
\end{table}

\paragraph{Prompt.} Each model gets the photograph (the 1024\,px frame, as a lossless PNG), then this text:

\smallskip
{\setlength{\parskip}{2pt}\noindent\textbf{RawTherapee} (after the image).\par\noindent{\ttfamily\footnotesize Retouch this image as best as possible using these RawTherapee 5.8 sliders:\\
\mbox{}\\
- White Balance: Temperature (1500 to 60000 K, currently 6504 K), Tint (0.02 to 10, currently 1)\\
- Exposure: Compensation ($-$5 to 12 EV), Contrast ($-$100 to 100), Saturation ($-$100 to 100), Black ($-$16384 to 32768), Highlight compression (0 to 500)\\
- Shadows \& Highlights: Shadows (0 to 100), Highlights (0 to 100)\\
- Sharpening: Amount (1 to 1000)\\
- Local Contrast: Amount (0 to 1)\\
- Haze Removal: Strength (0 to 100)\\
- Noise Reduction: Luminance (0 to 100)\\
\mbox{}\\
Reply with a JSON object with these 13 keys: \{\textquotedbl{}temperature\textquotedbl{}: \textless{}number\textgreater{}, \textquotedbl{}tint\textquotedbl{}: \textless{}number\textgreater{}, \textquotedbl{}exposure\_compensation\textquotedbl{}: \textless{}number\textgreater{}, \textquotedbl{}contrast\textquotedbl{}: \textless{}number\textgreater{}, \textquotedbl{}saturation\textquotedbl{}: \textless{}number\textgreater{}, \textquotedbl{}black\textquotedbl{}: \textless{}number\textgreater{}, \textquotedbl{}highlight\_compression\textquotedbl{}: \textless{}number\textgreater{}, \textquotedbl{}shadows\textquotedbl{}: \textless{}number\textgreater{}, \textquotedbl{}highlights\textquotedbl{}: \textless{}number\textgreater{}, \textquotedbl{}sharpening\_amount\textquotedbl{}: \textless{}number\textgreater{}, \textquotedbl{}local\_contrast\_amount\textquotedbl{}: \textless{}number\textgreater{}, \textquotedbl{}dehaze\_strength\textquotedbl{}: \textless{}number\textgreater{}, \textquotedbl{}denoise\_luminance\textquotedbl{}: \textless{}number\textgreater{}\}}\par\medskip
\noindent\textbf{darktable} (after the image).\par\noindent{\ttfamily\footnotesize Retouch this image as best as possible using these darktable 3.8 sliders:\\
\mbox{}\\
- White balance: Red, Green, Blue channel coefficients (0 to 8 each, currently 1)\\
- Exposure: Exposure ($-$3 to 4 EV), Black level correction ($-$0.1 to 0.1)\\
- Contrast brightness saturation: Contrast ($-$1 to 1), Saturation ($-$1 to 1)\\
- Shadows and highlights: Shadows ($-$100 to 100), Highlights ($-$100 to 100), White point adjustment ($-$10 to 10)\\
- Sharpen: Amount (0 to 2)\\
- Local contrast: Detail (0\% to 500\%)\\
- Haze removal: Strength ($-$1 to 1)\\
- Denoise (non-local means): Luma (0\% to 100\%)\\
\mbox{}\\
Reply with a JSON object with these 14 keys: \{\textquotedbl{}wb\_red\textquotedbl{}: \textless{}number\textgreater{}, \textquotedbl{}wb\_green\textquotedbl{}: \textless{}number\textgreater{}, \textquotedbl{}wb\_blue\textquotedbl{}: \textless{}number\textgreater{}, \textquotedbl{}exposure\textquotedbl{}: \textless{}number\textgreater{}, \textquotedbl{}black\_level\textquotedbl{}: \textless{}number\textgreater{}, \textquotedbl{}contrast\textquotedbl{}: \textless{}number\textgreater{}, \textquotedbl{}saturation\textquotedbl{}: \textless{}number\textgreater{}, \textquotedbl{}shadows\textquotedbl{}: \textless{}number\textgreater{}, \textquotedbl{}highlights\textquotedbl{}: \textless{}number\textgreater{}, \textquotedbl{}white\_point\textquotedbl{}: \textless{}number\textgreater{}, \textquotedbl{}sharpen\_amount\textquotedbl{}: \textless{}number\textgreater{}, \textquotedbl{}local\_contrast\_detail\_percent\textquotedbl{}: \textless{}number\textgreater{}, \textquotedbl{}haze\_strength\textquotedbl{}: \textless{}number\textgreater{}, \textquotedbl{}denoise\_luma\_percent\textquotedbl{}: \textless{}number\textgreater{}\}}\par\medskip
\noindent\textbf{Settings.}
\begin{itemize}[leftmargin=*,itemsep=0pt,topsep=2pt]
  \item One call per photograph and editor. No system message.
  \item Claude Opus 5: adaptive thinking at effort max. GPT-6 Astra: reasoning effort high.
  \item At most 16,000 output tokens, returned as JSON with one key per slider.
  \item The slider values are written unchanged into the editor's own settings file.
\end{itemize}
}
\smallskip

\paragraph{Prompt design.} The prompt states the task and nothing more:
\begin{itemize}[leftmargin=*,itemsep=0pt,topsep=2pt]
  \item \textbf{The editor's own sliders.} It lists 13 of RawTherapee's sliders (14 of darktable's), named and ranged
  as in the editor. The answer goes into the editor unchanged, with no mapping of ours.
  \item \textbf{The current white balance.} Without it, both models took a lower temperature as neutral and turned
  the photographs blue in a \numPilotPhotos-photograph pilot.
  \item \textbf{No extras.} No role, hint, example value, dataset, judge or style.
\end{itemize}

\paragraph{Answers.} Claude returned a valid answer for \numClaudeValid{} of \numClaudeCalls{} photographs (median
\numClaudeSec\,s per call). GPT-6 Astra returned \numGPTValid{} (median \numGPTSec\,s); its content filter refused
\numGPTFailed{} photographs, which count as unedited. \Cref{tab:claudesliders} shows the spread of Claude's answers.

\begin{table}[h]
\centering
\caption{\textbf{Claude Opus 5's answers} over all test photographs: median and 10th--90th percentiles per slider.}
\label{tab:claudesliders}
\footnotesize
\begin{tabular}{@{}lccc@{}}
\toprule
Slider & median & p10 & p90 \\
\midrule
\texttt{temperature} & $6100$ & $5600$ & $6900$ \\
\texttt{tint} & $1.02$ & $1.00$ & $1.02$ \\
\texttt{exposure\_compensation} & $0.25$ & $-0.15$ & $0.65$ \\
\texttt{contrast} & $14.00$ & $12.00$ & $22.00$ \\
\texttt{saturation} & $10.00$ & $6.00$ & $18.00$ \\
\texttt{black} & $180$ & $-600$ & $450$ \\
\texttt{highlight\_compression} & $35$ & $25$ & $60$ \\
\texttt{shadows} & $28.00$ & $20.00$ & $55.00$ \\
\texttt{highlights} & $30.00$ & $25.00$ & $45.00$ \\
\texttt{sharpening\_amount} & $220$ & $200$ & $420$ \\
\texttt{local\_contrast\_amount} & $0.25$ & $0.18$ & $0.35$ \\
\texttt{dehaze\_strength} & $25.00$ & $12.00$ & $45.00$ \\
\texttt{denoise\_luminance} & $25.00$ & $15.00$ & $40.00$ \\
\bottomrule
\end{tabular}

\end{table}

\section{Baselines}
\label{app:baselines}

We compare with seven learned retouchers, all run with the authors' code and released weights:
\begin{itemize}[leftmargin=*,itemsep=0pt,topsep=2pt]
  \item \textbf{Trained on expert edits:} CSRNet~\cite{he2020csrnet}, AdaInt~\cite{yang2022adaint} and
  NeurOP~\cite{wang2022neurop}, with FiveK weights. On PPR10K, AdaInt and NeurOP use the authors' PPR10K weights.
  \item \textbf{Trained without paired edits:} UEGAN~\cite{ni2020uegan}, EnlightenGAN~\cite{jiang2021enlightengan},
  Zero-DCE~\cite{guo2020zerodce} and SCI~\cite{ma2022sci}.
\end{itemize}
We leave out the image-adaptive 3D LUT~\cite{zeng2020lut}: run with current versions of PyTorch, its released
weights produce corrupted images.

\paragraph{Input size.} Our test photographs are at full resolution, but most baselines were trained on much smaller
images, and some work poorly on large ones. So we run each method in two ways and keep the better one:
\begin{itemize}[leftmargin=*,itemsep=0pt,topsep=2pt]
  \item \textbf{Native.} The method edits the full photograph.
  \item \textbf{Small copy.} The method edits a copy at its training size, and we carry the edit over to the full
  photograph. For AdaInt, CSRNet and NeurOP the edit is a per-pixel colour mapping, so it carries over exactly.
  Zero-DCE and SCI predict curve or illumination maps, which we upsample, as Zero-DCE++ does. For the GANs we
  upsample the change they made.
\end{itemize}
The better way is the one with the higher score, averaged over the three test sets and the five held-out
\texttt{pyiqa} judges. We fixed this rule before scoring any baseline. CSRNet, UEGAN, EnlightenGAN and SCI run
native; AdaInt, NeurOP and Zero-DCE run on a small copy.

\paragraph{PPR10K reference.} We edit the PPR10K photographs at 1024\,px, but PPR10K's own expert edits are only
360\,px. MMArt~\cite{lin2025jarvisart} is the only release with PPR10K expert edits at 1024\,px, so we use it as the
reference.

\section{Test sets}
\label{app:sets}

We test on three public sets (\mainsec{sec:experiments}{4}). FiveK and PPR10K have expert edits to compare against; SPAQ
has phone photographs with no reference edit.
\begin{itemize}[leftmargin=*,itemsep=0pt,topsep=2pt]
  \item \textbf{FiveK.} The 500 photographs of the standard test split, as converted from the raw files by a public
  copy of FiveK on Hugging Face. One of them (a4949) is corrupt in that copy, which leaves 499. For three photographs
  the expert's edit is cropped or resized differently from the original, so we leave them out of comparisons with the
  expert.
  \item \textbf{PPR10K.} The 382 portraits of the official validation groups that MMArt~\cite{lin2025jarvisart}
  released (groups 1520--1680, from 148 photo sessions). Photographs from one session look alike, so the bootstrap
  resamples whole sessions.
  \item \textbf{SPAQ.} A fixed list of 1000 phone photographs. We rotate the
  \numSpaqRotated{} photographs stored sideways and convert the \numSpaqWideGamut{} wide-gamut files to sRGB.
\end{itemize}

\paragraph{Test photographs in training.} None of the three sets is used to train ATune. AdaInt and NeurOP train on
FiveK photographs a0001--a4500 and on PPR10K's training groups; we checked that neither contains a test photograph.
CSRNet's authors split FiveK at random, so its released weights were trained on \numCsrSeen{} of our \numCsrN{} FiveK
test photographs; the main paper marks it with $^\ddagger$. SPAQ test photographs were used to test the attribute
measures (\cref{app:attrsel}), but not to train anything.

\section{Against the expert: all methods}
\label{app:expertfull}

\begin{table*}[t]
\centering
\caption{\textbf{Against the expert} (FiveK expert C, $n{=}496$; PPR10K retoucher, $n{=}382$). Judges: median
score change, edit minus expert. \textbf{Bold}: best; first row: the original.}
\label{tab:expertfull}
\setlength{\tabcolsep}{4.2pt}\footnotesize\renewcommand{\arraystretch}{1.12}
\begin{adjustbox}{max width=\textwidth}
\begin{tabular}{@{}lccccccIcccccc@{}}
\toprule
 & \multicolumn{6}{c}{FiveK, vs expert C} & \multicolumn{6}{c}{PPR10K, vs retoucher} \\
\cmidrule(lr){2-7}\cmidrule(l){8-13}
Method & Q-Align & LIQE & PSNR$\uparrow$ & SSIM$\uparrow$ & LPIPS$\downarrow$ & $\Delta E_{00}\!\downarrow$ & Q-Align & LIQE & PSNR$\uparrow$ & SSIM$\uparrow$ & LPIPS$\downarrow$ & $\Delta E_{00}\!\downarrow$ \\
\midrule
Unedited original & $-$0.091 & $-$0.18 & 18.45 & 0.730 & 0.135 & 10.72 & +0.004 & +0.01 & 20.73 & 0.932 & 0.101 & 7.97 \\
\midrule
\textbf{ATune} (10M, $\tau{=}0.99$) & \textbf{+0.078} & \textbf{+0.43} & 18.68 & 0.709 & 0.164 & 10.79 & \textbf{+0.129} & \textbf{+0.34} & 19.50 & \textbf{0.926} & 0.126 & 9.64 \\
\hdashline
Claude Opus 5, same editor & +0.074 & +0.24 & 18.77 & 0.737 & 0.132 & 10.40 & +0.104 & +0.31 & 18.44 & 0.896 & \textbf{0.112} & 10.35 \\
GPT-6 Astra, same editor & +0.031 & +0.13 & 19.41 & 0.746 & 0.123 & 9.92 & +0.082 & +0.23 & 18.25 & 0.914 & 0.124 & 11.09 \\
\hdashline
CSRNet~\cite{he2020csrnet}$^\ddagger$ & \bad{$-$0.199} & \bad{$-$0.52} & 15.80 & 0.719 & 0.173 & 15.40 & \bad{$-$0.415} & \bad{$-$0.58} & 15.50 & 0.878 & 0.243 & 16.35 \\
AdaInt~\cite{yang2022adaint}$^\dagger$ & +0.023 & +0.04 & 19.74 & 0.766 & 0.093 & 9.18 & +0.082 & +0.11 & 17.31 & 0.907 & 0.152 & 11.68 \\
NeurOP~\cite{wang2022neurop}$^\dagger$ & +0.027 & +0.02 & \textbf{20.43} & \textbf{0.768} & \textbf{0.089} & \textbf{8.77} & +0.055 & +0.12 & 18.11 & 0.911 & 0.143 & 10.72 \\
\hdashline
UEGAN~\cite{ni2020uegan} & +0.028 & +0.04 & 19.02 & 0.744 & 0.107 & 10.14 & +0.039 & \nsig{+0.02} & \textbf{19.97} & 0.899 & \textbf{0.112} & \textbf{9.04} \\
EnlightenGAN~\cite{jiang2021enlightengan} & \bad{$-$0.407} & \bad{$-$0.61} & 11.25 & 0.605 & 0.203 & 23.07 & \bad{$-$0.224} & \bad{$-$0.22} & 15.10 & 0.853 & 0.176 & 14.64 \\
Zero-DCE~\cite{guo2020zerodce} & \bad{$-$0.428} & \bad{$-$0.25} & 13.14 & 0.646 & 0.186 & 18.63 & \bad{$-$0.176} & \nsig{+0.01} & 15.86 & 0.861 & 0.192 & 14.92 \\
SCI~\cite{ma2022sci} & \bad{$-$1.282} & \bad{$-$1.00} & 7.52 & 0.518 & 0.386 & 31.97 & \bad{$-$1.745} & \bad{$-$0.93} & 9.92 & 0.661 & 0.431 & 22.58 \\
\bottomrule
\end{tabular}

\end{adjustbox}
\end{table*}

A judge can reward an edit that looks nothing like what an expert would do. So we also compare every edit with the
expert's own edit of the same photograph, in two ways (\cref{tab:expertfull}):
\begin{itemize}[leftmargin=*,itemsep=0pt,topsep=2pt]
  \item \textbf{Pixel distance.} How far each edit is from the expert's, by PSNR, SSIM, LPIPS and
  CIEDE2000~\cite{wang2004ssim,zhang2018lpips,sharma2005ciede2000}. On FiveK, ATune's edits are about as far from
  expert C as the original is (PSNR $\numExpOrigPSNR\to\numExpOursPSNR$\,dB, CIEDE2000
  $\numExpOrigDE\to\numExpOursDE$). LPIPS rises ($\numExpOrigLPIPS\to\numExpOursLPIPS$) because ATune sharpens and
  the expert did not. The frontier models come a little closer, and NeurOP closest (\numExpNeurOPPSNR\,dB), as
  expected: it was trained to copy expert C. On PPR10K, every method ends further from the retoucher's pixels than
  the original. The retoucher's work is local, and global sliders cannot reproduce it.
  \item \textbf{Judge scores.} A judge scores both the edit and the expert's edit; we report the difference. A
  positive number means the judge rates the edit above the expert's. On FiveK, the
  original scores \numExpOrigQAFiveK{} below expert C under \qalign, and ATune's edits score \numExpOursQAFiveK{}
  above it. On PPR10K, ATune scores \numExpOursQAPPR{} above the retoucher (GPT-6 Astra \numExpGPTQAPPR). ATune is
  not alone here: both frontier models and three baselines also score above the expert under \qalign.
\end{itemize}
A judge's preference and an expert's choice are different targets, so we report both.

\section{Human study}
\label{app:human}

\paragraph{Setup.} In each trial, a rater saw two edits of the same photograph side by side, in random order: one by
ATune and one by another method. The rater answered ``Which image is better?'' with Image~1, Image~2 or About the same.
We compared ATune with three methods: Claude Opus 5, GPT-6 Astra, and the stronger supervised retoucher on each set,
which we chose in advance by \qalign{} (NeurOP on FiveK and AdaInt on PPR10K). The photographs were drawn at random
from each test set, and every pair was rated by at least five people. The images were JPEGs at quality 95, 2048\,px on
the long side, except PPR10K's, which kept their own 1024\,px.

\paragraph{Raters.} The raters came from a commercial crowdsourcing platform that admits only experienced
contributors, and were paid US\$15 for a session of 60 pairs. Each agreed to the platform's consent notice first, and
we replaced their account identifiers with random numbers before analysis. The platform marked \numHumInvalid{}
answers as invalid; without them, \numHumVotes{} votes from \numHumRaters{} raters remain. No IRB reviewed the study,
because it collected only preferences between photographs from public research datasets and no personal
information, and we judged it to pose minimal risk.

\paragraph{Analysis.} For each comparison we report the share of votes that went to ATune, leaving out the About the
same answers. Counting those as half a vote for each side instead would change no share by more than
\numHumTieShift{} points. Confidence intervals come from a bootstrap that resamples both raters and
photographs~\cite{owen2007pigeonhole}. Against Claude, ATune's share over all three sets has a 95\% CI of
\mbox{\numHumClaudeAllCI\%}, close to 50\%, so we treat the two as equally preferred~\cite{lakens2017equivalence}.
The comparisons with GPT-6 Astra and the supervised retouchers are corrected for multiple testing with Holm's
method. \Cref{tab:humanfull} gives every result.

\begin{table}[h]
\centering
\caption{\textbf{All human-study results.} ATune: its share of votes, ties left out. Ties as $\frac12$: the same
share with each tie split evenly. $p$: p-value for ATune's share differing from 50\%, before Holm's correction. $^\dagger$NeurOP on FiveK, AdaInt on PPR10K.}
\label{tab:humanfull}
\setlength{\tabcolsep}{3pt}\footnotesize
\begin{adjustbox}{max width=\linewidth}
\begin{tabular}{@{}llrrccc@{}}
\toprule
ATune vs & Set & Answers & Ties (\%) & ATune (\%), 95\% CI & Ties as $\frac12$ (\%) & $p$ \\
\midrule
Claude Opus 5 & FiveK & 1{,}345 & 8.3 & 44.2 (38.4--49.8) & 44.7 & 0.044 \\
 & PPR10K & 861 & 7.4 & 56.1 (49.3--62.8) & 55.6 & 0.078 \\
 & SPAQ & 1{,}325 & 17.1 & 51.2 (45.3--57.1) & 51.0 & 0.687 \\
 & pooled & 3{,}531 & 11.4 & 50.5 (46.8--54.2) & 50.4 & 0.791 \\
\midrule
GPT-6 Astra & FiveK & 1{,}333 & 7.0 & 54.0 (48.1--59.7) & 53.7 & 0.186 \\
 & PPR10K & 864 & 8.8 & 54.1 (47.0--61.3) & 53.7 & 0.272 \\
 & SPAQ & 1{,}327 & 15.5 & 55.0 (48.9--60.7) & 54.2 & 0.112 \\
 & pooled & 3{,}524 & 10.6 & 54.3 (50.3--58.3) & 53.9 & 0.031 \\
\midrule
NeurOP\,/\,AdaInt$^\dagger$ & FiveK & 1{,}338 & 7.1 & 47.2 (41.5--53.0) & 47.4 & 0.347 \\
 & PPR10K & 860 & 4.3 & 45.3 (38.3--52.6) & 45.5 & 0.205 \\
 & pooled & 2{,}198 & 6.0 & 46.3 (41.5--51.2) & 46.5 & 0.137 \\
\bottomrule
\end{tabular}

\end{adjustbox}
\end{table}

\paragraph{Judges and raters.} We also checked how often each judge agrees with people, following
DEAR~\cite{plohotnuk2025dear}. A judge agrees on a pair when it gives the higher score to the edit most raters chose.
For comparison, we measure how often a single rater agrees with the majority of the other raters
(\cref{tab:humanjudges}).

\begin{table}[h]
\centering
\caption{\textbf{How often each judge picks the raters' choice} (\% of pairs). Last row: one rater against the
others.}
\label{tab:humanjudges}
\setlength{\tabcolsep}{3pt}\footnotesize
\begin{adjustbox}{max width=\linewidth}
\begin{tabular}{@{}lcccc@{}}
\toprule
Judge & vs Claude & vs GPT-6 & vs NeurOP\,/\,AdaInt & All (95\% CI) \\
\midrule
Q-Align & 63.6 & 63.6 & 65.2 & 64.0 (61.5--66.4) \\
Q-Align\textsubscript{A} & 54.7 & 60.3 & 53.2 & 56.5 (53.9--59.0) \\
LIQE & 57.6 & 64.1 & 51.6 & 58.6 (55.9--61.3) \\
PaQ-2-PiQ & 51.9 & 56.1 & 50.1 & 53.1 (50.5--55.6) \\
TOPIQ & 56.2 & 59.5 & 55.7 & 57.3 (54.7--59.8) \\
LAION & 55.9 & 53.7 & 52.1 & 54.1 (51.4--56.8) \\
CLIP-IQA & 51.1 & 58.1 & 49.6 & 53.4 (50.7--56.1) \\
QualiCLIP & 59.8 & 62.7 & 51.3 & 58.9 (56.3--61.4) \\
DeQA & 59.9 & 61.8 & 65.3 & 61.9 (59.4--64.5) \\
\midrule
Vote of the nine judges & 61.6 & 64.5 & 55.7 & 61.3 (58.7--63.8) \\
\midrule
One rater vs the others' majority & & & & 67.3 (64.7--69.8) \\
\bottomrule
\end{tabular}

\end{adjustbox}
\end{table}

\paragraph{Sharpening.} ATune's edits are often sharper than the other method's, so we looked at whether this
sways raters and judges differently. We split the pairs into thirds by how much sharper ATune's edit is and compare
the two in each third (\cref{tab:humansharp}). We chose this analysis after seeing the votes.

\begin{table}[h]
\centering
\caption{\textbf{Raters and judges, by how much sharper ATune's edit is.} Pairs split into thirds by the sharpness
difference, ATune minus the other method, within each set. Raters: ATune's share of votes, ties left out, with 95\% CI.
Judges: share of pairs on which more of the nine judges favour ATune.}
\label{tab:humansharp}
\setlength{\tabcolsep}{3pt}\footnotesize
\begin{adjustbox}{max width=\linewidth}
\begin{tabular}{@{}llccc@{}}
\toprule
ATune vs & Third ($n$ pairs) & Median $\Delta$ & Raters (\%), 95\% CI & Judges (\% pairs) \\
\midrule
Claude Opus 5 & least (234) & $-$0.12 & 49.0 (43.0--55.0) & 39.3 \\
 & middle (232) & +0.28 & 54.7 (48.6--60.9) & 75.0 \\
 & most (234) & +0.74 & 45.4 (39.1--51.8) & 76.5 \\
\midrule
GPT-6 Astra & least (234) & +0.06 & 52.0 (45.7--58.2) & 56.0 \\
 & middle (232) & +0.45 & 58.5 (52.2--64.6) & 86.2 \\
 & most (234) & +0.94 & 52.5 (46.0--58.9) & 84.2 \\
\midrule
NeurOP\,/\,AdaInt$^\dagger$ & least (146) & +0.12 & 39.8 (32.7--47.2) & 60.3 \\
 & middle (145) & +0.44 & 50.0 (42.3--57.5) & 86.2 \\
 & most (146) & +0.72 & 49.8 (42.0--57.7) & 91.8 \\
\bottomrule
\end{tabular}

\end{adjustbox}
\end{table}

\paragraph{Limitations.} The study ran on an external platform, so we could not control viewing conditions, and
the platform has no attention checks; it screens raters by experience instead.

\section{Two exploratory analyses}
\label{app:analysis}

We ran two further analyses after seeing the results. They suggest explanations but do not prove them.

\begin{table}[t]
\centering
\caption{\textbf{ATune vs the frontier models, by how good the original is.} Each test set is split into thirds
by the original's \qalign{} score. Each cell: judges significantly preferring ATune\,/\,neither\,/\,the other model.}
\label{tab:quality}
\setlength{\tabcolsep}{5pt}\footnotesize
\begin{tabular}{@{}llcc@{}}
\toprule
Set & Third of originals ($n$) & vs Claude Opus 5 & vs GPT-6 Astra \\
\midrule
FiveK & weakest (163) & 4\,/\,4\,/\,1 & 9\,/\,0\,/\,0 \\
 & middle (169) & 4\,/\,5\,/\,0 & 8\,/\,1\,/\,0 \\
 & strongest (167) & 2\,/\,6\,/\,1 & 6\,/\,3\,/\,0 \\
\midrule
PPR10K & weakest (126) & 6\,/\,3\,/\,0 & 8\,/\,1\,/\,0 \\
 & middle (123) & 4\,/\,4\,/\,1 & 7\,/\,2\,/\,0 \\
 & strongest (133) & 1\,/\,7\,/\,1 & 4\,/\,4\,/\,1 \\
\midrule
SPAQ & weakest (333) & 4\,/\,5\,/\,0 & 6\,/\,3\,/\,0 \\
 & middle (333) & 6\,/\,3\,/\,0 & 9\,/\,0\,/\,0 \\
 & strongest (334) & 4\,/\,2\,/\,3 & 7\,/\,2\,/\,0 \\
\bottomrule
\end{tabular}

\end{table}

\paragraph{Does ATune help weak photographs more?} We split each test set into thirds by the original's
\qalign{} score, from weakest to strongest, and compare ATune with each frontier model in each third
(\cref{tab:quality}). Against Claude, ATune's lead is smallest on the strongest photographs, on every set. On FiveK, \numQLowFiveKOurs{} judges favour ATune and \numQLowFiveKClaude{} Claude in the weakest third,
against \numQHighFiveKOurs{} to \numQHighFiveKClaude{} in the strongest. On PPR10K it is \numQLowPPROurs{} to
\numQLowPPRClaude, against an even \numQHighPPROurs{} to \numQHighPPRClaude; on SPAQ, \numQLowSPAQOurs{} to
\numQLowSPAQClaude, against \numQHighSPAQOurs{} to \numQHighSPAQClaude. Against GPT-6 Astra, more judges favour ATune
in every third. This fits the idea that ATune learned what weak photographs need, while strong photographs have less
to gain.

\paragraph{What makes the judges prefer one edit?} ATune's and Claude's edits of the same photograph often get
different scores (averaged over the six held-out judges). To see why, we predict the score difference from the differences in the 13 attributes, with a
linear fit. The fit explains $R^2=\numGapRsqFiveK$, $\numGapRsqPPR$ and $\numGapRsqSPAQ$ of the score difference on
FiveK, PPR10K and SPAQ.
\begin{itemize}[leftmargin=*,itemsep=0pt,topsep=2pt]
  \item \textbf{Sharpness matters most.} The sharper edit tends to score higher ($r=\numGapSharpRFiveK$,
  $\numGapSharpRPPR$ and $\numGapSharpRSPAQ$). Sharpness is the strongest attribute in the fit on FiveK and PPR10K;
  on SPAQ, \numGapTopSPAQ{} is.
  \item \textbf{Highlights hardly matter.} Claude often pulls the highlights back so nothing clips. This barely
  changes the scores ($|r|\le\numGapClipR$).
\end{itemize}
Sharper edits please the judges but are not closer to the expert: against expert C, our edits' LPIPS rises from
$\numExpOrigLPIPS$ to $\numExpOursLPIPS$, while PSNR and colour error change little (\cref{tab:expertfull}).

\section{Training-set size: every judge}
\label{app:scaling}

\begin{figure*}[t]
\centering
\includegraphics[width=\textwidth]{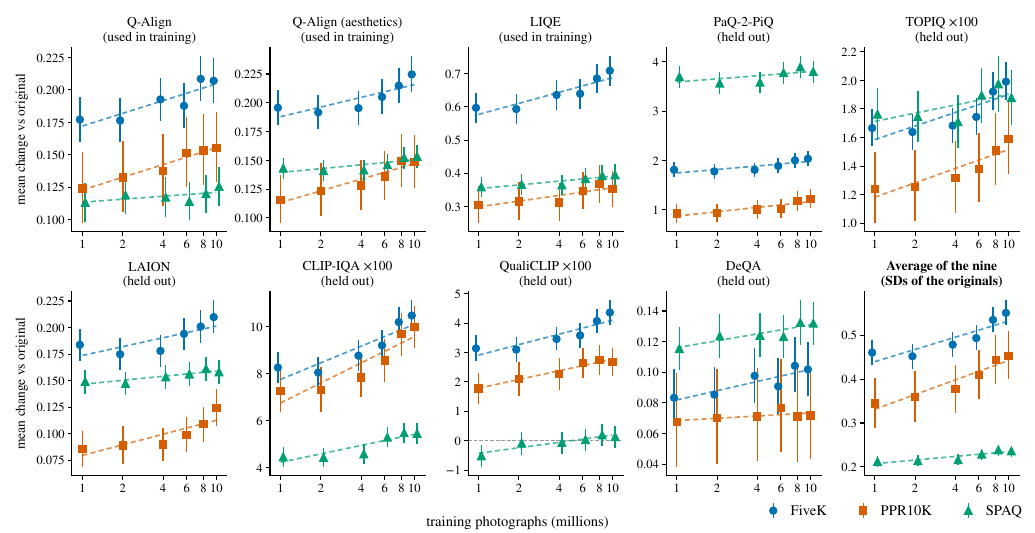}
\caption{\textbf{Every judge against training-set size.} Mean score change against the original, with 95\% CI, for
the six models of \mainsec{sec:ablations}{4.5}; the size axis is logarithmic. Dashed: a straight-line fit. Last panel: the average of the nine judges, each scaled by its spread on the originals.}
\label{fig:scaling}
\end{figure*}

\Cref{fig:scaling} shows, for each judge, how the score change grows with the number of training photographs,
from 1M to \numPaperSize. Neighbouring sizes have overlapping intervals here, so this plot cannot tell them apart. The
paired comparison in \maintab{tab:ablation}{5}a can, because it compares the two models on the same photographs.

\section{Robustness: judges given the full-size image}
\label{app:robust}

Three judges, PaQ-2-PiQ, CLIP-IQA and QualiCLIP, come with an input size set by their authors. In the main paper,
every judge sees the image at its authors' size. To see how much this matters, we also gave these three judges the
full-size image (\cref{tab:native}). Their scores change a lot. On FiveK, PaQ-2-PiQ's score gain for ATune's edits
grows from $\numPaqDesignFiveK$ to $\numPaqNativeFiveK$, enough to reverse its ranking of ATune and Claude. CLIP-IQA
and QualiCLIP also change with the image size. Because a judge's score depends on the size it sees, the main paper
always uses each judge's own size.

\begin{table}[h]
\centering
\caption{\textbf{Three judges at their authors' input size (design) and on the full-size image (native).} Median
score change against the original. \textbf{Bold}: largest gain; \underline{underlined}: second largest, per set and column.}
\label{tab:native}
\setlength{\tabcolsep}{2.6pt}\scriptsize
\begin{adjustbox}{max width=\linewidth}
\begin{tabular}{@{}lcccccc@{}}
\toprule
 & \multicolumn{2}{c}{PaQ-2-PiQ} & \multicolumn{2}{c}{CLIP-IQA\,$\times$100} & \multicolumn{2}{c}{QualiCLIP\,$\times$100} \\
 & design & native & design & native & design & native \\
\midrule
\multicolumn{7}{l}{\textit{FiveK}} \\
\textbf{ATune} (10M, $\tau{=}0.99$) & +1.78 & \textbf{+6.99} & \textbf{+9.72} & \textbf{+11.02} & \textbf{+4.08} & \underline{+3.36} \\
Claude Opus 5, same editor & \underline{+1.94} & +2.33 & +1.07 & \underline{+6.01} & \underline{+2.67} & +1.96 \\
GPT-6 Astra, same editor & +1.31 & +1.13 & \bad{$-$1.14} & +2.75 & +2.17 & +1.03 \\
CSRNet~\cite{he2020csrnet}$^\ddagger$ & \textbf{+2.57} & \underline{+3.56} & \nsig{+0.13} & +5.11 & \bad{$-$4.53} & \bad{$-$5.02} \\
AdaInt~\cite{yang2022adaint}$^\dagger$ & +1.87 & +1.79 & +2.35 & +3.57 & +0.84 & \bad{$-$1.60} \\
NeurOP~\cite{wang2022neurop}$^\dagger$ & +1.66 & +1.45 & +2.34 & +3.17 & +0.80 & \bad{$-$2.00} \\
UEGAN~\cite{ni2020uegan} & +1.46 & +1.59 & \underline{+2.99} & +1.80 & +0.74 & \bad{$-$0.57} \\
EnlightenGAN~\cite{jiang2021enlightengan} & +1.73 & \nsig{$-$0.16} & \bad{$-$4.39} & +1.38 & \bad{$-$2.75} & \textbf{+4.04} \\
Zero-DCE~\cite{guo2020zerodce} & +1.76 & +1.35 & \bad{$-$1.31} & +2.41 & \bad{$-$0.76} & \bad{$-$0.39} \\
SCI~\cite{ma2022sci} & \bad{$-$1.74} & \bad{$-$0.71} & \bad{$-$9.95} & +4.32 & \bad{$-$8.94} & \bad{$-$4.45} \\
\multicolumn{7}{l}{\textit{PPR10K}} \\
\textbf{ATune} (10M, $\tau{=}0.99$) & +1.02 & +1.21 & \textbf{+10.99} & \textbf{+10.91} & \textbf{+2.77} & \textbf{+0.76} \\
Claude Opus 5, same editor & \underline{+1.17} & \underline{+1.36} & \underline{+7.66} & \underline{+10.87} & \underline{+0.69} & \nsig{+0.09} \\
GPT-6 Astra, same editor & +0.68 & +0.80 & +3.41 & +6.07 & +0.53 & \nsig{\underline{+0.43}} \\
CSRNet~\cite{he2020csrnet}$^\ddagger$ & \textbf{+1.80} & \textbf{+2.00} & \bad{$-$1.08} & \bad{$-$1.63} & \bad{$-$7.80} & \bad{$-$11.09} \\
AdaInt~\cite{yang2022adaint}$^\dagger$ & +0.53 & +0.52 & +1.06 & +1.10 & +0.21 & \bad{$-$1.86} \\
NeurOP~\cite{wang2022neurop}$^\dagger$ & +0.25 & +0.31 & \nsig{$-$0.26} & \nsig{$-$0.23} & \bad{$-$0.17} & \bad{$-$1.86} \\
UEGAN~\cite{ni2020uegan} & +0.89 & +0.96 & +3.28 & +4.40 & \bad{$-$0.37} & \nsig{$-$0.11} \\
EnlightenGAN~\cite{jiang2021enlightengan} & +1.02 & +1.32 & \bad{$-$3.14} & \bad{$-$2.25} & \bad{$-$1.73} & \bad{$-$1.44} \\
Zero-DCE~\cite{guo2020zerodce} & +1.11 & +1.25 & \nsig{+0.11} & \nsig{$-$0.16} & \bad{$-$0.99} & \bad{$-$1.61} \\
SCI~\cite{ma2022sci} & \bad{$-$2.43} & \bad{$-$2.08} & \bad{$-$9.47} & \bad{$-$6.85} & \bad{$-$18.85} & \bad{$-$20.58} \\
\multicolumn{7}{l}{\textit{SPAQ}} \\
\textbf{ATune} (10M, $\tau{=}0.99$) & +2.83 & \textbf{+9.70} & \textbf{+4.23} & \textbf{+6.37} & \nsig{0.00} & +1.67 \\
Claude Opus 5, same editor & \textbf{+3.21} & \underline{+4.68} & +1.09 & +3.02 & \bad{$-$1.36} & \bad{$-$0.32} \\
GPT-6 Astra, same editor & +1.96 & +2.40 & +0.32 & +1.05 & \bad{$-$0.81} & \bad{$-$0.37} \\
CSRNet~\cite{he2020csrnet}$^\ddagger$ & \underline{+3.17} & +4.01 & +1.87 & \nsig{+0.29} & \bad{$-$5.54} & \bad{$-$1.53} \\
AdaInt~\cite{yang2022adaint}$^\dagger$ & +2.20 & +2.51 & +3.24 & +0.90 & \nsig{\underline{+0.06}} & \bad{$-$0.26} \\
NeurOP~\cite{wang2022neurop}$^\dagger$ & +2.17 & +2.59 & \underline{+3.39} & +0.79 & \nsig{$-$0.05} & \bad{$-$0.28} \\
UEGAN~\cite{ni2020uegan} & +1.54 & +2.08 & +3.04 & +3.06 & \textbf{+1.12} & +0.43 \\
EnlightenGAN~\cite{jiang2021enlightengan} & +1.80 & \bad{$-$0.30} & \nsig{$-$0.16} & +3.52 & \bad{$-$3.06} & \textbf{+3.57} \\
Zero-DCE~\cite{guo2020zerodce} & +2.36 & +4.11 & +0.91 & +3.30 & \bad{$-$1.14} & \underline{+1.96} \\
SCI~\cite{ma2022sci} & \bad{$-$1.55} & +3.77 & \bad{$-$1.50} & \underline{+5.67} & \bad{$-$12.65} & \bad{$-$1.09} \\
\bottomrule
\end{tabular}

\end{adjustbox}
\end{table}

\section{Qualitative examples}
\label{app:qual}

\begin{figure*}[p]
\centering\setlength{\tabcolsep}{1.5pt}
\begin{tabular}{@{}p{0.242\linewidth}@{\hspace{2pt}}p{0.242\linewidth}@{\hspace{12pt}}p{0.242\linewidth}@{\hspace{2pt}}p{0.242\linewidth}@{}}
\centering\footnotesize Original & \centering\footnotesize ATune & \centering\footnotesize Original & \centering\footnotesize ATune \tabularnewline[-1pt]
\centering \adjustbox{cframe=black!30 0.3pt}{\includegraphics[width=\dimexpr\linewidth-0.6pt\relax,height=6.6cm,keepaspectratio]{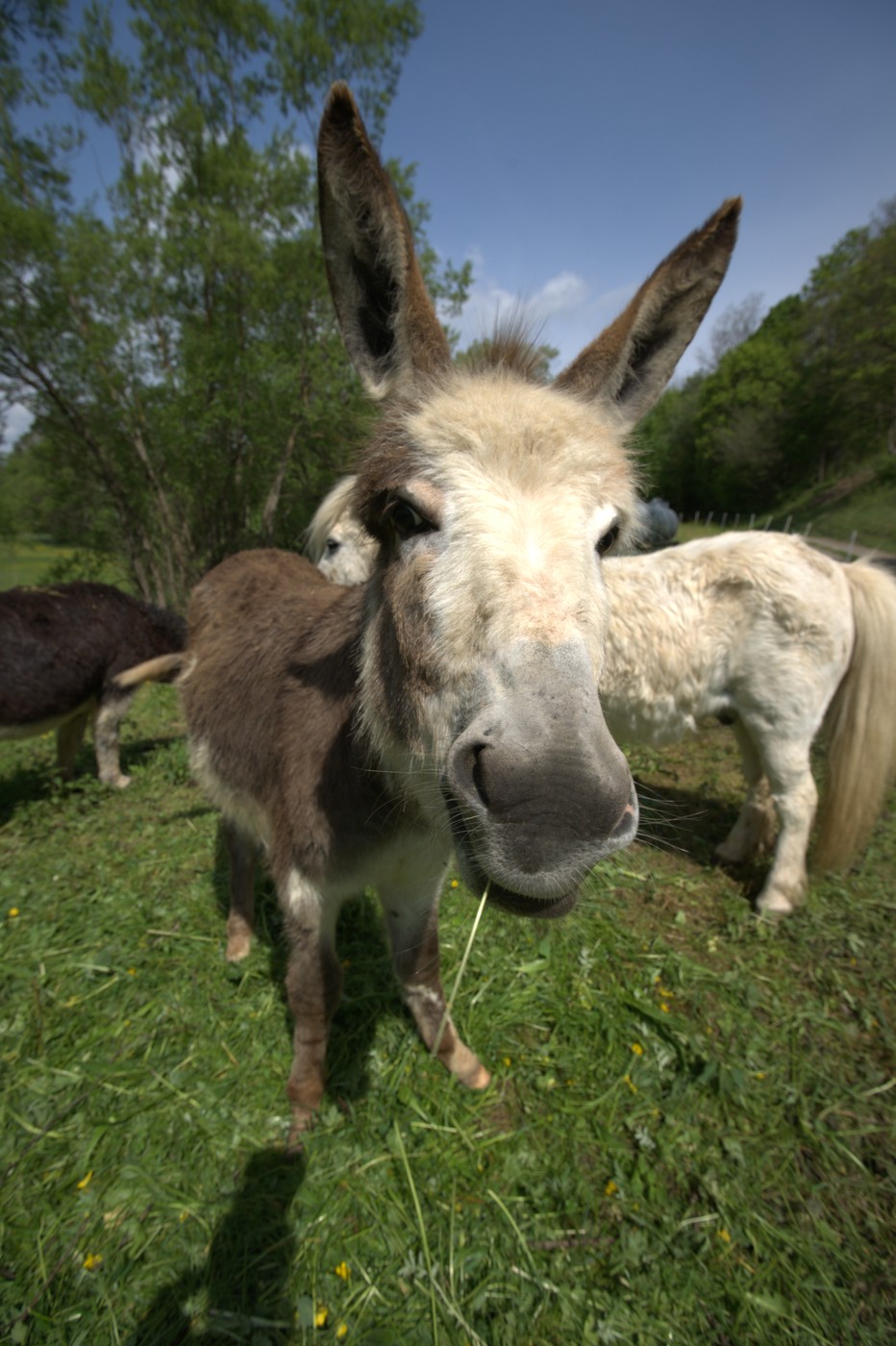}} & \centering \adjustbox{cframe=black!30 0.3pt}{\includegraphics[width=\dimexpr\linewidth-0.6pt\relax,height=6.6cm,keepaspectratio]{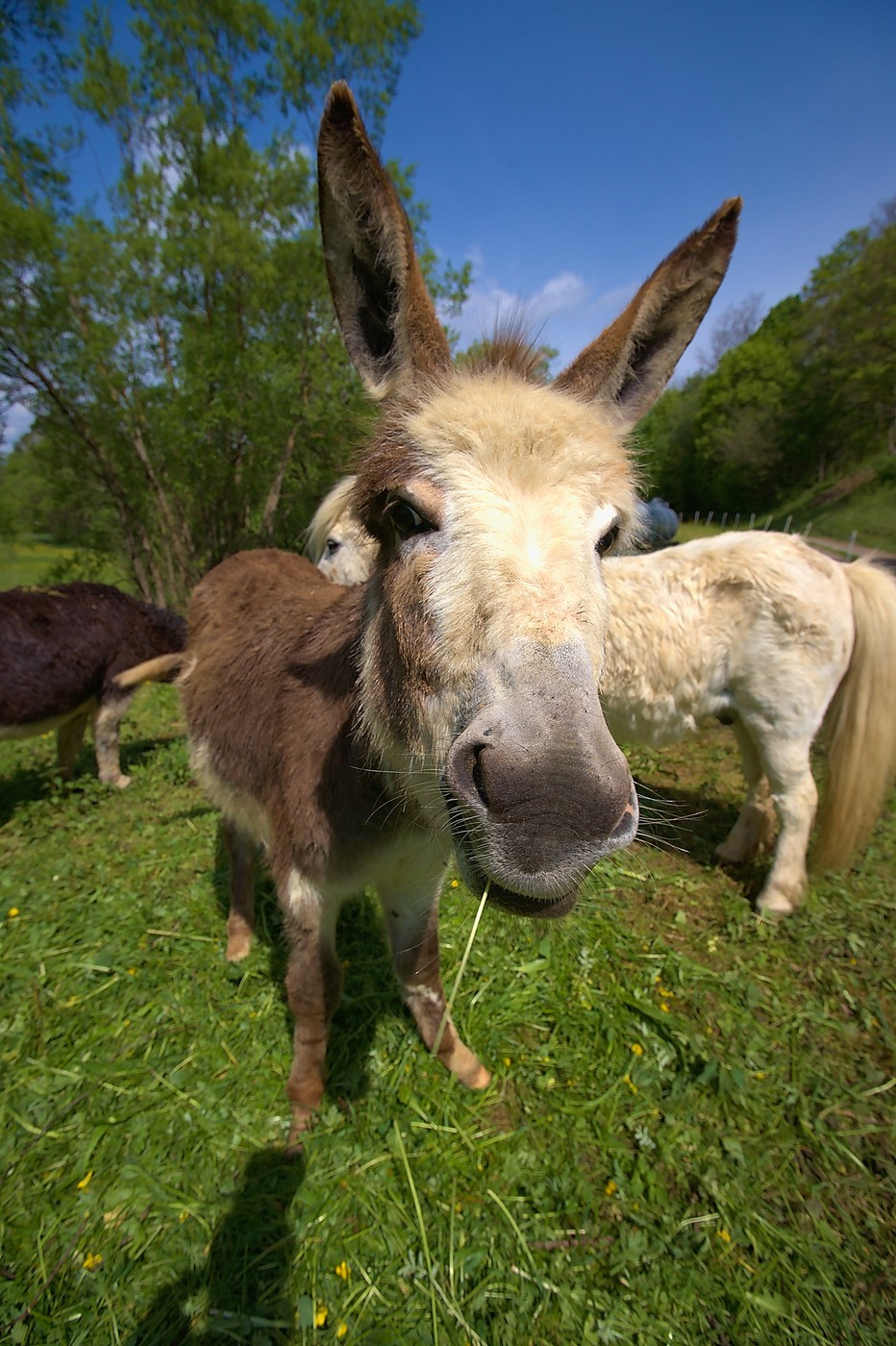}} & \centering \adjustbox{cframe=black!30 0.3pt}{\includegraphics[width=\dimexpr\linewidth-0.6pt\relax,height=6.6cm,keepaspectratio]{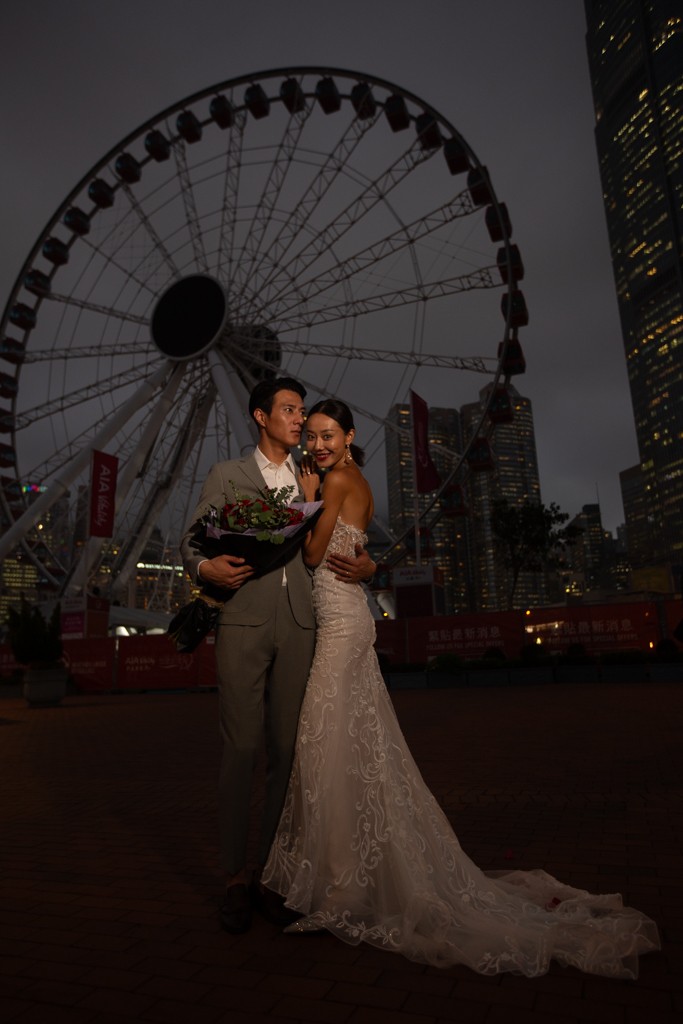}} & \centering \adjustbox{cframe=black!30 0.3pt}{\includegraphics[width=\dimexpr\linewidth-0.6pt\relax,height=6.6cm,keepaspectratio]{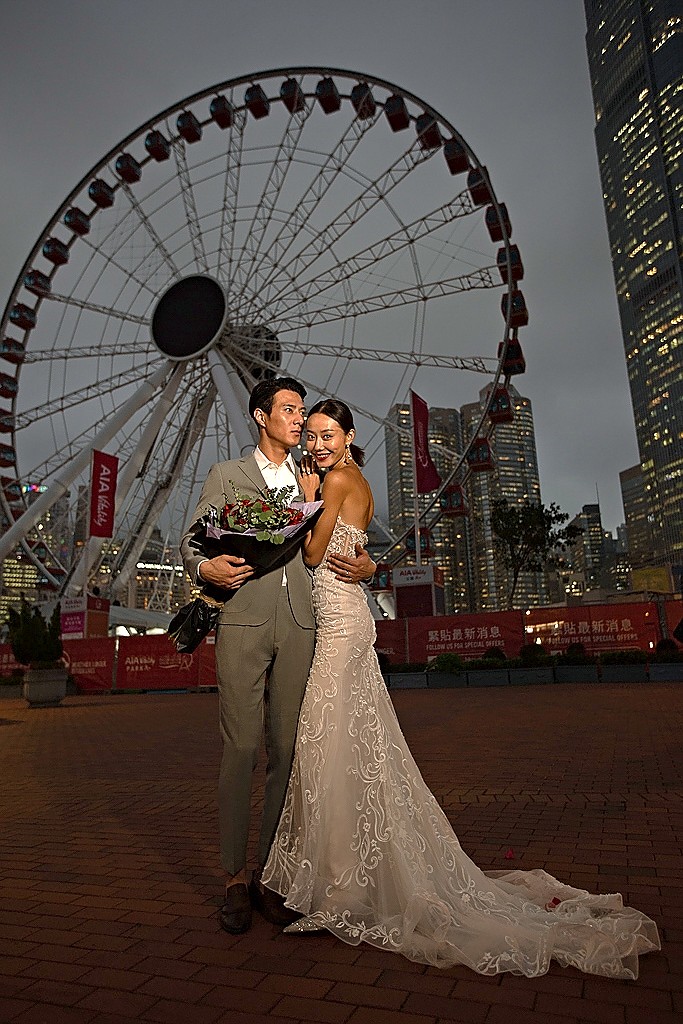}} \tabularnewline
\end{tabular}\\[6pt]
\begin{tabular}{@{}p{0.242\linewidth}@{\hspace{2pt}}p{0.242\linewidth}@{\hspace{12pt}}p{0.242\linewidth}@{\hspace{2pt}}p{0.242\linewidth}@{}}
\centering\footnotesize Original & \centering\footnotesize ATune & \centering\footnotesize Original & \centering\footnotesize ATune \tabularnewline[-1pt]
\centering \adjustbox{cframe=black!30 0.3pt}{\includegraphics[width=\dimexpr\linewidth-0.6pt\relax,height=6.6cm,keepaspectratio]{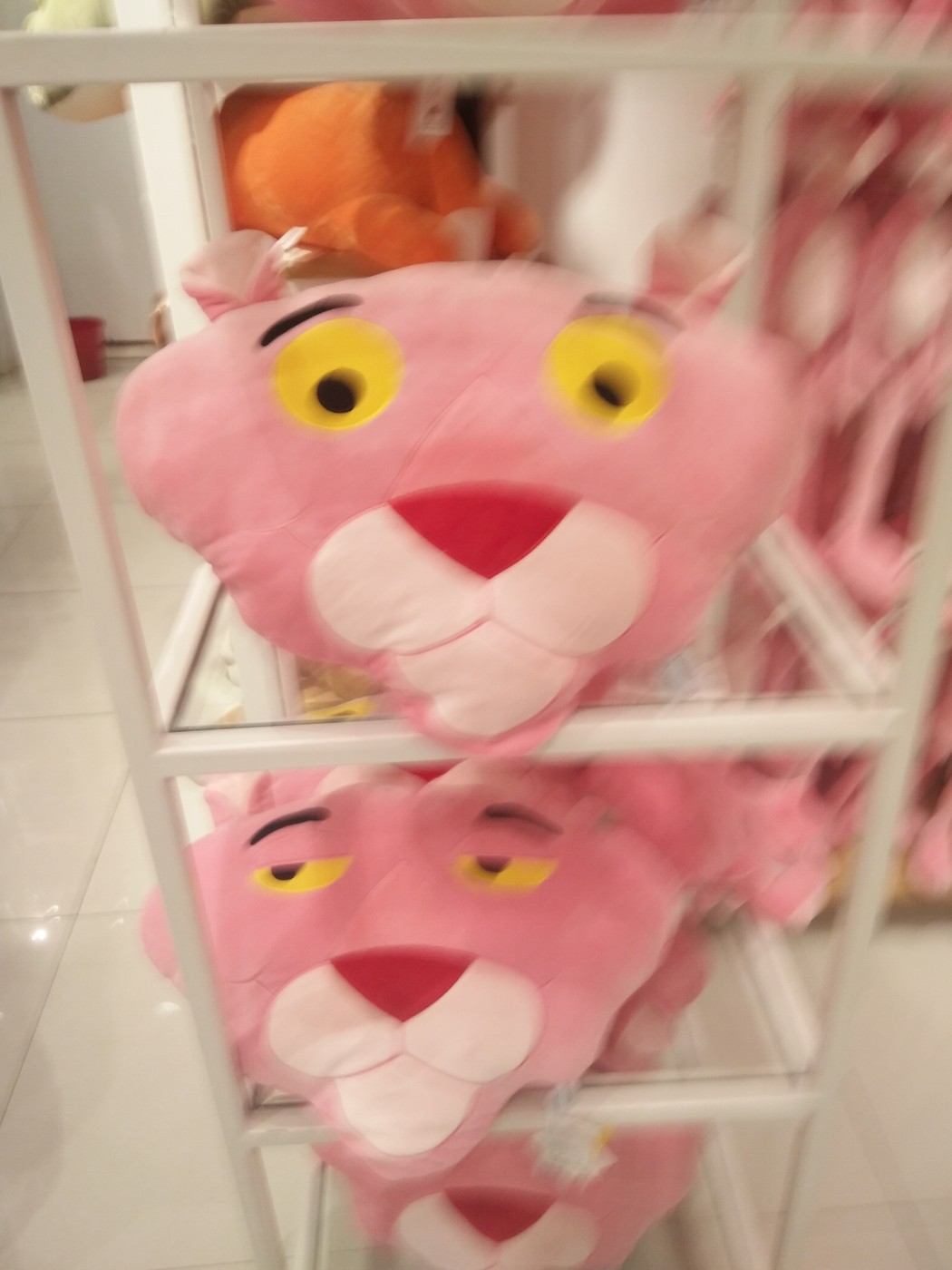}} & \centering \adjustbox{cframe=black!30 0.3pt}{\includegraphics[width=\dimexpr\linewidth-0.6pt\relax,height=6.6cm,keepaspectratio]{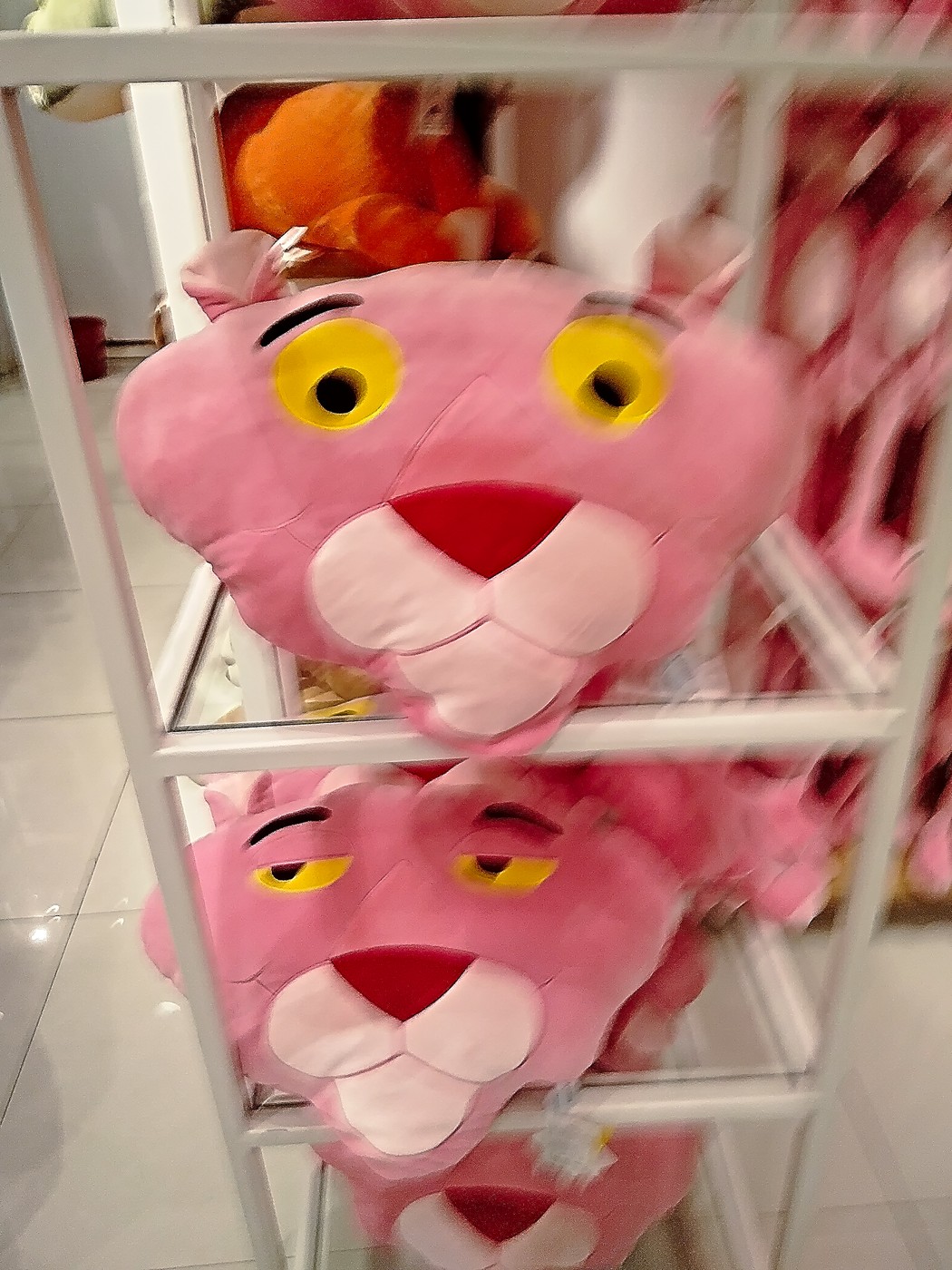}} & \centering \adjustbox{cframe=black!30 0.3pt}{\includegraphics[width=\dimexpr\linewidth-0.6pt\relax,height=6.6cm,keepaspectratio]{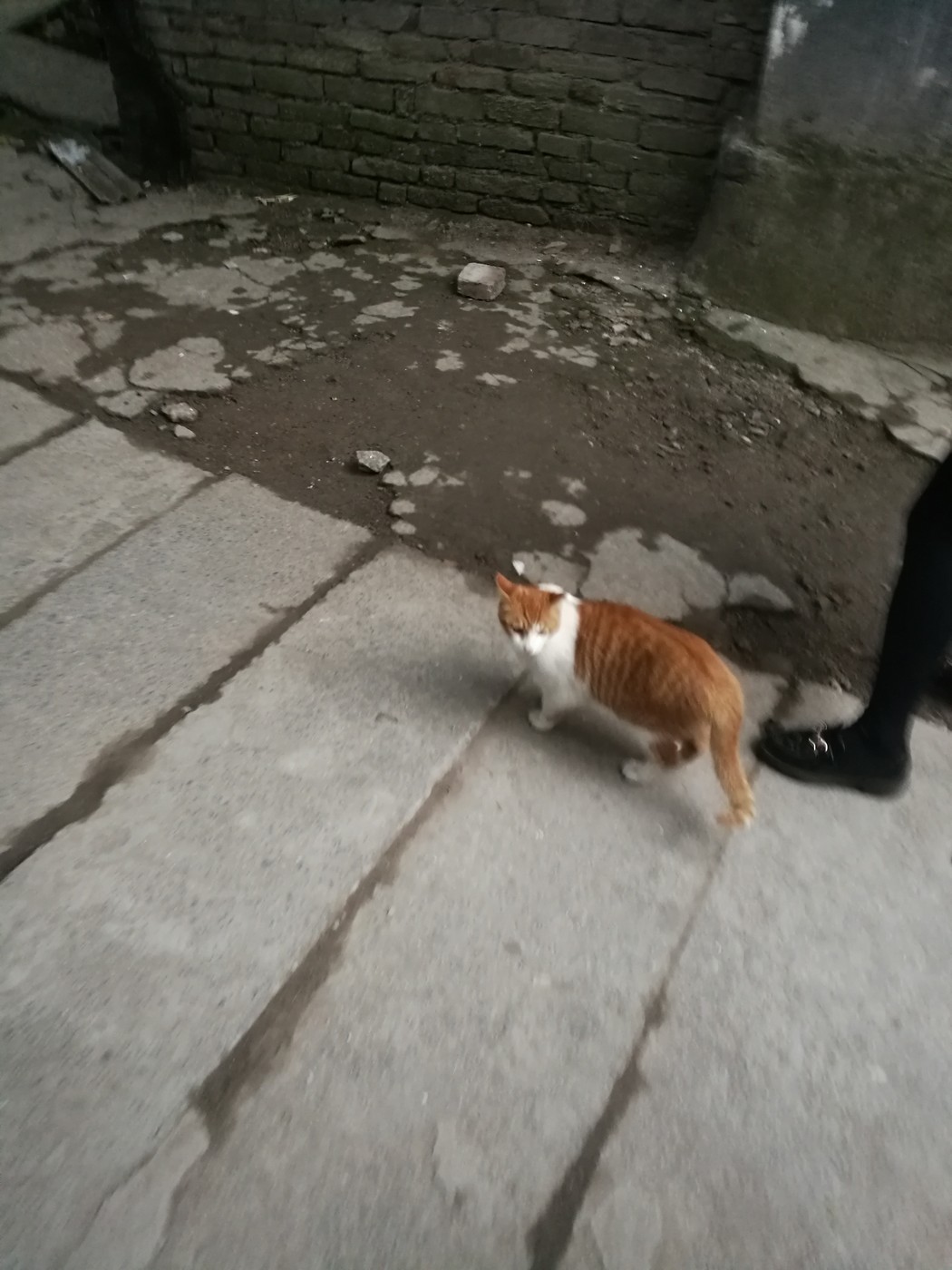}} & \centering \adjustbox{cframe=black!30 0.3pt}{\includegraphics[width=\dimexpr\linewidth-0.6pt\relax,height=6.6cm,keepaspectratio]{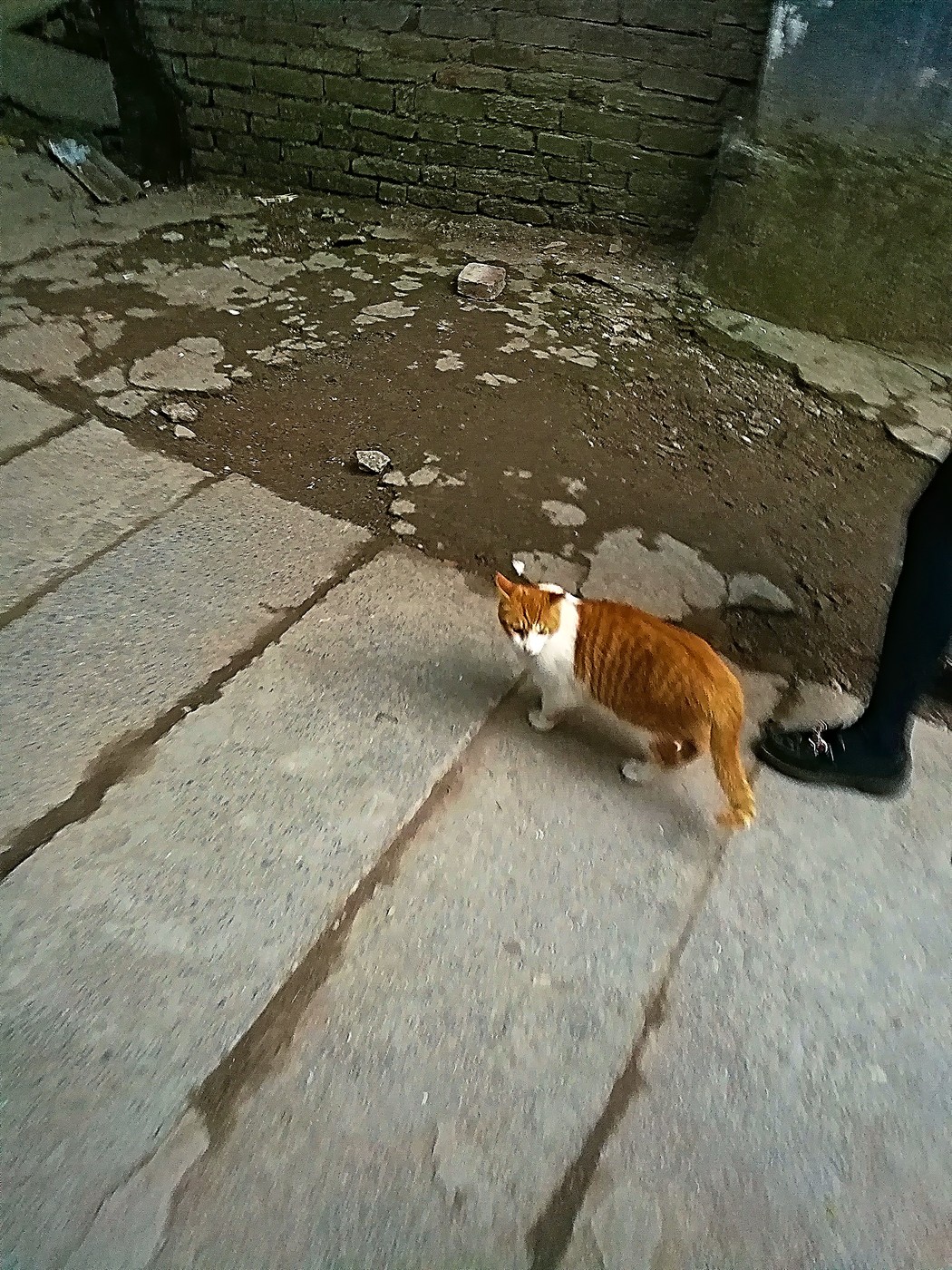}} \tabularnewline
\end{tabular}\\[6pt]
\begin{tabular}{@{}p{0.242\linewidth}@{\hspace{2pt}}p{0.242\linewidth}@{\hspace{12pt}}p{0.242\linewidth}@{\hspace{2pt}}p{0.242\linewidth}@{}}
\centering\footnotesize Original & \centering\footnotesize ATune & \centering\footnotesize Original & \centering\footnotesize ATune \tabularnewline[-1pt]
\centering \adjustbox{cframe=black!30 0.3pt}{\includegraphics[width=\dimexpr\linewidth-0.6pt\relax,height=6.6cm,keepaspectratio]{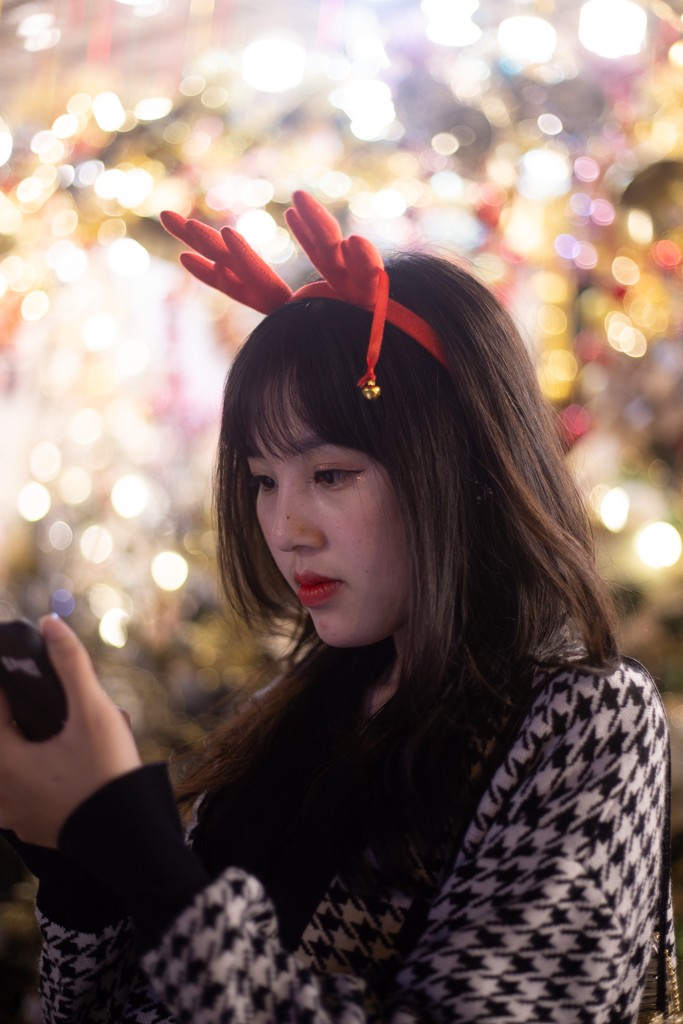}} & \centering \adjustbox{cframe=black!30 0.3pt}{\includegraphics[width=\dimexpr\linewidth-0.6pt\relax,height=6.6cm,keepaspectratio]{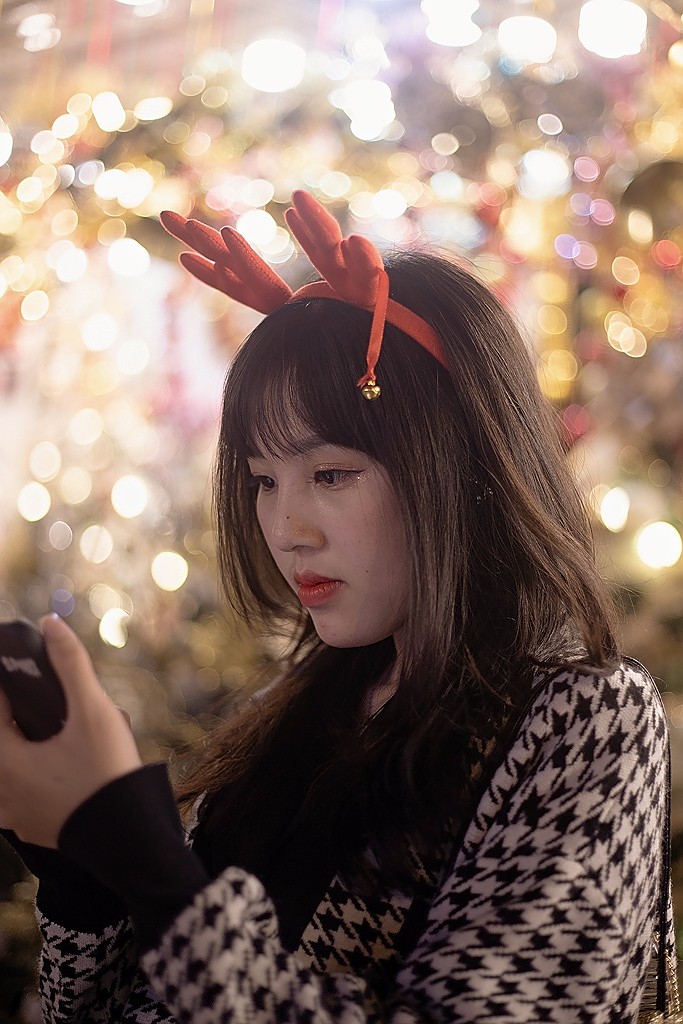}} &  &  \tabularnewline
\end{tabular}
\caption{\textbf{Selected before-and-after examples,} chosen for variety among photographs where the judges prefer ATune's edit to Claude's. \Cref{fig:qual_fivek,fig:qual_ppr10k,fig:qual_spaq} show examples picked by a fixed rule.}
\label{fig:showcase}
\end{figure*}
\begin{figure*}[p]
\centering\setlength{\tabcolsep}{1.5pt}
\begin{tabular}{@{}p{0.350\linewidth}@{\hspace{2pt}}p{0.350\linewidth}@{}}
\centering\footnotesize Original & \centering\footnotesize ATune \tabularnewline[-1pt]
\centering \adjustbox{cframe=black!30 0.3pt}{\includegraphics[width=\dimexpr\linewidth-0.6pt\relax,height=4.5cm,keepaspectratio]{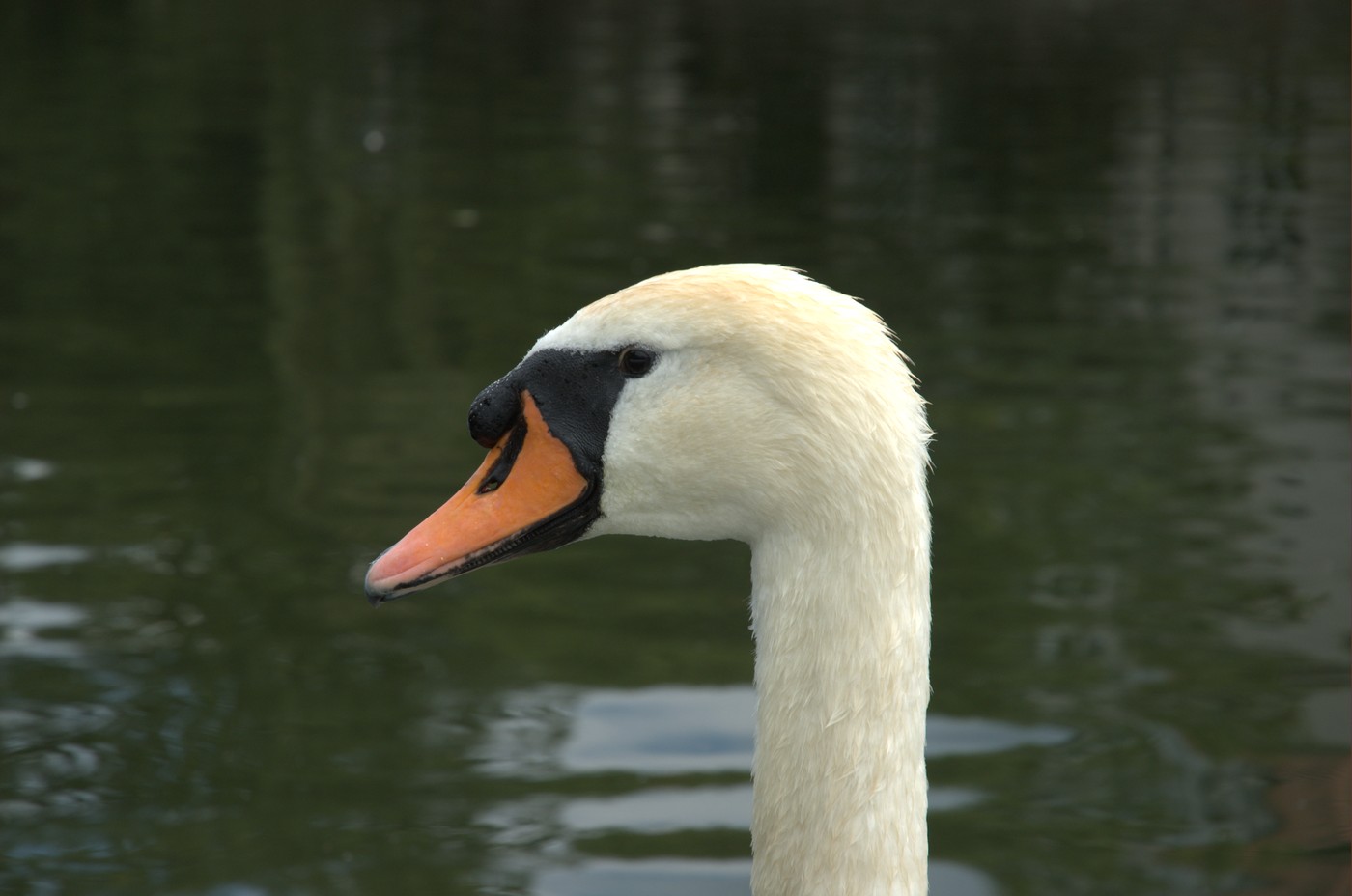}} & \centering \adjustbox{cframe=black!30 0.3pt}{\includegraphics[width=\dimexpr\linewidth-0.6pt\relax,height=4.5cm,keepaspectratio]{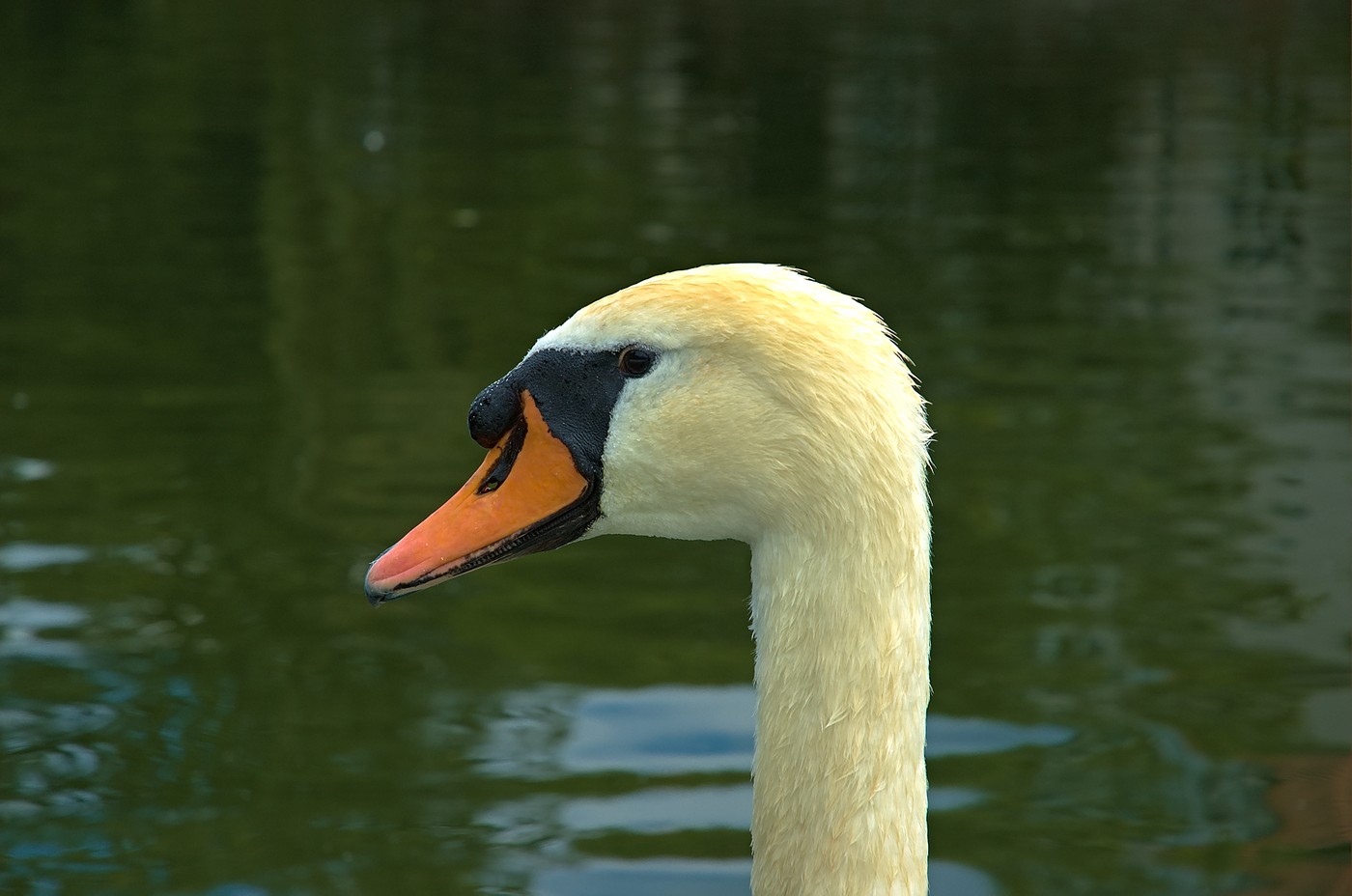}} \tabularnewline
\end{tabular}\\[6pt]
\begin{tabular}{@{}p{0.350\linewidth}@{\hspace{2pt}}p{0.350\linewidth}@{}}
\centering\footnotesize Original & \centering\footnotesize ATune \tabularnewline[-1pt]
\centering \adjustbox{cframe=black!30 0.3pt}{\includegraphics[width=\dimexpr\linewidth-0.6pt\relax,height=4.5cm,keepaspectratio]{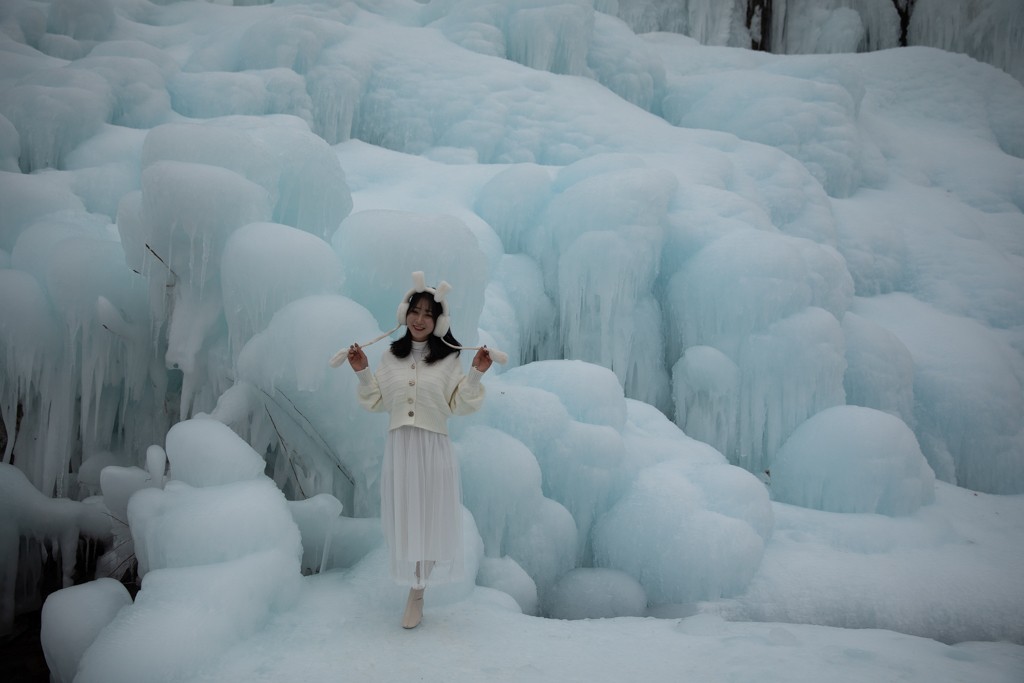}} & \centering \adjustbox{cframe=black!30 0.3pt}{\includegraphics[width=\dimexpr\linewidth-0.6pt\relax,height=4.5cm,keepaspectratio]{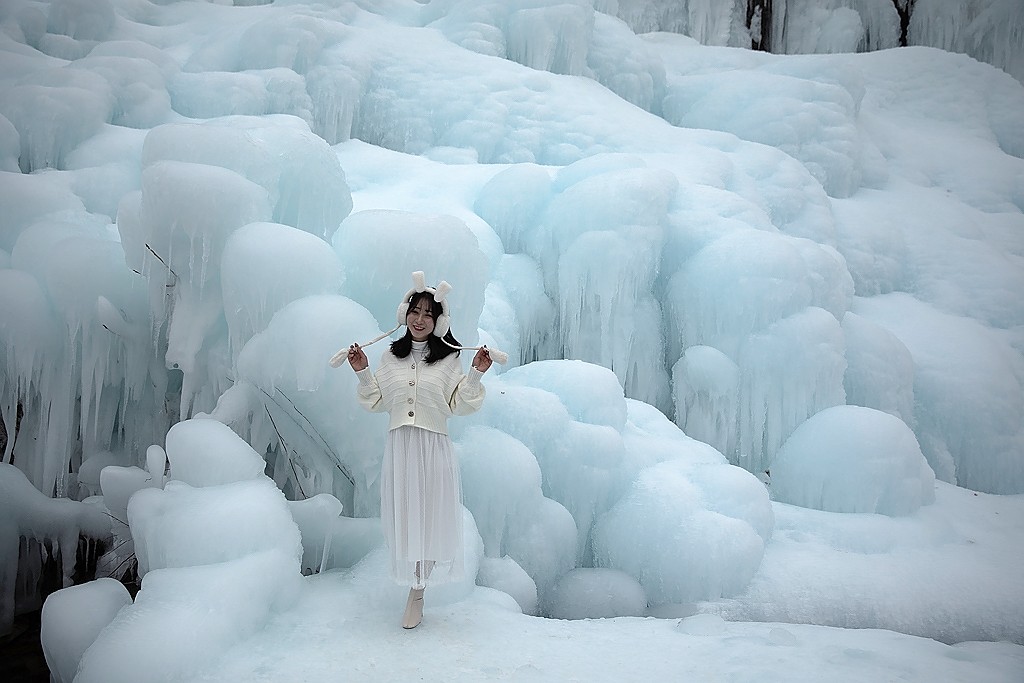}} \tabularnewline
\end{tabular}
\caption{\textbf{Selected before-and-after examples (continued).}}
\label{fig:showcase2}
\end{figure*}

\begin{figure*}[p]
\centering
\setlength{\tabcolsep}{1.5pt}
\begin{tabular}{@{}m{0.035\linewidth}m{0.233\linewidth}m{0.233\linewidth}m{0.233\linewidth}m{0.233\linewidth}@{}}
 & \centering\small Original & \centering\small ATune & \centering\small Claude Opus 5 & \centering\small Expert C \tabularnewline[2pt]
\rotatebox{90}{\parbox{3.35cm}{\centering\scriptsize Prefer ours\\(90th pct.)}} & \centering \adjustbox{cframe=black!25 0.3pt}{\includegraphics[width=\dimexpr\linewidth-0.6pt\relax,height=3.35cm,keepaspectratio]{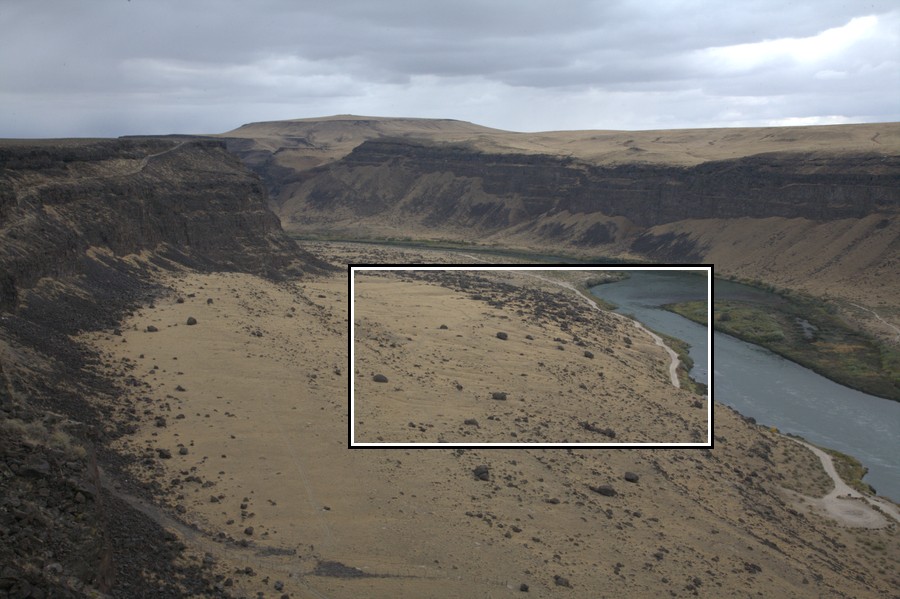}}\\[1.5pt]\adjustbox{cframe=black!25 0.3pt}{\includegraphics[width=\dimexpr\linewidth-0.6pt\relax]{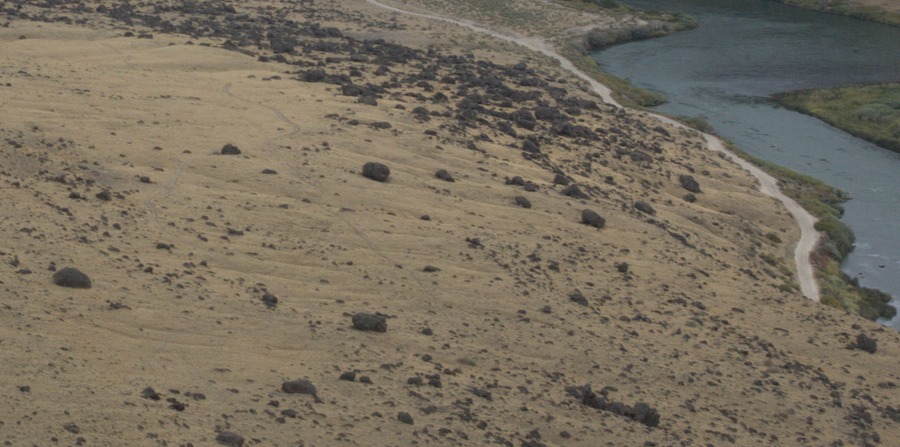}} & \centering \begin{tikzpicture}[baseline=(im.center)]\node[inner sep=0] (im) {\adjustbox{cframe=black!25 0.3pt}{\includegraphics[width=\dimexpr\linewidth-0.6pt\relax,height=3.35cm,keepaspectratio]{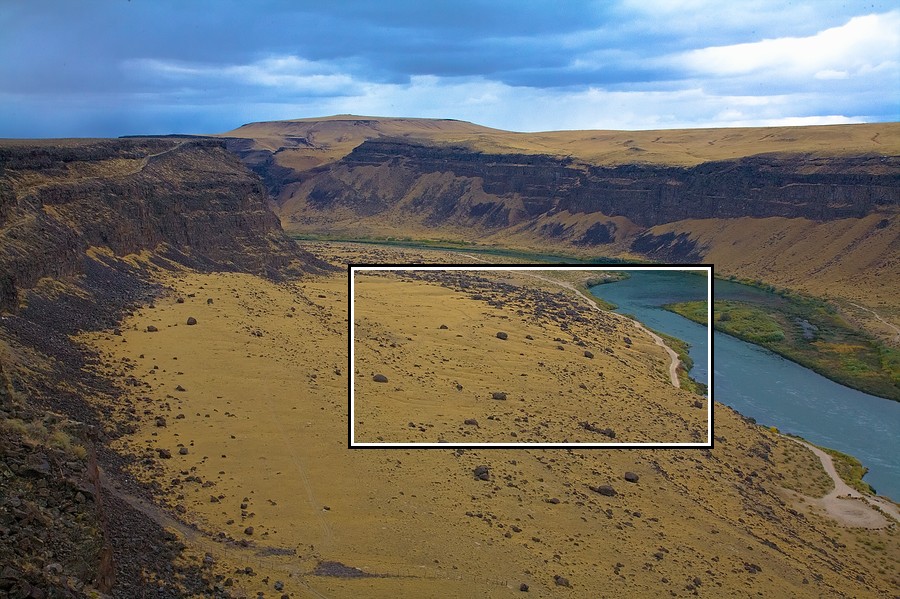}}};\node[anchor=south east, inner sep=1.2pt, fill=posgreen, text=white, font=\scriptsize\bfseries, rounded corners=0.8pt] at ([xshift=-2pt, yshift=2pt]im.south east) {9/9 judges};\end{tikzpicture}\\[1.5pt]\adjustbox{cframe=black!25 0.3pt}{\includegraphics[width=\dimexpr\linewidth-0.6pt\relax]{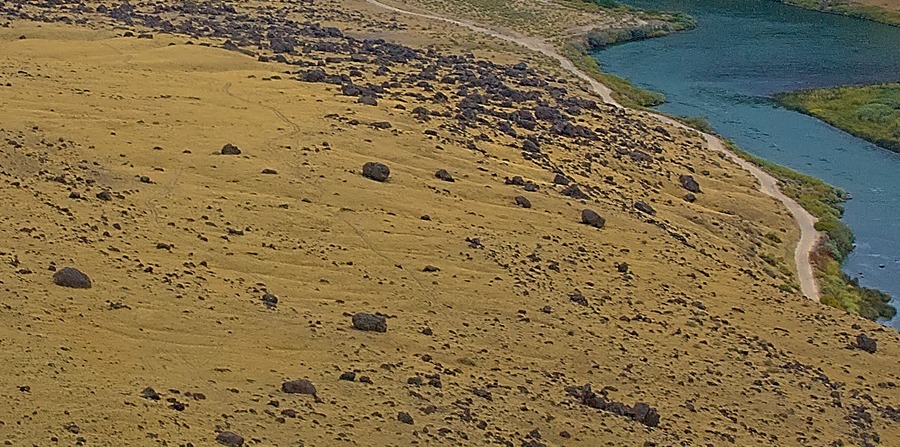}} & \centering \begin{tikzpicture}[baseline=(im.center)]\node[inner sep=0] (im) {\adjustbox{cframe=black!25 0.3pt}{\includegraphics[width=\dimexpr\linewidth-0.6pt\relax,height=3.35cm,keepaspectratio]{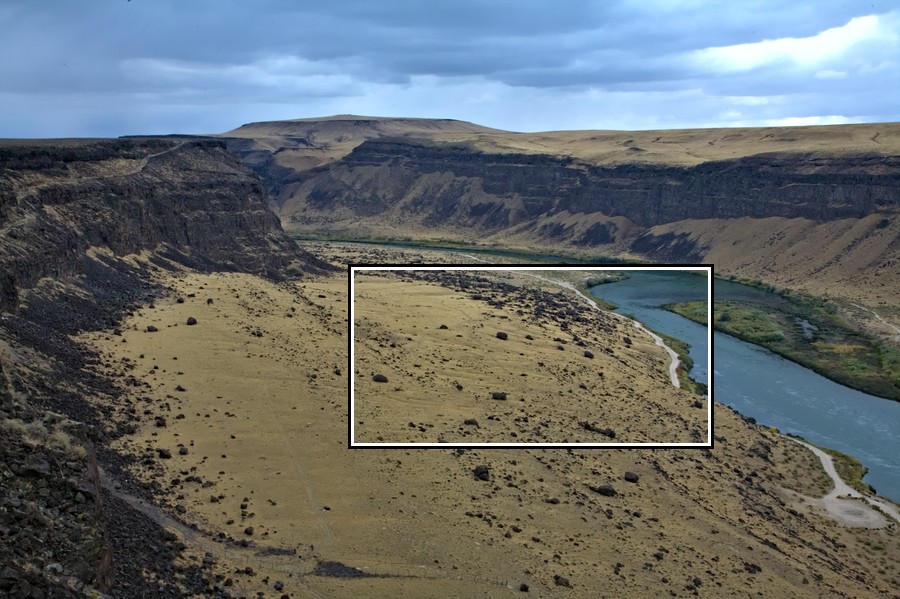}}};\node[anchor=south east, inner sep=1.2pt, fill=black!55, text=white, font=\scriptsize\bfseries, rounded corners=0.8pt] at ([xshift=-2pt, yshift=2pt]im.south east) {0/9 judges};\end{tikzpicture}\\[1.5pt]\adjustbox{cframe=black!25 0.3pt}{\includegraphics[width=\dimexpr\linewidth-0.6pt\relax]{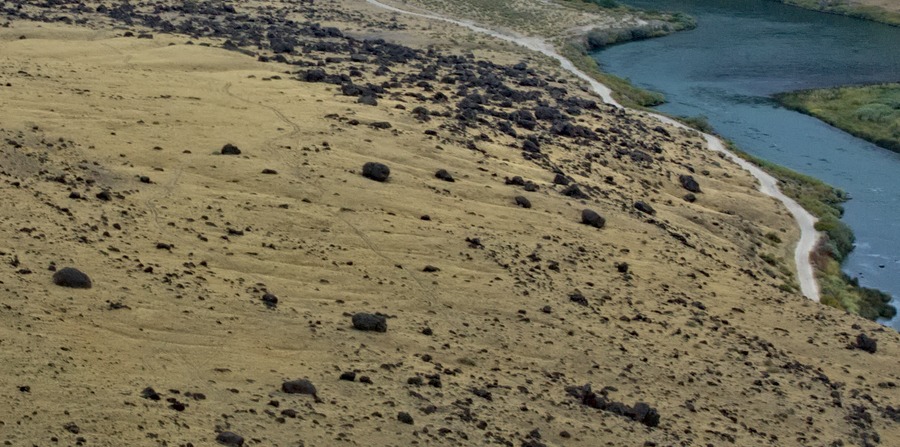}} & \centering \adjustbox{cframe=black!25 0.3pt}{\includegraphics[width=\dimexpr\linewidth-0.6pt\relax,height=3.35cm,keepaspectratio]{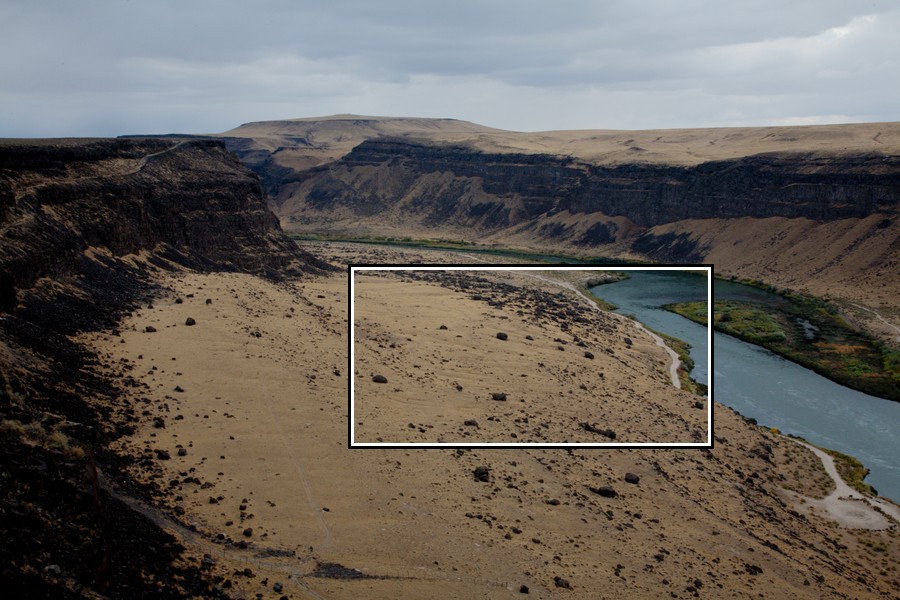}}\\[1.5pt]\adjustbox{cframe=black!25 0.3pt}{\includegraphics[width=\dimexpr\linewidth-0.6pt\relax]{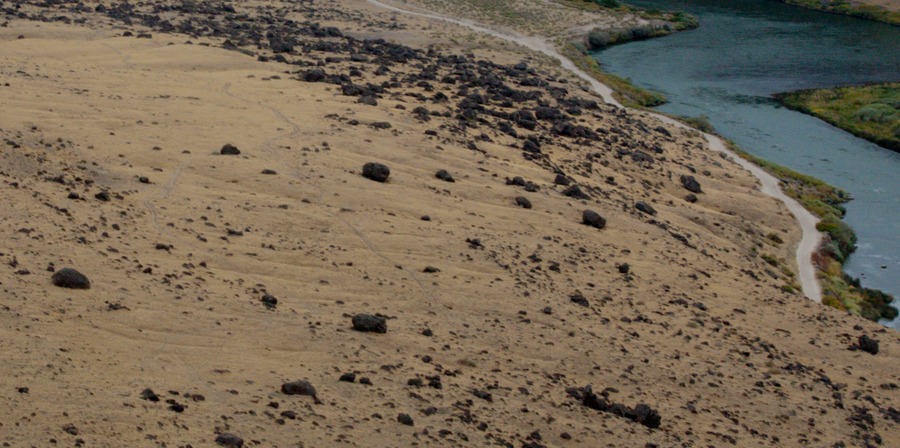}} \tabularnewline[7pt]
\rotatebox{90}{\parbox{3.35cm}{\centering\scriptsize Typical\\(median)}} & \centering \adjustbox{cframe=black!25 0.3pt}{\includegraphics[width=\dimexpr\linewidth-0.6pt\relax,height=3.35cm,keepaspectratio]{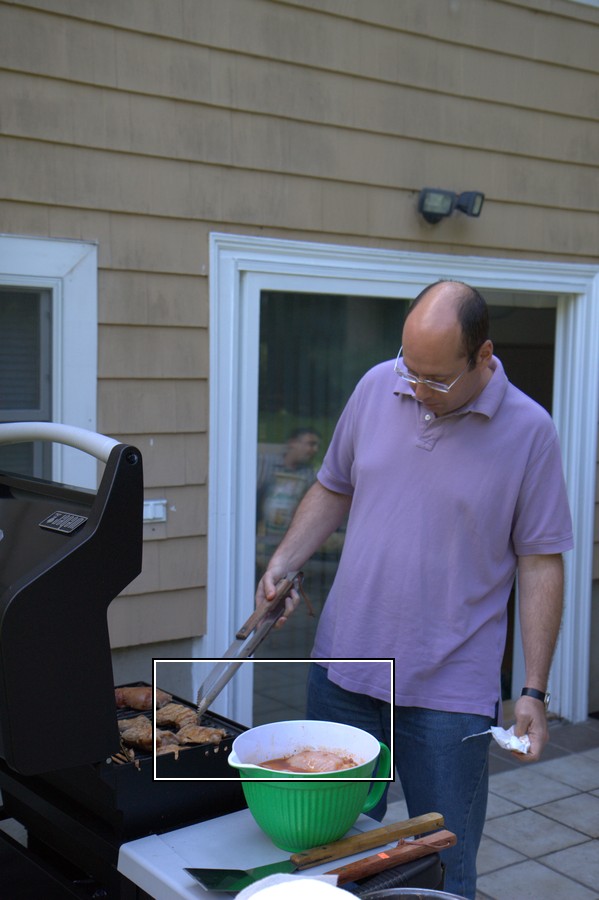}}\\[1.5pt]\adjustbox{cframe=black!25 0.3pt}{\includegraphics[width=\dimexpr\linewidth-0.6pt\relax]{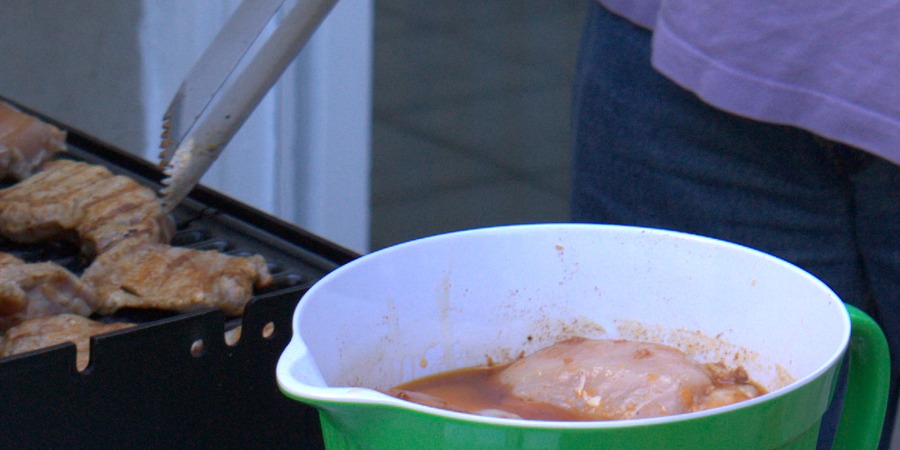}} & \centering \begin{tikzpicture}[baseline=(im.center)]\node[inner sep=0] (im) {\adjustbox{cframe=black!25 0.3pt}{\includegraphics[width=\dimexpr\linewidth-0.6pt\relax,height=3.35cm,keepaspectratio]{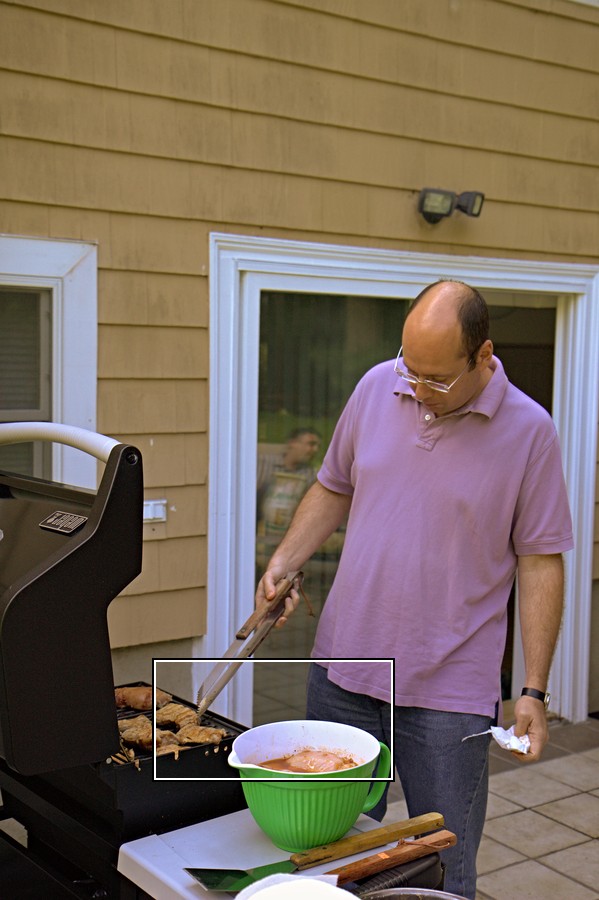}}};\node[anchor=south east, inner sep=1.2pt, fill=posgreen, text=white, font=\scriptsize\bfseries, rounded corners=0.8pt] at ([xshift=-2pt, yshift=2pt]im.south east) {6/9 judges};\end{tikzpicture}\\[1.5pt]\adjustbox{cframe=black!25 0.3pt}{\includegraphics[width=\dimexpr\linewidth-0.6pt\relax]{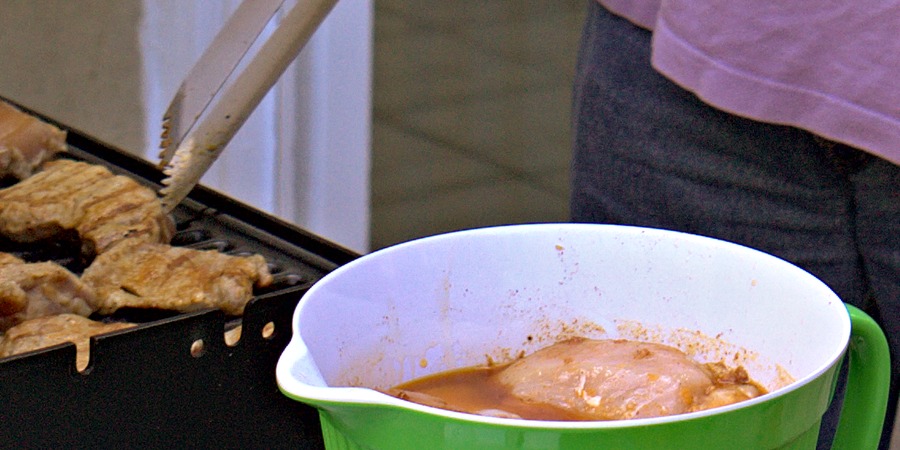}} & \centering \begin{tikzpicture}[baseline=(im.center)]\node[inner sep=0] (im) {\adjustbox{cframe=black!25 0.3pt}{\includegraphics[width=\dimexpr\linewidth-0.6pt\relax,height=3.35cm,keepaspectratio]{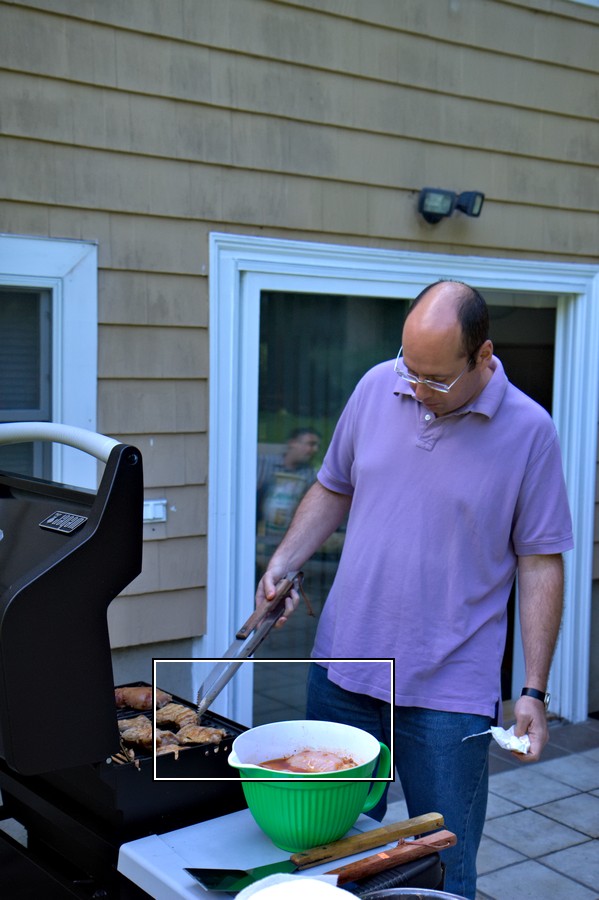}}};\node[anchor=south east, inner sep=1.2pt, fill=black!55, text=white, font=\scriptsize\bfseries, rounded corners=0.8pt] at ([xshift=-2pt, yshift=2pt]im.south east) {3/9 judges};\end{tikzpicture}\\[1.5pt]\adjustbox{cframe=black!25 0.3pt}{\includegraphics[width=\dimexpr\linewidth-0.6pt\relax]{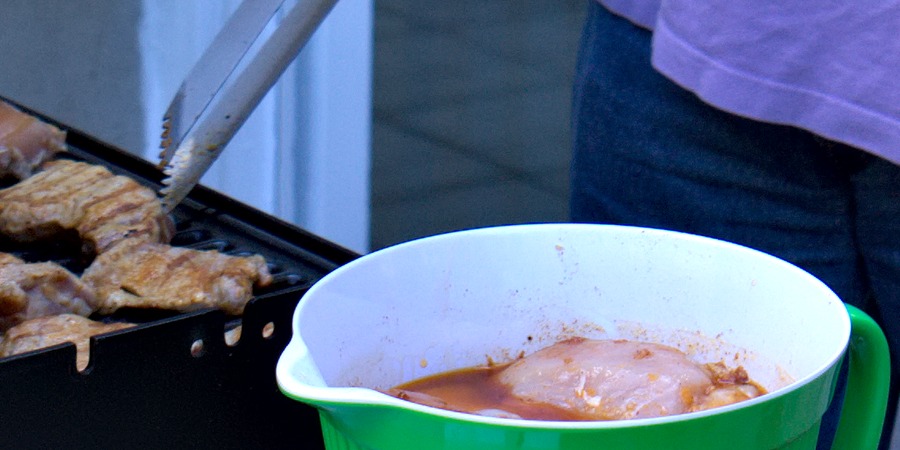}} & \centering \adjustbox{cframe=black!25 0.3pt}{\includegraphics[width=\dimexpr\linewidth-0.6pt\relax,height=3.35cm,keepaspectratio]{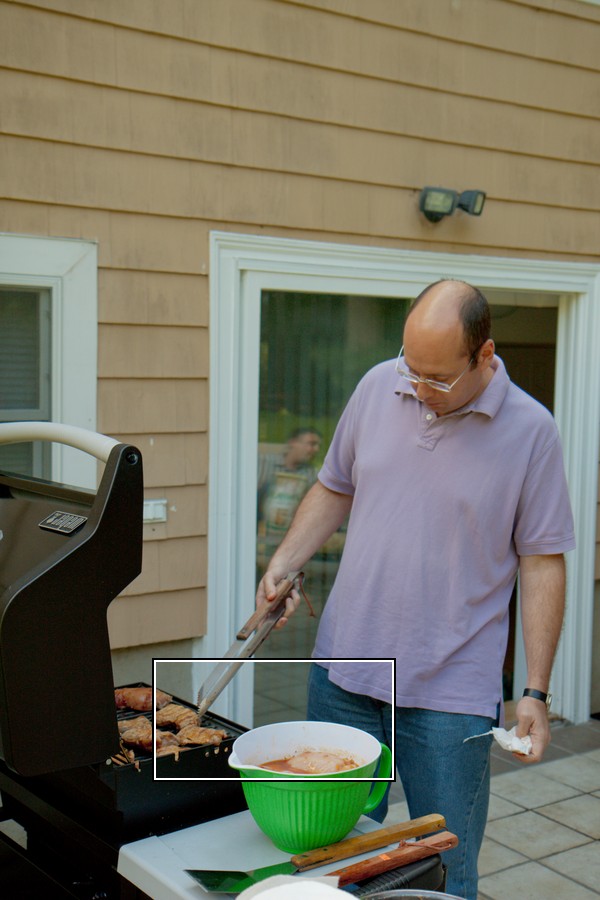}}\\[1.5pt]\adjustbox{cframe=black!25 0.3pt}{\includegraphics[width=\dimexpr\linewidth-0.6pt\relax]{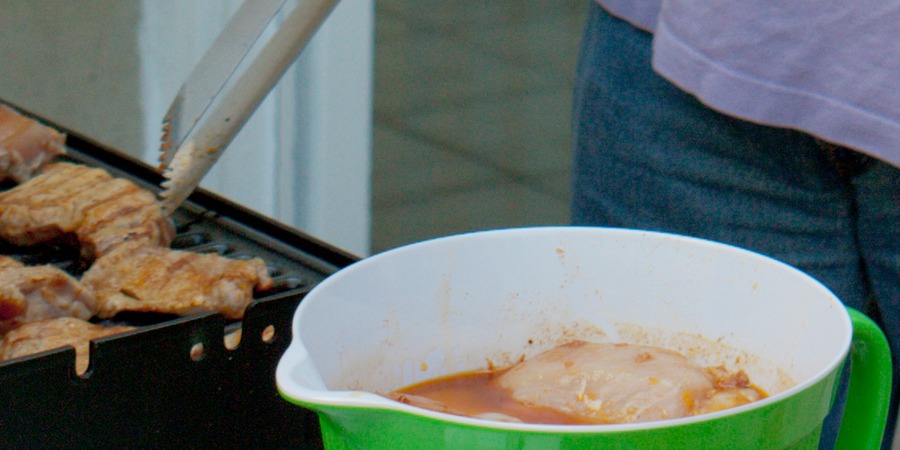}} \tabularnewline[7pt]
\rotatebox{90}{\parbox{3.35cm}{\centering\scriptsize Prefer Claude\\(10th pct.)}} & \centering \adjustbox{cframe=black!25 0.3pt}{\includegraphics[width=\dimexpr\linewidth-0.6pt\relax,height=3.35cm,keepaspectratio]{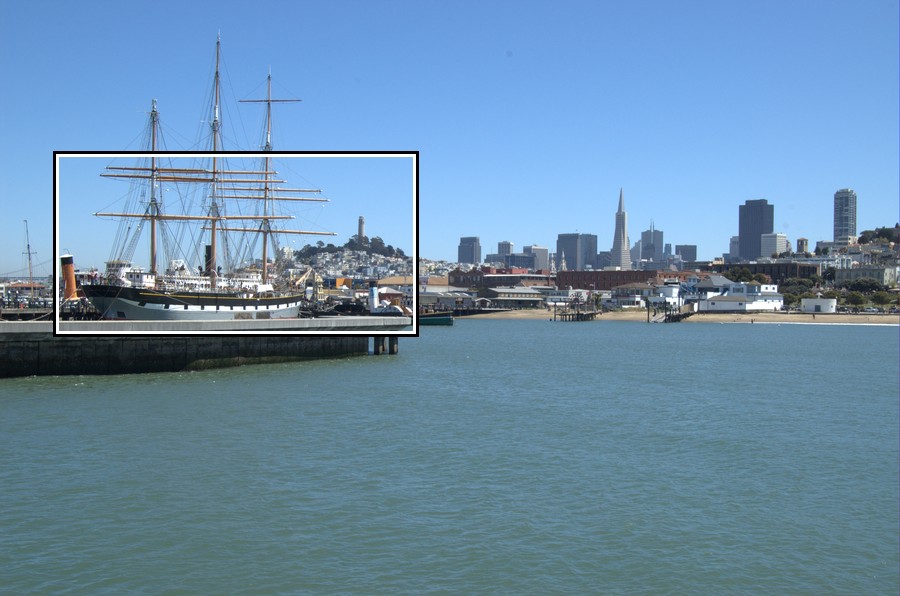}}\\[1.5pt]\adjustbox{cframe=black!25 0.3pt}{\includegraphics[width=\dimexpr\linewidth-0.6pt\relax]{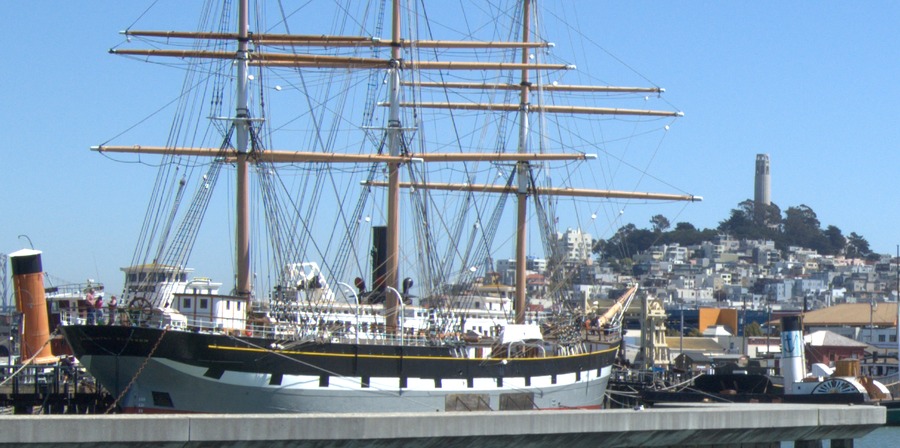}} & \centering \begin{tikzpicture}[baseline=(im.center)]\node[inner sep=0] (im) {\adjustbox{cframe=black!25 0.3pt}{\includegraphics[width=\dimexpr\linewidth-0.6pt\relax,height=3.35cm,keepaspectratio]{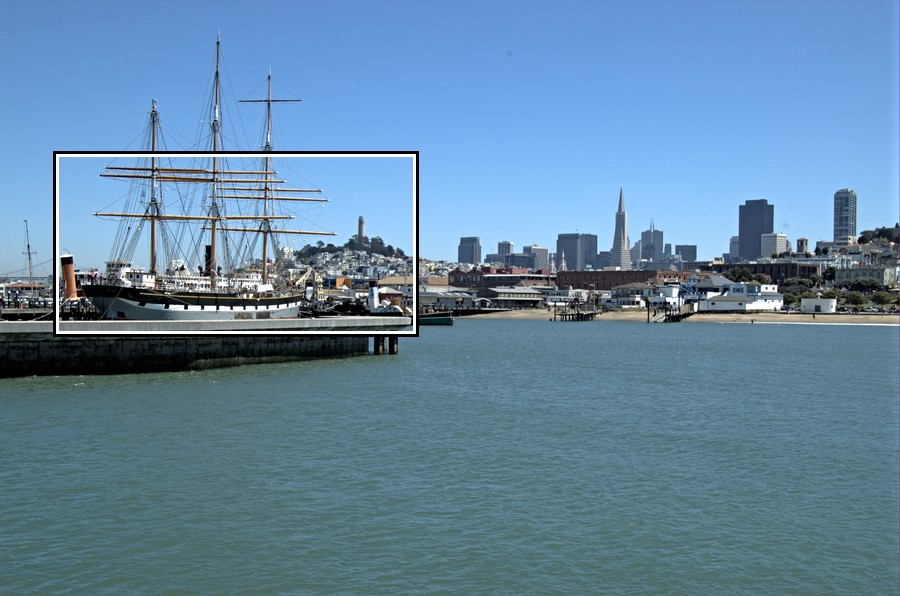}}};\node[anchor=south east, inner sep=1.2pt, fill=black!55, text=white, font=\scriptsize\bfseries, rounded corners=0.8pt] at ([xshift=-2pt, yshift=2pt]im.south east) {3/9 judges};\end{tikzpicture}\\[1.5pt]\adjustbox{cframe=black!25 0.3pt}{\includegraphics[width=\dimexpr\linewidth-0.6pt\relax]{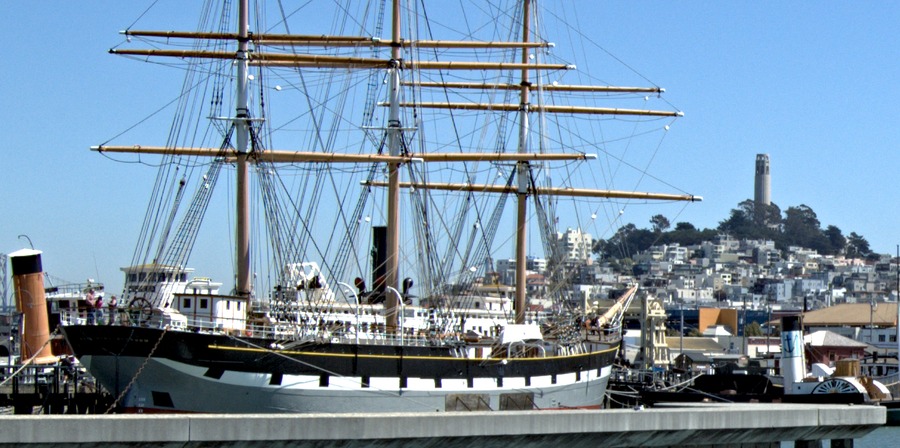}} & \centering \begin{tikzpicture}[baseline=(im.center)]\node[inner sep=0] (im) {\adjustbox{cframe=black!25 0.3pt}{\includegraphics[width=\dimexpr\linewidth-0.6pt\relax,height=3.35cm,keepaspectratio]{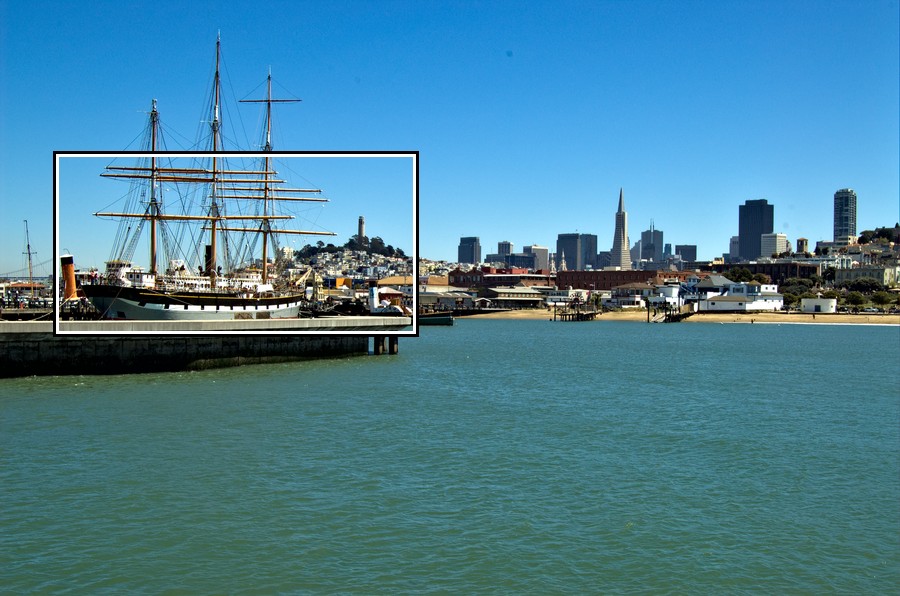}}};\node[anchor=south east, inner sep=1.2pt, fill=posgreen, text=white, font=\scriptsize\bfseries, rounded corners=0.8pt] at ([xshift=-2pt, yshift=2pt]im.south east) {6/9 judges};\end{tikzpicture}\\[1.5pt]\adjustbox{cframe=black!25 0.3pt}{\includegraphics[width=\dimexpr\linewidth-0.6pt\relax]{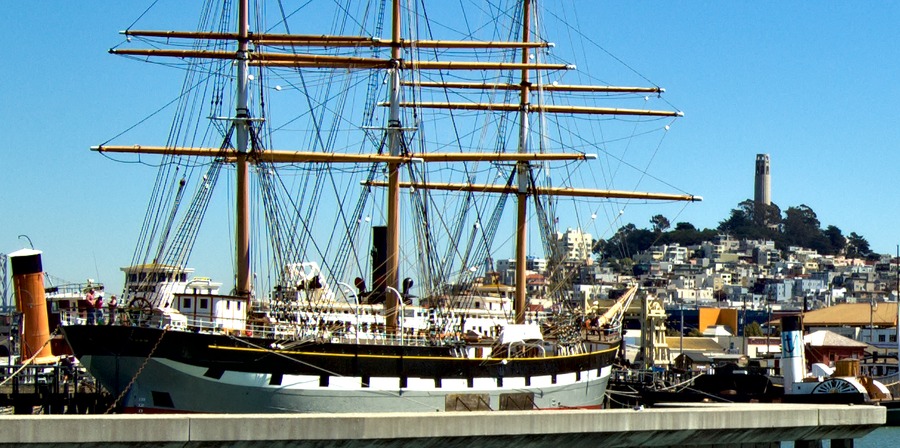}} & \centering \adjustbox{cframe=black!25 0.3pt}{\includegraphics[width=\dimexpr\linewidth-0.6pt\relax,height=3.35cm,keepaspectratio]{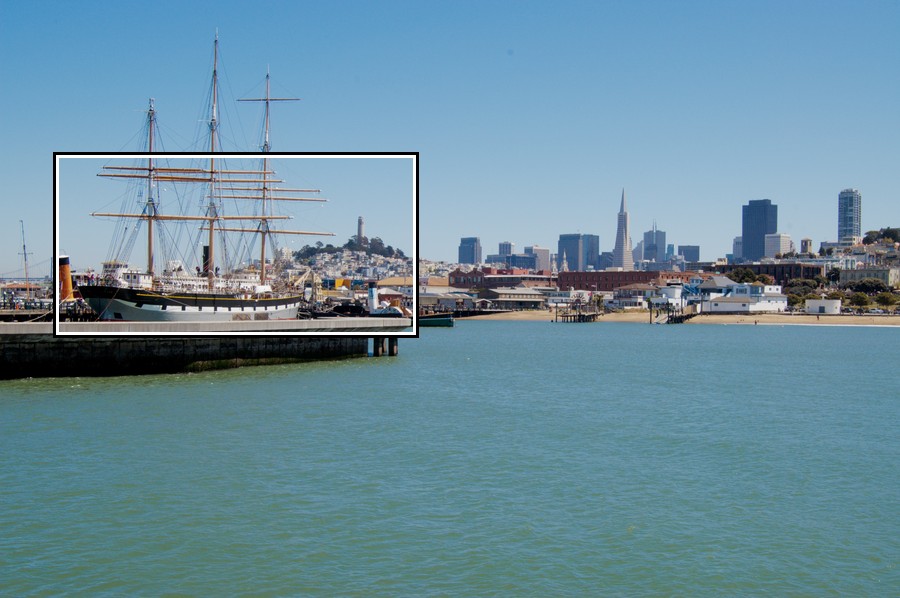}}\\[1.5pt]\adjustbox{cframe=black!25 0.3pt}{\includegraphics[width=\dimexpr\linewidth-0.6pt\relax]{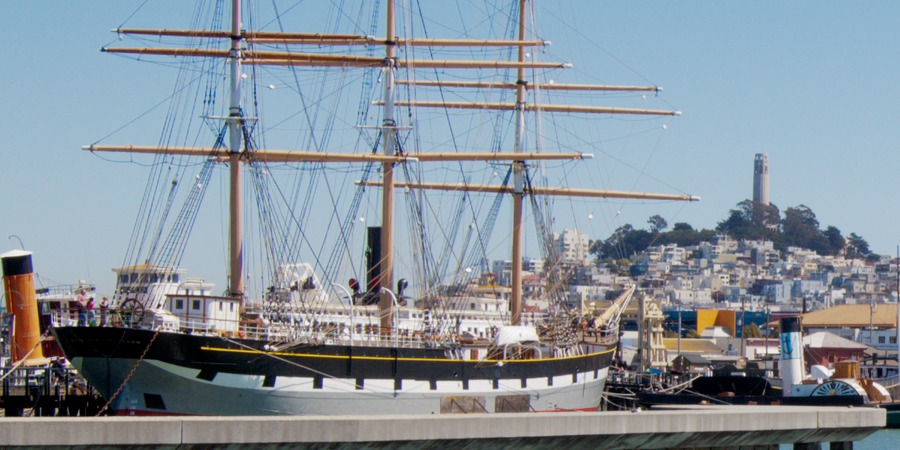}} \tabularnewline[7pt]
\end{tabular}
\caption{\textbf{MIT-Adobe FiveK: where the judges clearly prefer ATune (top), a typical photograph (middle) and where they clearly prefer Claude (bottom).} Right: the expert's edit.}
\label{fig:qual_fivek}
\end{figure*}

\begin{figure*}[p]
\centering
\setlength{\tabcolsep}{1.5pt}
\begin{tabular}{@{}m{0.035\linewidth}m{0.233\linewidth}m{0.233\linewidth}m{0.233\linewidth}m{0.233\linewidth}@{}}
 & \centering\small Original & \centering\small ATune & \centering\small Claude Opus 5 & \centering\small Retoucher \tabularnewline[2pt]
\rotatebox{90}{\parbox{3.35cm}{\centering\scriptsize Prefer ours\\(90th pct.)}} & \centering \adjustbox{cframe=black!25 0.3pt}{\includegraphics[width=\dimexpr\linewidth-0.6pt\relax,height=3.35cm,keepaspectratio]{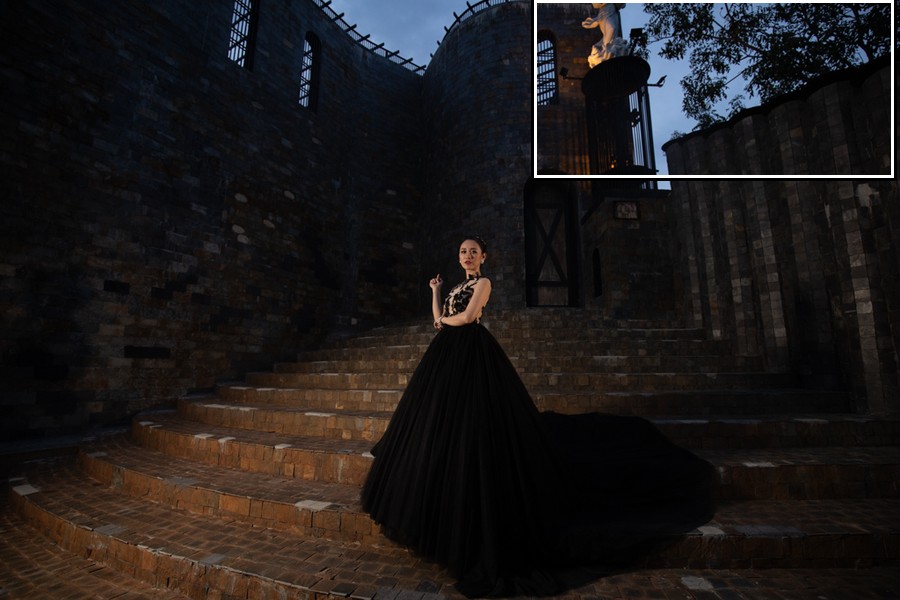}}\\[1.5pt]\adjustbox{cframe=black!25 0.3pt}{\includegraphics[width=\dimexpr\linewidth-0.6pt\relax]{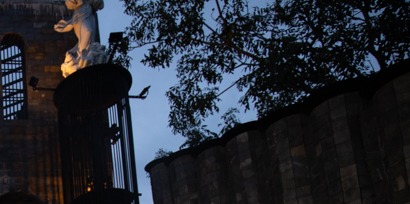}} & \centering \begin{tikzpicture}[baseline=(im.center)]\node[inner sep=0] (im) {\adjustbox{cframe=black!25 0.3pt}{\includegraphics[width=\dimexpr\linewidth-0.6pt\relax,height=3.35cm,keepaspectratio]{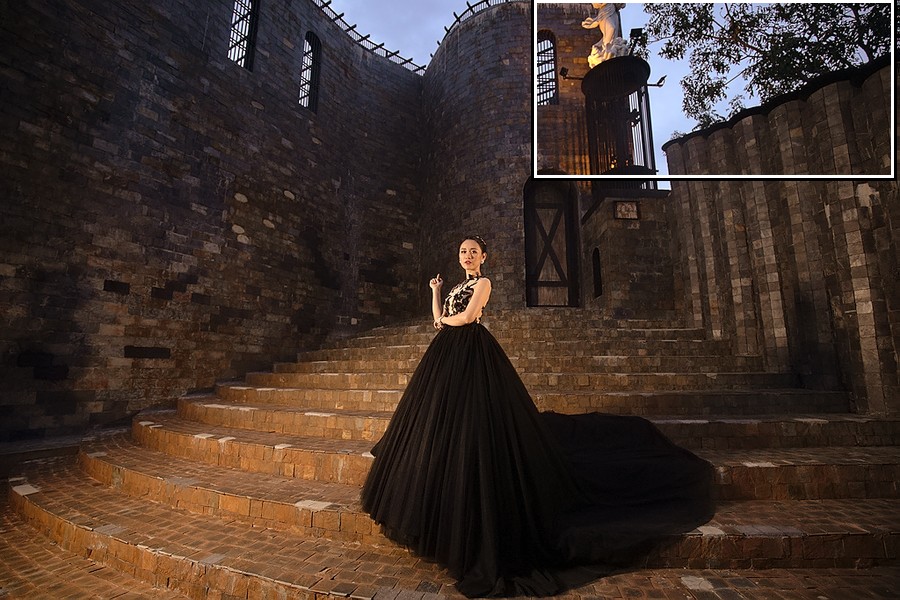}}};\node[anchor=south east, inner sep=1.2pt, fill=posgreen, text=white, font=\scriptsize\bfseries, rounded corners=0.8pt] at ([xshift=-2pt, yshift=2pt]im.south east) {8/9 judges};\end{tikzpicture}\\[1.5pt]\adjustbox{cframe=black!25 0.3pt}{\includegraphics[width=\dimexpr\linewidth-0.6pt\relax]{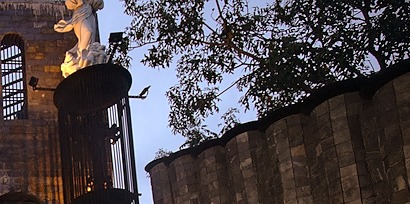}} & \centering \begin{tikzpicture}[baseline=(im.center)]\node[inner sep=0] (im) {\adjustbox{cframe=black!25 0.3pt}{\includegraphics[width=\dimexpr\linewidth-0.6pt\relax,height=3.35cm,keepaspectratio]{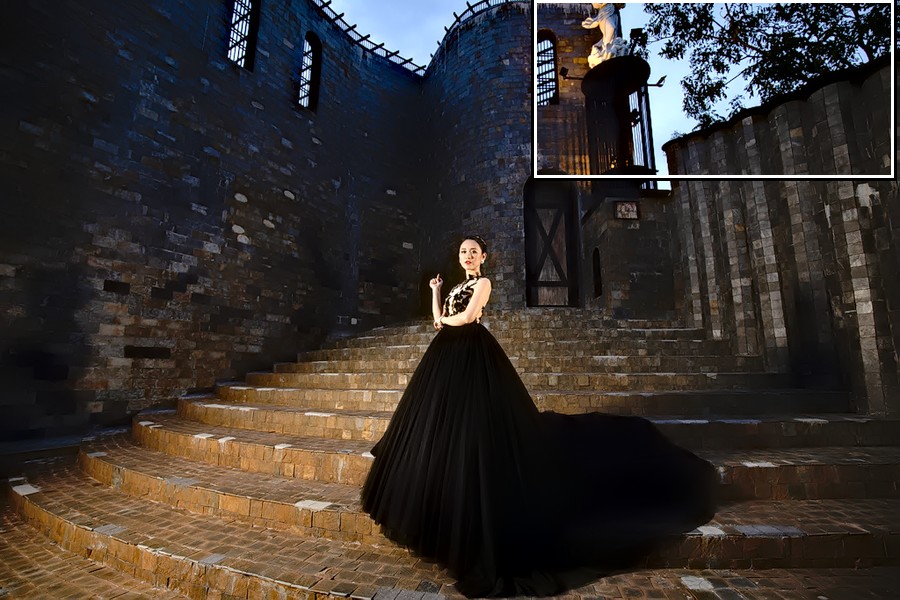}}};\node[anchor=south east, inner sep=1.2pt, fill=black!55, text=white, font=\scriptsize\bfseries, rounded corners=0.8pt] at ([xshift=-2pt, yshift=2pt]im.south east) {1/9 judges};\end{tikzpicture}\\[1.5pt]\adjustbox{cframe=black!25 0.3pt}{\includegraphics[width=\dimexpr\linewidth-0.6pt\relax]{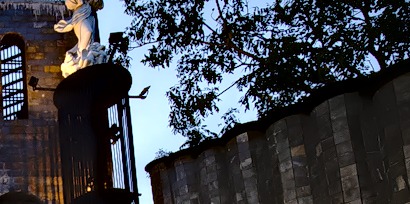}} & \centering \adjustbox{cframe=black!25 0.3pt}{\includegraphics[width=\dimexpr\linewidth-0.6pt\relax,height=3.35cm,keepaspectratio]{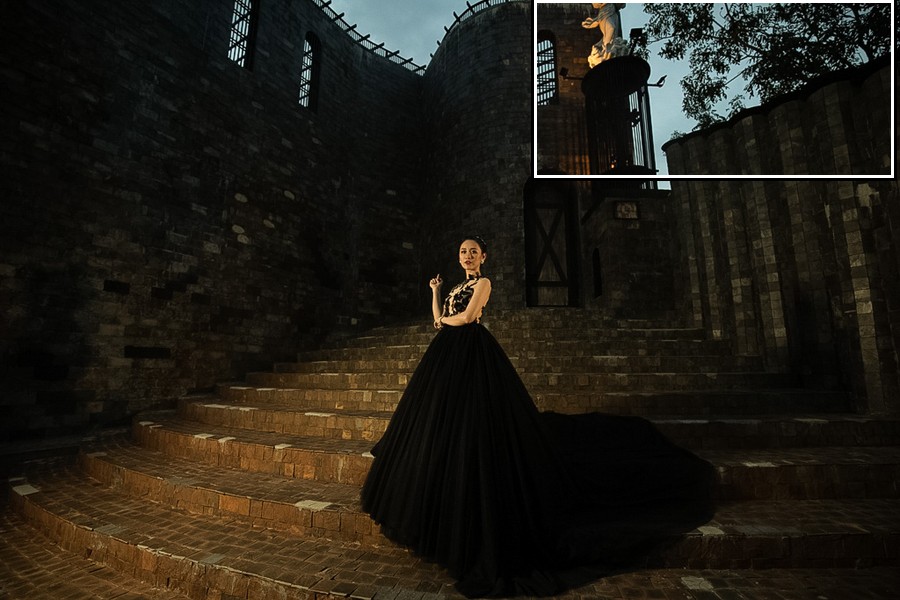}}\\[1.5pt]\adjustbox{cframe=black!25 0.3pt}{\includegraphics[width=\dimexpr\linewidth-0.6pt\relax]{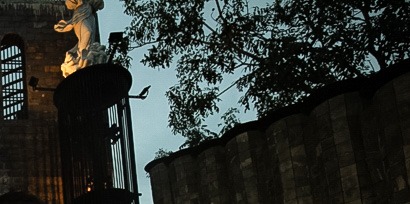}} \tabularnewline[7pt]
\rotatebox{90}{\parbox{3.35cm}{\centering\scriptsize Typical\\(median)}} & \centering \adjustbox{cframe=black!25 0.3pt}{\includegraphics[width=\dimexpr\linewidth-0.6pt\relax,height=3.35cm,keepaspectratio]{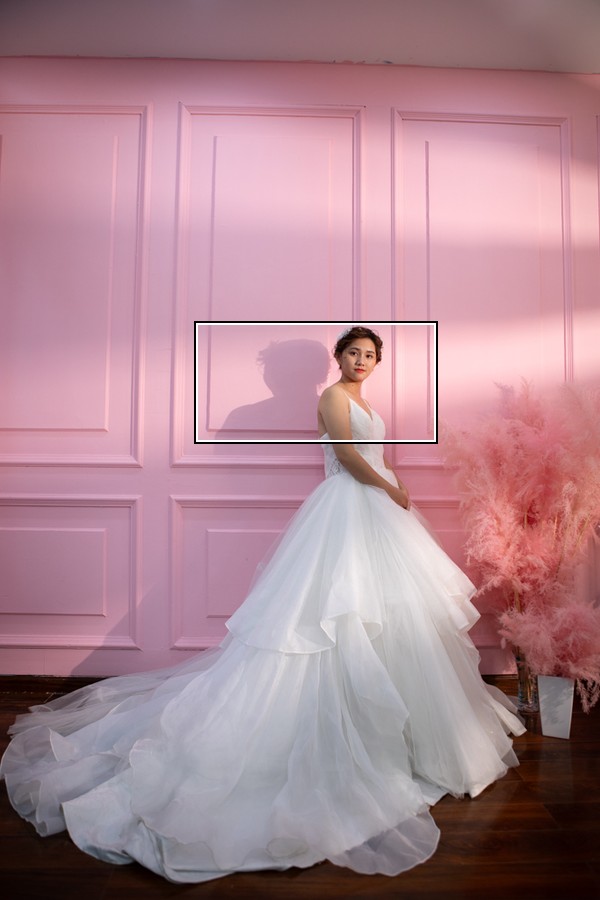}}\\[1.5pt]\adjustbox{cframe=black!25 0.3pt}{\includegraphics[width=\dimexpr\linewidth-0.6pt\relax]{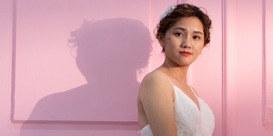}} & \centering \begin{tikzpicture}[baseline=(im.center)]\node[inner sep=0] (im) {\adjustbox{cframe=black!25 0.3pt}{\includegraphics[width=\dimexpr\linewidth-0.6pt\relax,height=3.35cm,keepaspectratio]{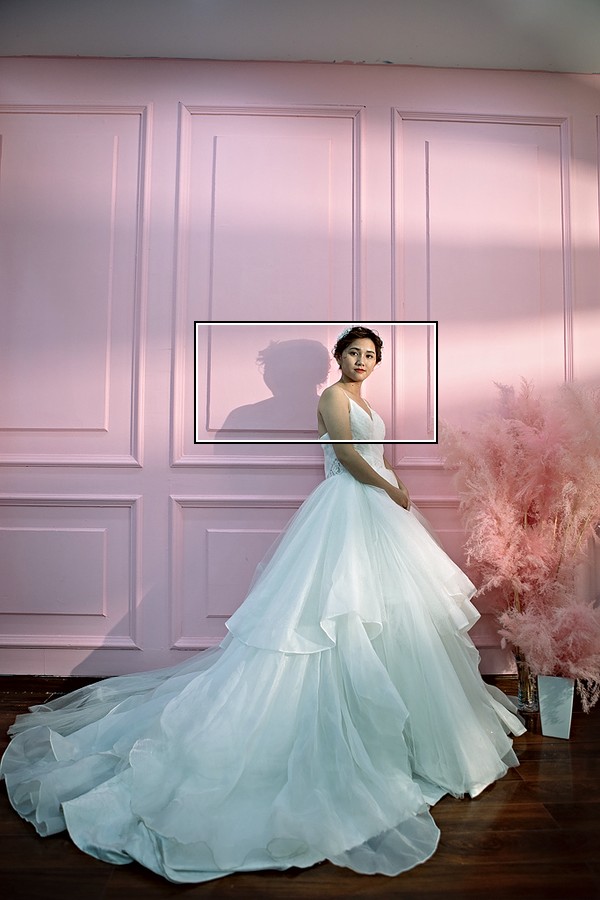}}};\node[anchor=south east, inner sep=1.2pt, fill=posgreen, text=white, font=\scriptsize\bfseries, rounded corners=0.8pt] at ([xshift=-2pt, yshift=2pt]im.south east) {5/9 judges};\end{tikzpicture}\\[1.5pt]\adjustbox{cframe=black!25 0.3pt}{\includegraphics[width=\dimexpr\linewidth-0.6pt\relax]{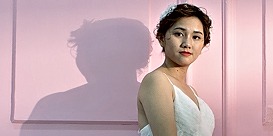}} & \centering \begin{tikzpicture}[baseline=(im.center)]\node[inner sep=0] (im) {\adjustbox{cframe=black!25 0.3pt}{\includegraphics[width=\dimexpr\linewidth-0.6pt\relax,height=3.35cm,keepaspectratio]{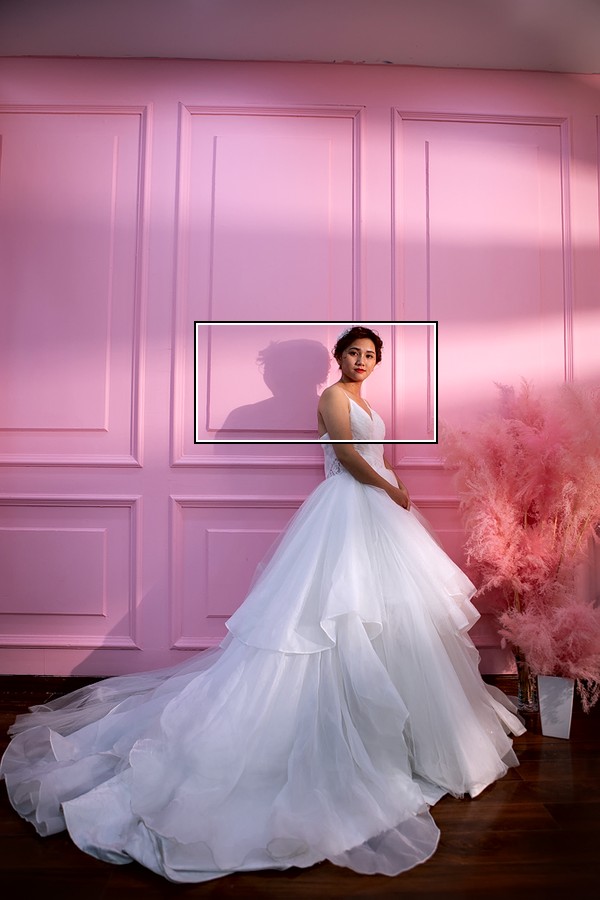}}};\node[anchor=south east, inner sep=1.2pt, fill=black!55, text=white, font=\scriptsize\bfseries, rounded corners=0.8pt] at ([xshift=-2pt, yshift=2pt]im.south east) {4/9 judges};\end{tikzpicture}\\[1.5pt]\adjustbox{cframe=black!25 0.3pt}{\includegraphics[width=\dimexpr\linewidth-0.6pt\relax]{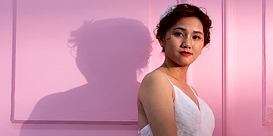}} & \centering \adjustbox{cframe=black!25 0.3pt}{\includegraphics[width=\dimexpr\linewidth-0.6pt\relax,height=3.35cm,keepaspectratio]{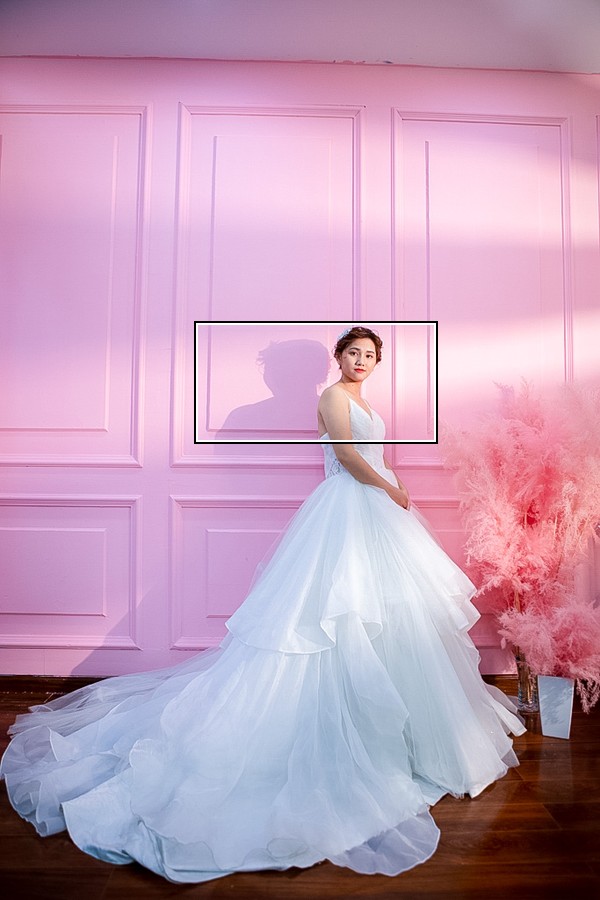}}\\[1.5pt]\adjustbox{cframe=black!25 0.3pt}{\includegraphics[width=\dimexpr\linewidth-0.6pt\relax]{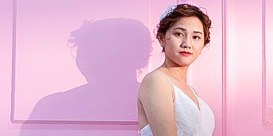}} \tabularnewline[7pt]
\rotatebox{90}{\parbox{3.35cm}{\centering\scriptsize Prefer Claude\\(10th pct.)}} & \centering \adjustbox{cframe=black!25 0.3pt}{\includegraphics[width=\dimexpr\linewidth-0.6pt\relax,height=3.35cm,keepaspectratio]{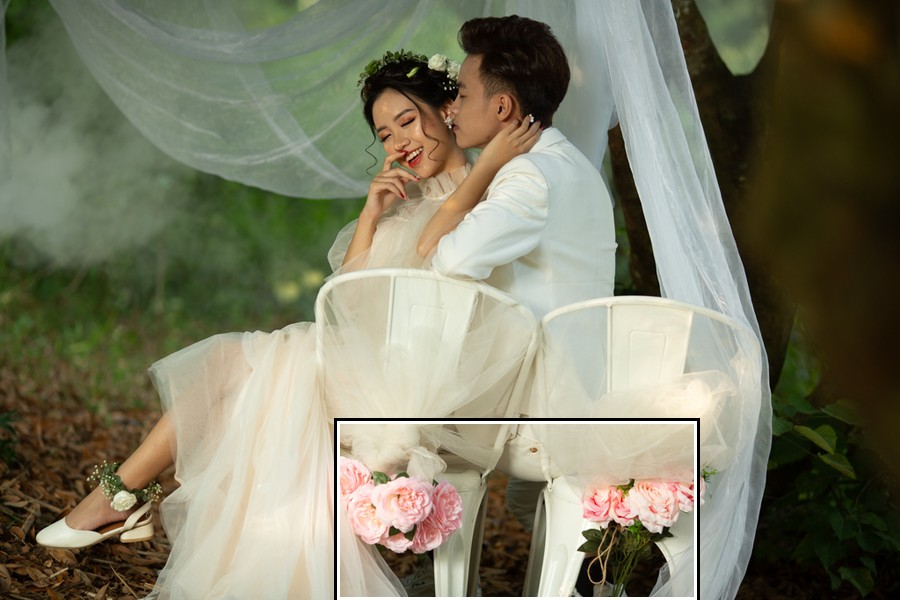}}\\[1.5pt]\adjustbox{cframe=black!25 0.3pt}{\includegraphics[width=\dimexpr\linewidth-0.6pt\relax]{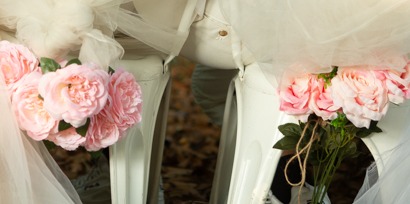}} & \centering \begin{tikzpicture}[baseline=(im.center)]\node[inner sep=0] (im) {\adjustbox{cframe=black!25 0.3pt}{\includegraphics[width=\dimexpr\linewidth-0.6pt\relax,height=3.35cm,keepaspectratio]{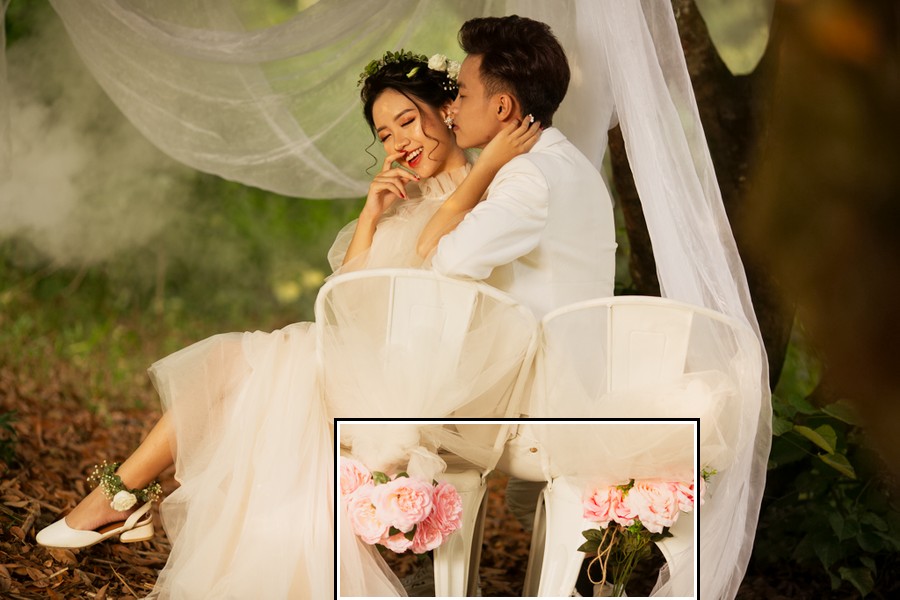}}};\node[anchor=south east, inner sep=1.2pt, fill=black!55, text=white, font=\scriptsize\bfseries, rounded corners=0.8pt] at ([xshift=-2pt, yshift=2pt]im.south east) {1/9 judges};\end{tikzpicture}\\[1.5pt]\adjustbox{cframe=black!25 0.3pt}{\includegraphics[width=\dimexpr\linewidth-0.6pt\relax]{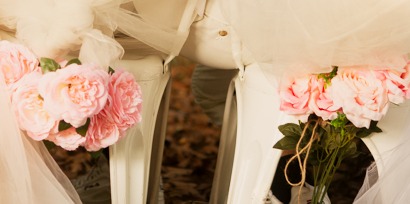}} & \centering \begin{tikzpicture}[baseline=(im.center)]\node[inner sep=0] (im) {\adjustbox{cframe=black!25 0.3pt}{\includegraphics[width=\dimexpr\linewidth-0.6pt\relax,height=3.35cm,keepaspectratio]{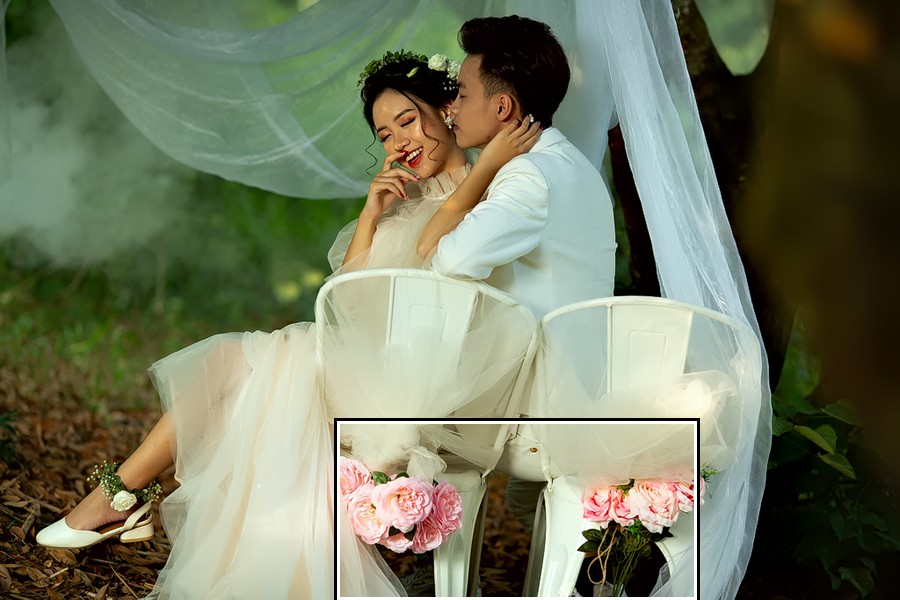}}};\node[anchor=south east, inner sep=1.2pt, fill=posgreen, text=white, font=\scriptsize\bfseries, rounded corners=0.8pt] at ([xshift=-2pt, yshift=2pt]im.south east) {8/9 judges};\end{tikzpicture}\\[1.5pt]\adjustbox{cframe=black!25 0.3pt}{\includegraphics[width=\dimexpr\linewidth-0.6pt\relax]{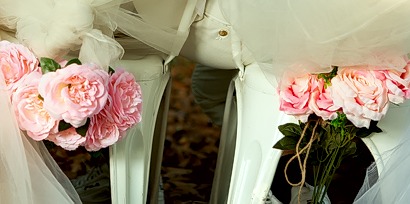}} & \centering \adjustbox{cframe=black!25 0.3pt}{\includegraphics[width=\dimexpr\linewidth-0.6pt\relax,height=3.35cm,keepaspectratio]{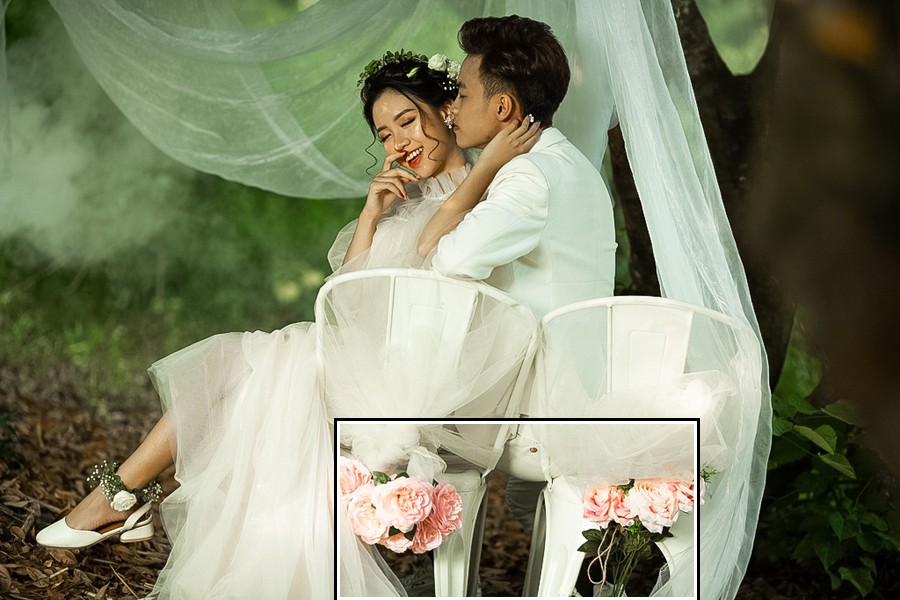}}\\[1.5pt]\adjustbox{cframe=black!25 0.3pt}{\includegraphics[width=\dimexpr\linewidth-0.6pt\relax]{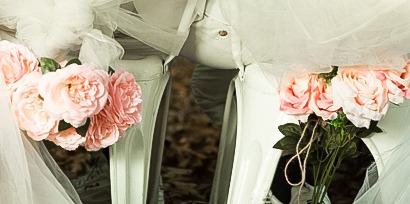}} \tabularnewline[7pt]
\end{tabular}
\caption{\textbf{PPR10K: where the judges clearly prefer ATune (top), a typical photograph (middle) and where they clearly prefer Claude (bottom).} Right: the expert's edit.}
\label{fig:qual_ppr10k}
\end{figure*}

\begin{figure*}[p]
\centering
\setlength{\tabcolsep}{1.5pt}
\begin{tabular}{@{}m{0.035\linewidth}m{0.310\linewidth}m{0.310\linewidth}m{0.310\linewidth}@{}}
 & \centering\small Original & \centering\small ATune & \centering\small Claude Opus 5 \tabularnewline[2pt]
\rotatebox{90}{\parbox{3.35cm}{\centering\scriptsize Prefer ours\\(90th pct.)}} & \centering \adjustbox{cframe=black!25 0.3pt}{\includegraphics[width=\dimexpr\linewidth-0.6pt\relax,height=3.35cm,keepaspectratio]{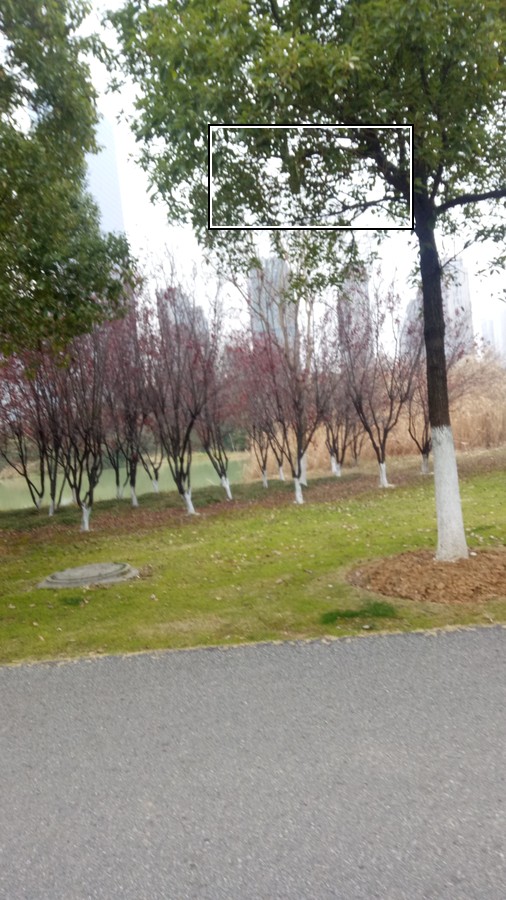}}\\[1.5pt]\adjustbox{cframe=black!25 0.3pt}{\includegraphics[width=\dimexpr\linewidth-0.6pt\relax]{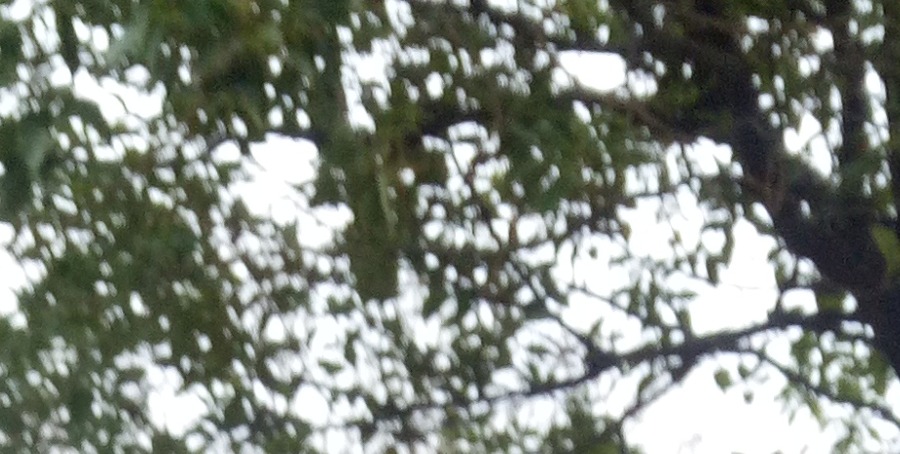}} & \centering \begin{tikzpicture}[baseline=(im.center)]\node[inner sep=0] (im) {\adjustbox{cframe=black!25 0.3pt}{\includegraphics[width=\dimexpr\linewidth-0.6pt\relax,height=3.35cm,keepaspectratio]{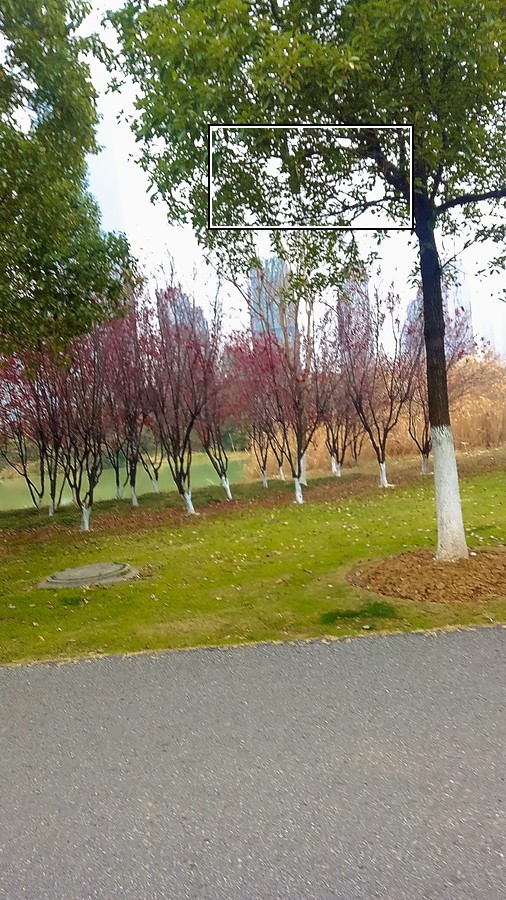}}};\node[anchor=south east, inner sep=1.2pt, fill=posgreen, text=white, font=\scriptsize\bfseries, rounded corners=0.8pt] at ([xshift=-2pt, yshift=2pt]im.south east) {7/9 judges};\end{tikzpicture}\\[1.5pt]\adjustbox{cframe=black!25 0.3pt}{\includegraphics[width=\dimexpr\linewidth-0.6pt\relax]{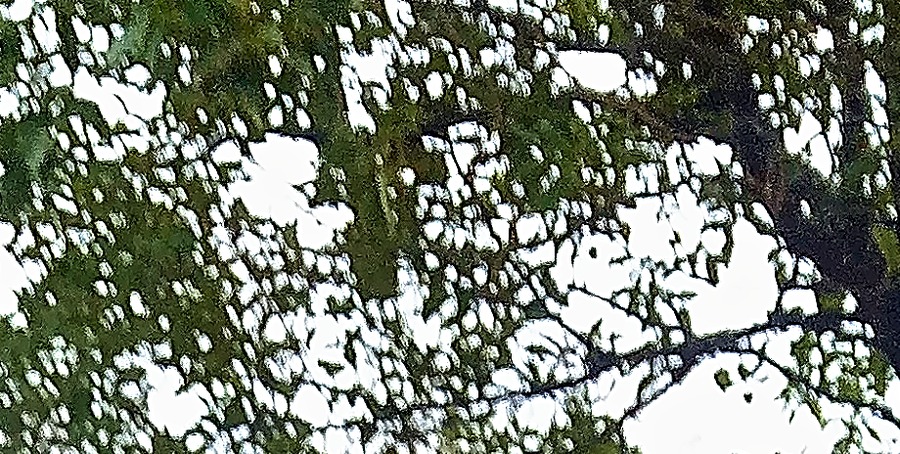}} & \centering \begin{tikzpicture}[baseline=(im.center)]\node[inner sep=0] (im) {\adjustbox{cframe=black!25 0.3pt}{\includegraphics[width=\dimexpr\linewidth-0.6pt\relax,height=3.35cm,keepaspectratio]{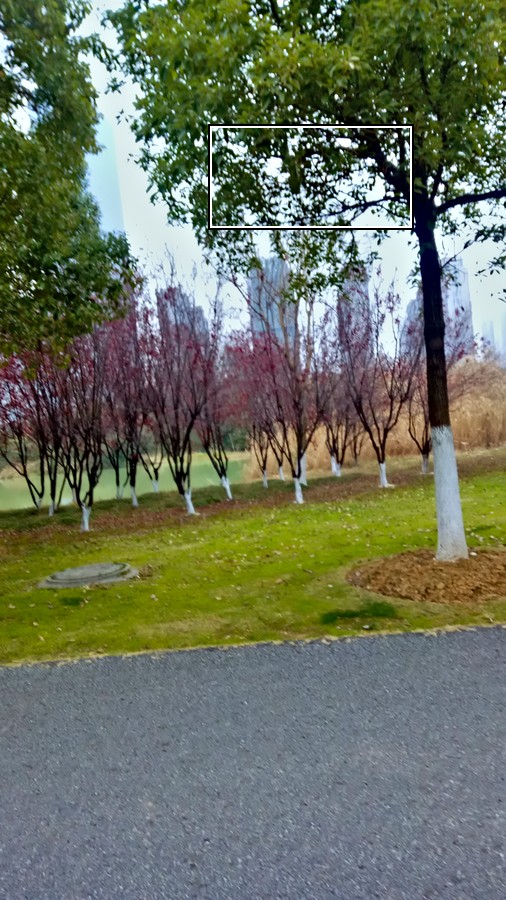}}};\node[anchor=south east, inner sep=1.2pt, fill=black!55, text=white, font=\scriptsize\bfseries, rounded corners=0.8pt] at ([xshift=-2pt, yshift=2pt]im.south east) {2/9 judges};\end{tikzpicture}\\[1.5pt]\adjustbox{cframe=black!25 0.3pt}{\includegraphics[width=\dimexpr\linewidth-0.6pt\relax]{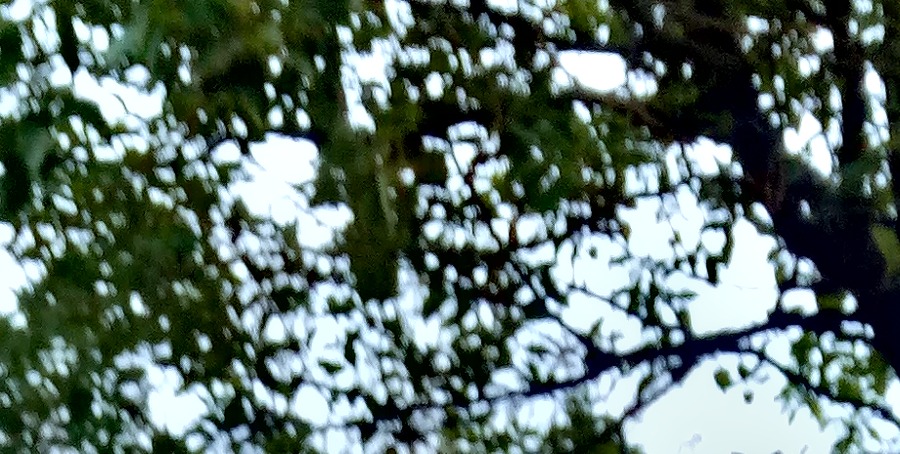}} \tabularnewline[7pt]
\rotatebox{90}{\parbox{3.35cm}{\centering\scriptsize Typical\\(median)}} & \centering \adjustbox{cframe=black!25 0.3pt}{\includegraphics[width=\dimexpr\linewidth-0.6pt\relax,height=3.35cm,keepaspectratio]{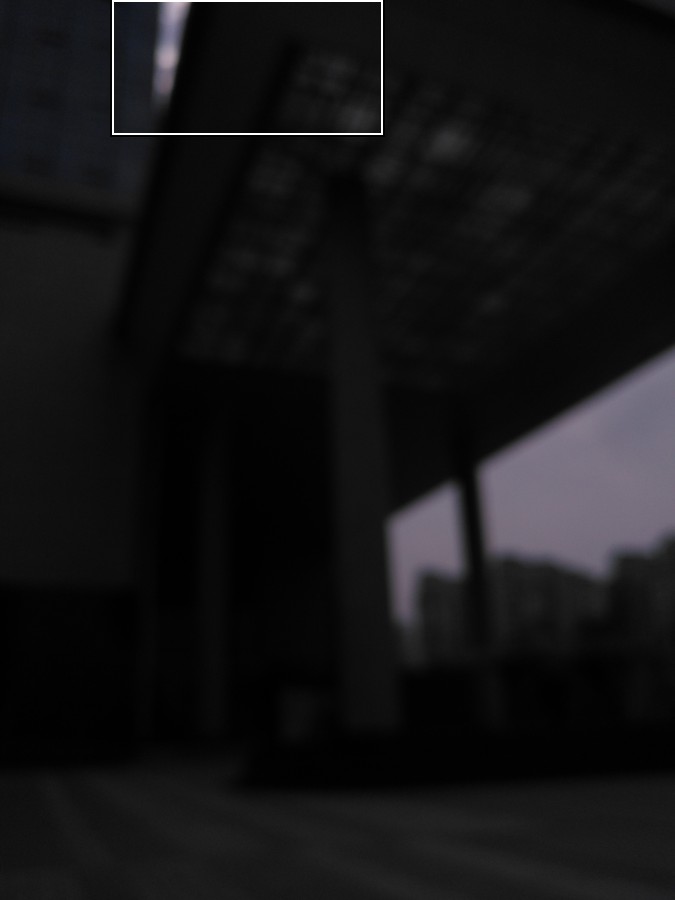}}\\[1.5pt]\adjustbox{cframe=black!25 0.3pt}{\includegraphics[width=\dimexpr\linewidth-0.6pt\relax]{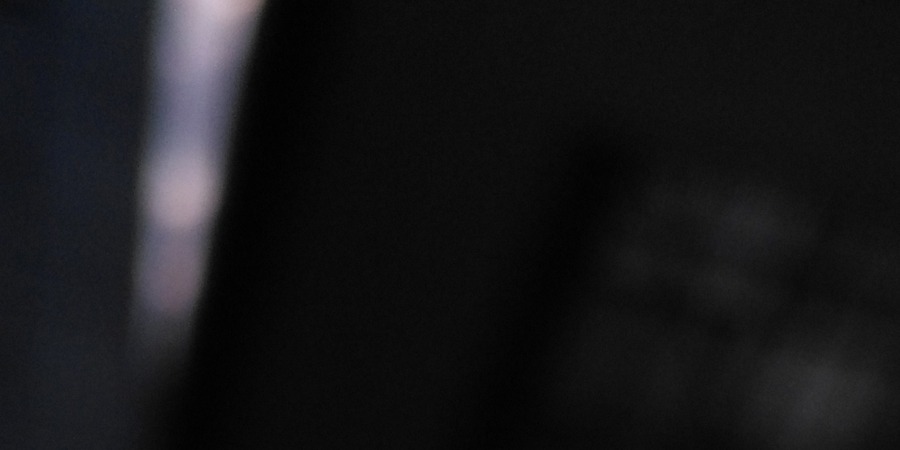}} & \centering \begin{tikzpicture}[baseline=(im.center)]\node[inner sep=0] (im) {\adjustbox{cframe=black!25 0.3pt}{\includegraphics[width=\dimexpr\linewidth-0.6pt\relax,height=3.35cm,keepaspectratio]{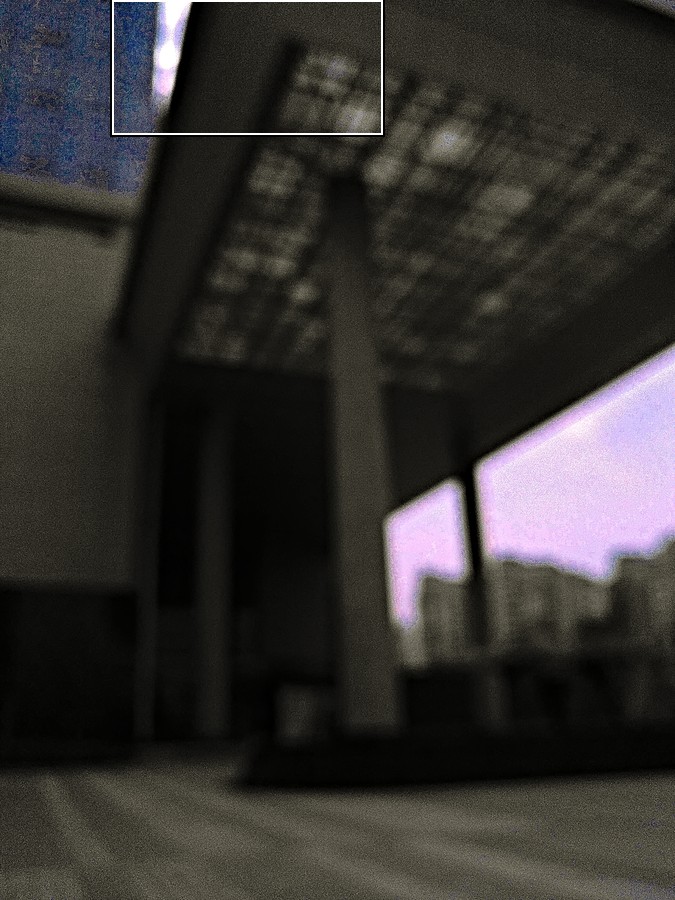}}};\node[anchor=south east, inner sep=1.2pt, fill=black!55, text=white, font=\scriptsize\bfseries, rounded corners=0.8pt] at ([xshift=-2pt, yshift=2pt]im.south east) {4/9 judges};\end{tikzpicture}\\[1.5pt]\adjustbox{cframe=black!25 0.3pt}{\includegraphics[width=\dimexpr\linewidth-0.6pt\relax]{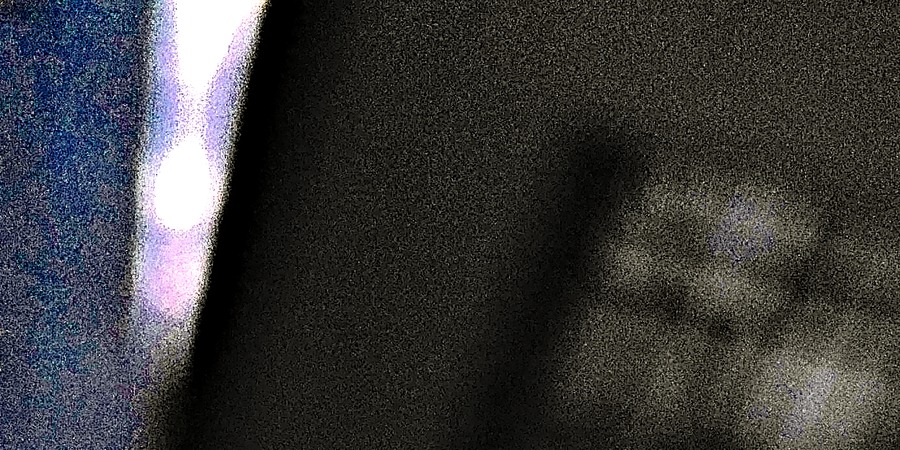}} & \centering \begin{tikzpicture}[baseline=(im.center)]\node[inner sep=0] (im) {\adjustbox{cframe=black!25 0.3pt}{\includegraphics[width=\dimexpr\linewidth-0.6pt\relax,height=3.35cm,keepaspectratio]{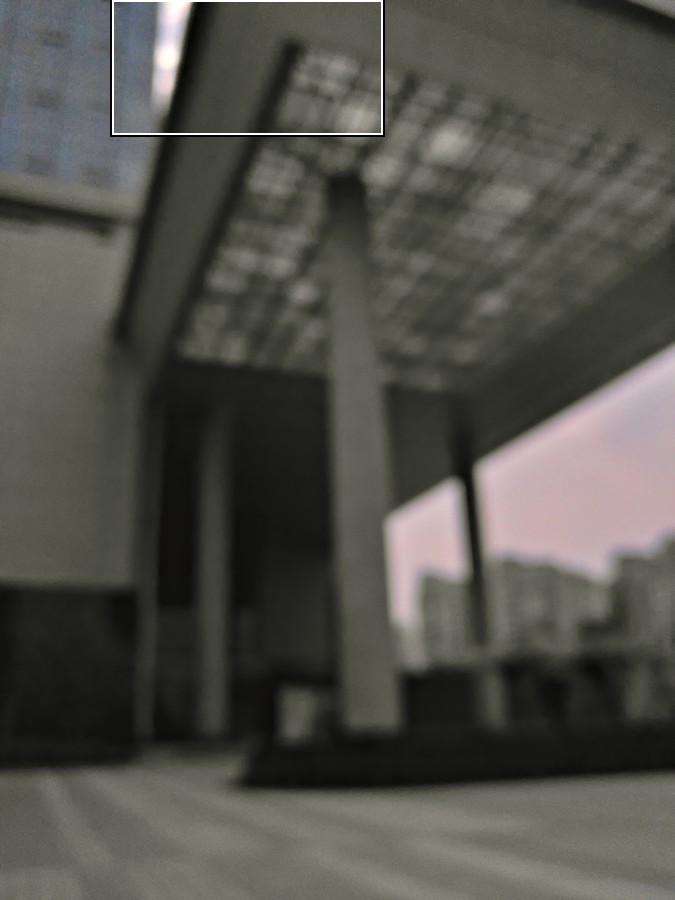}}};\node[anchor=south east, inner sep=1.2pt, fill=posgreen, text=white, font=\scriptsize\bfseries, rounded corners=0.8pt] at ([xshift=-2pt, yshift=2pt]im.south east) {5/9 judges};\end{tikzpicture}\\[1.5pt]\adjustbox{cframe=black!25 0.3pt}{\includegraphics[width=\dimexpr\linewidth-0.6pt\relax]{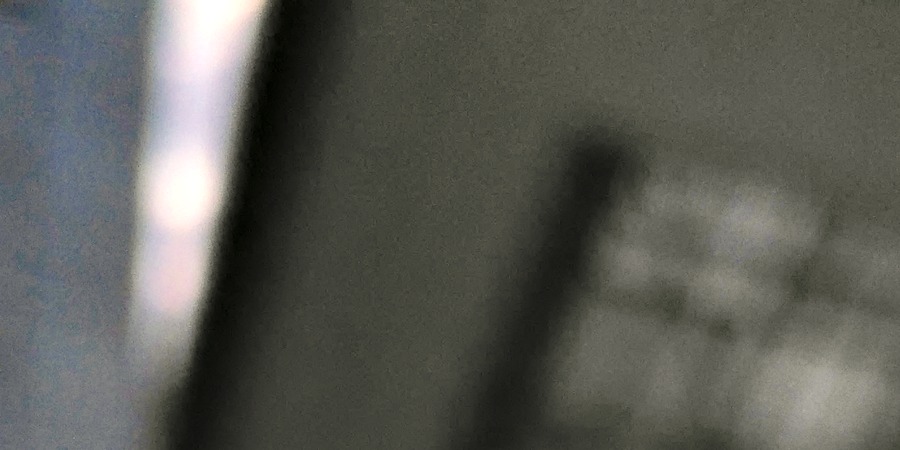}} \tabularnewline[7pt]
\rotatebox{90}{\parbox{3.35cm}{\centering\scriptsize Prefer Claude\\(10th pct.)}} & \centering \adjustbox{cframe=black!25 0.3pt}{\includegraphics[width=\dimexpr\linewidth-0.6pt\relax,height=3.35cm,keepaspectratio]{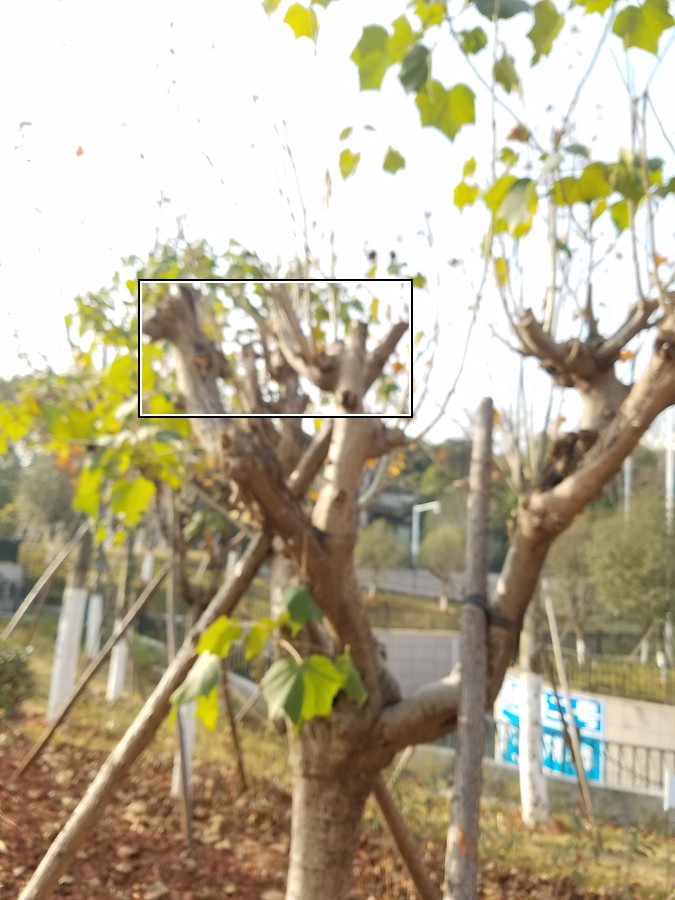}}\\[1.5pt]\adjustbox{cframe=black!25 0.3pt}{\includegraphics[width=\dimexpr\linewidth-0.6pt\relax]{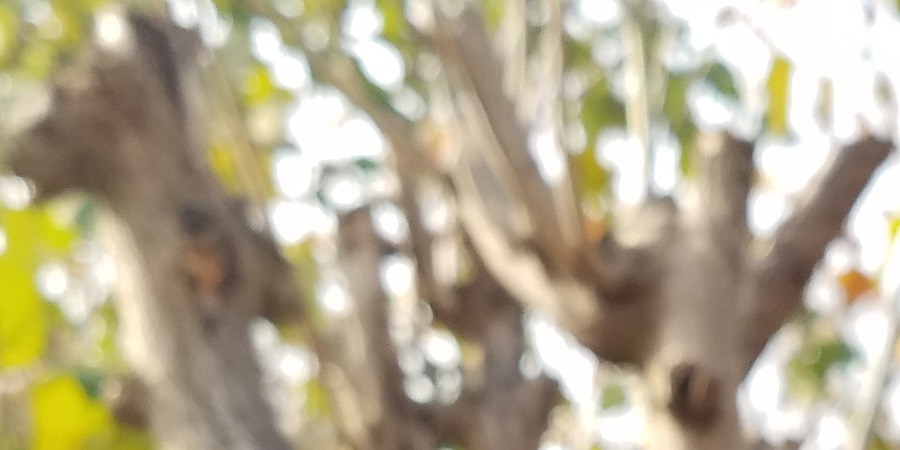}} & \centering \begin{tikzpicture}[baseline=(im.center)]\node[inner sep=0] (im) {\adjustbox{cframe=black!25 0.3pt}{\includegraphics[width=\dimexpr\linewidth-0.6pt\relax,height=3.35cm,keepaspectratio]{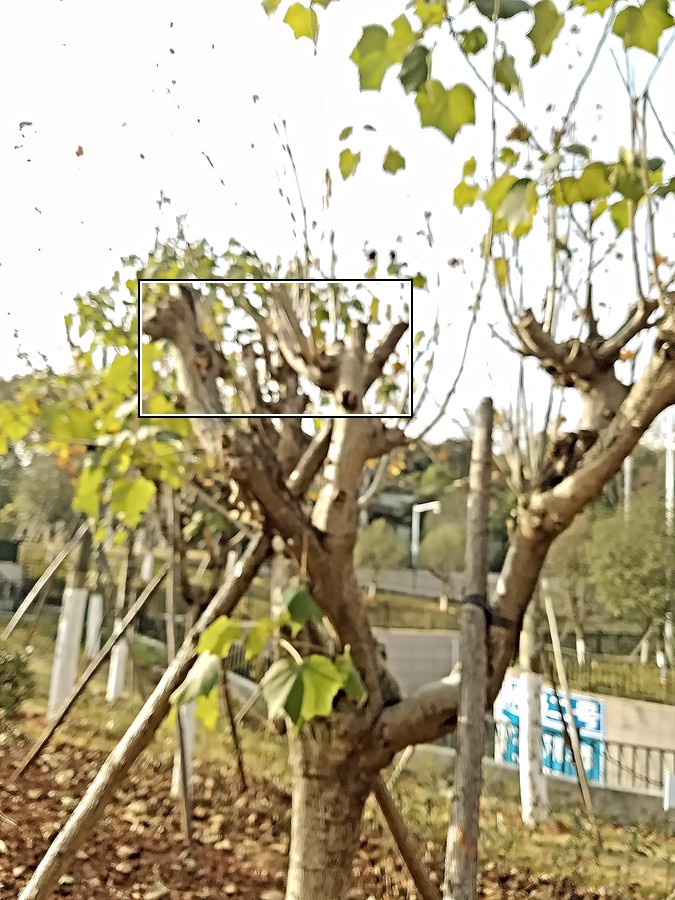}}};\node[anchor=south east, inner sep=1.2pt, fill=black!55, text=white, font=\scriptsize\bfseries, rounded corners=0.8pt] at ([xshift=-2pt, yshift=2pt]im.south east) {1/9 judges};\end{tikzpicture}\\[1.5pt]\adjustbox{cframe=black!25 0.3pt}{\includegraphics[width=\dimexpr\linewidth-0.6pt\relax]{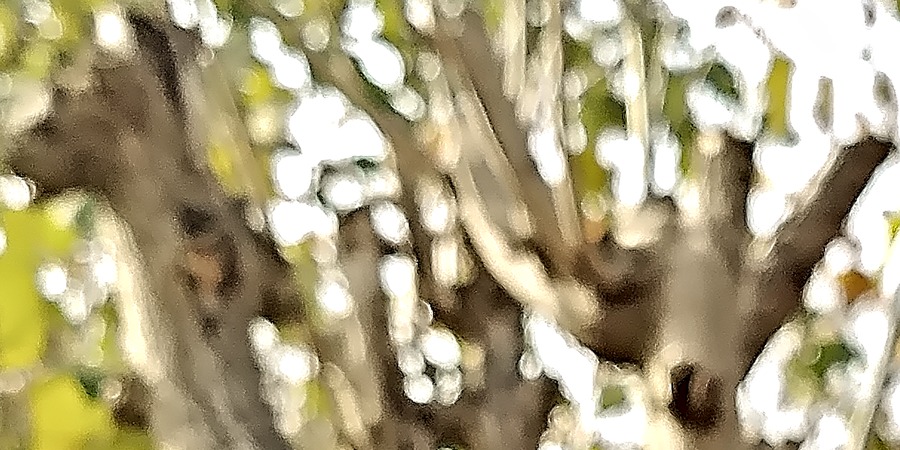}} & \centering \begin{tikzpicture}[baseline=(im.center)]\node[inner sep=0] (im) {\adjustbox{cframe=black!25 0.3pt}{\includegraphics[width=\dimexpr\linewidth-0.6pt\relax,height=3.35cm,keepaspectratio]{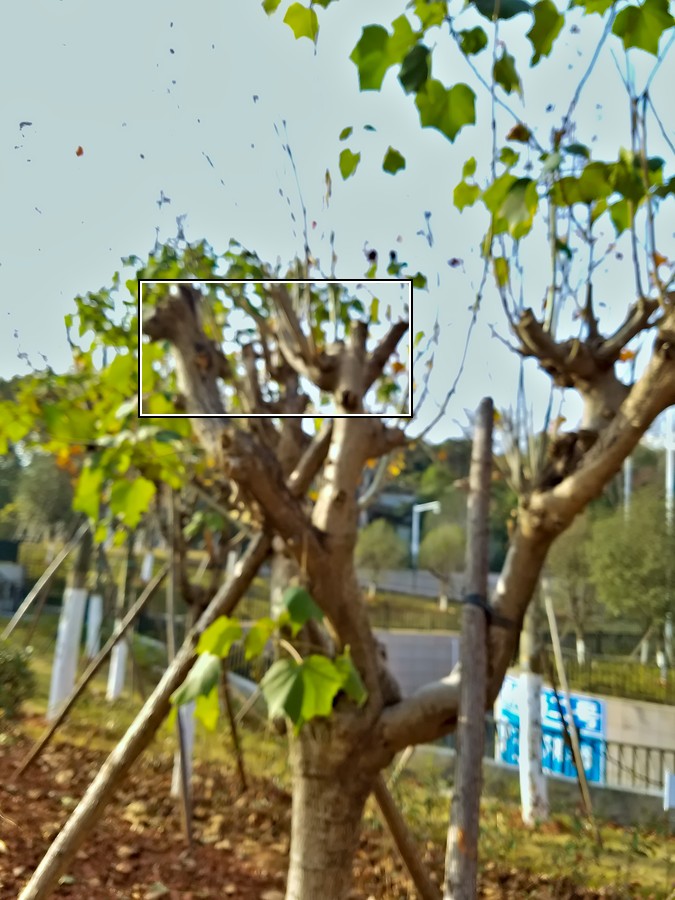}}};\node[anchor=south east, inner sep=1.2pt, fill=posgreen, text=white, font=\scriptsize\bfseries, rounded corners=0.8pt] at ([xshift=-2pt, yshift=2pt]im.south east) {8/9 judges};\end{tikzpicture}\\[1.5pt]\adjustbox{cframe=black!25 0.3pt}{\includegraphics[width=\dimexpr\linewidth-0.6pt\relax]{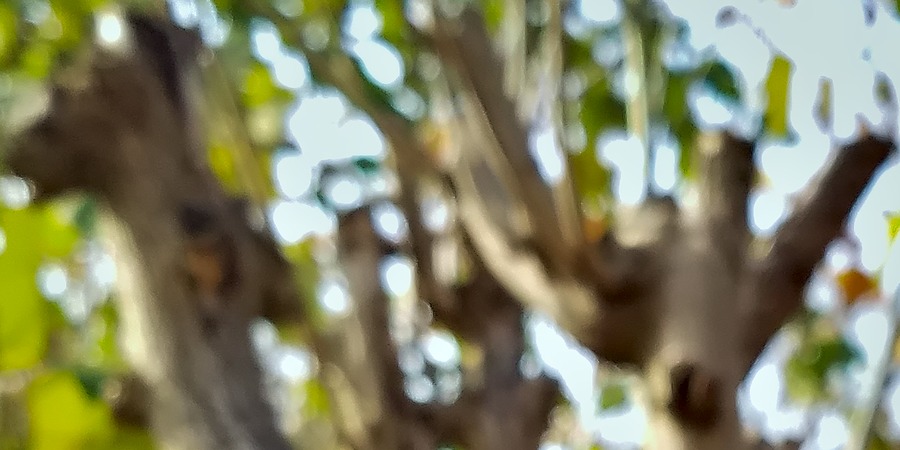}} \tabularnewline[7pt]
\end{tabular}
\caption{\textbf{SPAQ: where the judges clearly prefer ATune (top), a typical photograph (middle) and where they clearly prefer Claude (bottom).} SPAQ has no expert edit.}
\label{fig:qual_spaq}
\end{figure*}

\Cref{fig:showcase,fig:showcase2} show before-and-after pairs that we picked by hand for variety, among photographs where the
judges prefer ATune's edit to Claude's. \Cref{fig:qual_fivek,fig:qual_ppr10k,fig:qual_spaq} are not picked by hand.
For each test set, we rank every photograph by how much the nine judges prefer ATune's edit to Claude's, and show the
photographs at the \numQualPcts{} percentiles: one where the judges clearly prefer ATune, a typical one, and one where
they clearly prefer Claude.

\section{Reproducibility, data and impact}
\label{app:repro}

\paragraph{What we will release.} The code for every step: the attribute measures, the defects, training, the
solver, the editor drivers, the frontier-model runner, and the scripts that build every table. We will also release the
bucket scales, the test-set lists, and the per-photograph results behind each table.

\paragraph{Software and judges.} RawTherapee 5.8 (\texttt{rawtherapee-cli}) and darktable 3.8
(\texttt{darktable-cli}). \qalign{} (\texttt{q-future/one-align}, fp16), \liqe{} and the other five \texttt{pyiqa}
judges with their authors' weights, and DeQA-Score (\texttt{zhiyuanyou/DeQA-Score-Mix3}, fp16) with its authors'
scoring code.

\paragraph{Data.} The training photographs come from a licensed commercial stock collection and cannot be shared;
\mainsec{sec:corpus}{3.2} describes exactly how we sampled them. FiveK, PPR10K (with MMArt's exports), SPAQ and
KADID-10k are research datasets, used under their terms. The only data we collected about people is the human study's
votes, and we replaced the raters' account identifiers with random numbers (\cref{app:human}). PPR10K and SPAQ show
people, photographed and released by the datasets' authors; we only change the global colour and tone of their
photographs.

\paragraph{Impact.} ATune edits photographs only through global sliders, whose values are listed and can be undone.
It cannot add, remove or change what is in the scene, so it is harder to misuse for deception than generative
editors. Applied at scale, automatic improvement could make photographs look more alike, close to the style of the
training photographs, and for portraits could reinforce narrow ideas of appearance. The user keeps control: they set
the dial, see every slider value, and can leave a photograph unchanged.

\end{document}